\documentclass[9pt,twocolumn]{extarticle}
\usepackage{modernstyle}

\usepackage[round]{natbib}

\usepackage{hyperref}
\usepackage{url}
\usepackage{adjustbox}

\usepackage{booktabs}
\usepackage{multirow}
\usepackage{pifont}

\usepackage{bm}
\usepackage{subcaption}
\usepackage{siunitx}
\usepackage{amsfonts}
\usepackage{graphicx}
\usepackage{enumitem}
\usepackage{xcolor}
\usepackage{url}
\usepackage{fontawesome5}
\usepackage{bbold}
\usepackage{acmart-taps}
\usepackage{amssymb}

\DeclareMathAlphabet{\mathbb}{U}{bbold}{m}{n}

\usepackage{amsmath,amsfonts,bm}

\def\eqref#1{equation~\ref{#1}}
\def\1{\bm{1}}

\def\va{{\bm{a}}}
\def\vb{{\bm{b}}}
\def\vc{{\bm{c}}}

\def\vf{{\bm{f}}}

\def\vi{{\bm{i}}}

\def\vl{{\bm{l}}}
\def\vm{{\bm{m}}}

\def\vp{{\bm{p}}}

\def\vs{{\bm{s}}}

\def\vv{{\bm{v}}}
\def\vw{{\bm{w}}}
\def\vx{{\bm{x}}}

\def\eva{{a}}

\def\evf{{f}}

\def\evi{{i}}

\def\evl{{l}}

\def\evp{{p}}

\def\evs{{s}}

\def\evv{{v}}

\def\mW{{\bm{W}}}

\DeclareMathAlphabet{\mathsfit}{\encodingdefault}{\sfdefault}{m}{sl}
\SetMathAlphabet{\mathsfit}{bold}{\encodingdefault}{\sfdefault}{bx}{n}

\def\emW{{W}}

\DeclareMathOperator*{\argmax}{arg\,max}

\DeclareSIUnit\mmHg{mmHg}

\newcommand\GELU{\text{GELU}}

\definecolor{okgreen}{RGB}{2, 137, 59}
\definecolor{errorred}{RGB}{165, 25, 0}

\newcommand{\cmark}{\textcolor{okgreen}{\ding{51}}}%
\newcommand{\xmark}{\textcolor{errorred}{\ding{55}}}%

\newcommand\targetneuron{\ensuremath{t}}
\newcommand\model{\ensuremath{\mathcal{M}}}
\newcommand\expmethod{\ensuremath{\mathcal{E}}}
\newcommand\ldim[1]{\ensuremath{\text{dim}(#1)}}
\newcommand\vectdim[1]{\ensuremath{\mathbb{R}^{#1}}}
\newcommand\matdim[2]{\ensuremath{\mathbb{R}^{#1 \times #2}}}
\newcommand\oinputs[2]{\ensuremath{\vi^{(#1), #2}}}
\newcommand\oinputselem[3]{\ensuremath{\evi^{(#1), #2}_{#3}}}
\newcommand\indvector[2]{\ensuremath{\vv^{(#1), #2}}}
\newcommand\indvectorelem[3]{\ensuremath{\evv^{(#1), #2}_{#3}}}
\newcommand\nlayers{\ensuremath{N}}
\newcommand\currlayer{\ensuremath{n}}

\newcommand\scoresvec{\ensuremath{\vs}}
\newcommand\scoresvecelem[2]{\ensuremath{\evs^{{#1}}_{#2}}}
\newcommand\scoresvecn[2]{\ensuremath{\vs}^{(#1), {#2}}}
\newcommand\scoresvecnelem[3]{\ensuremath{\evs^{(#1), {#2}}_{#3}}}
\newcommand\scorebias{\ensuremath{s_{\text{bias}}}}
\newcommand\sconf{\ensuremath{s_{\text{conf}}}}
\newcommand\reference{\ensuremath{\text{ref}}}
\newcommand\featidx{\ensuremath{f}}
\newcommand\poss{\ensuremath{+}}
\newcommand\negs{\ensuremath{-}}
\newcommand\combs{\ensuremath{*}}
\newcommand\layer[1]{\ensuremath{L^{(#1)}}}
\newcommand\inputelem[1]{\ensuremath{x_{#1}}}
\newcommand\refinputelem[1]{\ensuremath{\inputelem{\reference, #1}}}
\newcommand\inputvec{\ensuremath{\vx}}
\newcommand\refinputvec{\ensuremath{\inputvec_{\reference}}}
\newcommand\actvec[1]{\ensuremath{\va^{(#1)}}}
\newcommand\refactvec[1]{\ensuremath{\actvec{#1}_{\reference}}}
\newcommand\actelem[2]{\ensuremath{\eva^{(#1)}_{#2}}}
\newcommand\refactelem[2]{\ensuremath{\eva^{(#1)}_{\reference, #2}}}

\newcommand\biasvec[1]{\ensuremath{\vb^{(#1)}}}

\newcommand\weightmat{\ensuremath{\mW}}
\newcommand\weightelem[3]{\ensuremath{\emW^{(#1)}_{#2, #3}}}
\newcommand\idxone{\ensuremath{i}}
\newcommand\idxtwo{\ensuremath{j}}
\newcommand\idxthree{\ensuremath{k}}
\newcommand\idxfour{\ensuremath{l}}
\newcommand\idxfive{\ensuremath{m}}
\newcommand\simpleactfun{\phi}

\newcommand\actvalelem[2]{\ensuremath{\eva^{(#1)}_{#2}}}

\newcommand\peak[4]{\ensuremath{\evp^{(#1), (#2 \rightarrow #3)}_{#4}}}
\newcommand\vpeak[3]{\ensuremath{\vp^{(#1), (#2 \rightarrow #3)}}}
\newcommand\linear[4]{\ensuremath{\evl^{(#1), (#2 \rightarrow #3)}_{#4}}}
\newcommand\vlinear[3]{\ensuremath{\vl^{(#1), (#2 \rightarrow #3)}}}
\newcommand\multiplier[4]{\ensuremath{m^{(#1), (#2 \rightarrow #3)}_{#4}}}
\newcommand\vmultiplier[3]{\ensuremath{\vm^{(#1), (#2 \rightarrow #3)}}}
\newcommand\cancelconst[1]{\ensuremath{c}}

\newcommand\vfoundation{\vf}
\newcommand\vfoundationelem[1]{\evf_{#1}}
\newcommand\relu{\text{ReLU}}

\newcommand\clip{\text{\normalfont clip}}
\newcommand\warning{\textcolor{errorred}{\faExclamationTriangle}}
\DeclareMathOperator{\sgn}{sgn}
\DeclareMathOperator{\remove}{rm}
\DeclareMathOperator{\conf}{conf}
\DeclareMathOperator{\counter}{counter}
\DeclareMathOperator{\countermag}{countermag}
\DeclareMathOperator{\cmag}{cmag}

\usepackage{amsthm}
\theoremstyle{plain}
\newtheorem{theorem}{Theorem}
\newtheorem{proposition}{Proposition}

\newtheorem{myproof}{Proof}
\newtheorem{example}{Example}
\theoremstyle{definition}
\newtheorem{definition}{Definition}

\newtheorem{property}{Property}
\theoremstyle{remark}

\usepackage{fancyhdr}
\fancypagestyle{firstpage}{%
	\fancyhf{}
	\setlength{\headsep}{0.3in}
	\cfoot{\thepage}
}
\fancypagestyle{otherpages}{%
	\fancyhf{}
	\setlength{\headsep}{0.3in}
	\rhead{Hidden Conflicts in Neural Networks}
	\lhead{A. Dejl, D. Zhang, H. Ayoobi, M. Williams and F. Toni}
	\cfoot{\thepage}
}

\title{Hidden Conflicts in Neural Networks\\ and Their Implications for Explainability}

\author {
	Adam Dejl$^1$, Dekai Zhang$^1$, Hamed Ayoobi$^1$, Matthew Williams$^2$ and Francesca Toni$^1$\\
	\large
	$^1$Department of Computing, Imperial College London \\
	$^2$Department of Surgery \& Cancer, Imperial College London \\
	\medskip
	\{adam.dejl18, dz819, h.ayoobi, matthew.williams, ft\}@imperial.ac.uk
}

\begin{document}
	\pagestyle{otherpages}
	\maketitle
	
	\begin{abstract}
		Artificial Neural Networks (ANNs) often represent conflicts between features, arising naturally during training as the network learns to integrate diverse and potentially disagreeing inputs to better predict the target variable. Despite their relevance to the ``reasoning'' processes of these models, the properties and implications of conflicts for understanding and explaining ANNs remain underexplored. In this paper, we develop a rigorous theory of conflicts in ANNs and demonstrate their impact on ANN explainability through two case studies. In the first case study, we use our theory of conflicts to inspire the design of a novel feature attribution method, which we call Conflict-Aware Feature-wise Explanations (CAFE). CAFE separates the positive and negative influences of features and biases, enabling more faithful explanations for models applied to tabular data. In the second case study, we take preliminary steps towards understanding the role of conflicts in out-of-distribution (OOD) scenarios. Through our experiments, we identify potentially useful connections between model conflicts and different kinds of distributional shifts in tabular and image data. Overall, our findings demonstrate the importance of accounting for conflicts in the development of more reliable explanation methods for AI systems, which are crucial for the beneficial use of these systems in the society.
	\end{abstract}
	
	\section{Introduction}
	\label{sec:introduction}
	During both training and inference, artificial neural networks (ANNs) are commonly faced with conflicts, as the different input features can often provide partially disagreeing accounts of the considered scenario. For example, consider a diagnostic AI system predicting whether a patient is likely to be suffering from a heart attack based on their symptoms. In the presence of chest pain, this system may predict a higher likelihood of a heart attack, but this effect may be counteracted by an indication that patient's symptoms worsen with breathing and movement (making cardiac causes of the pain considerably less likely). Such conflicts can be external (between different input features) or internal (between different internal representations) and are an important factor in the ``reasoning'' of a model.
	
	\begin{figure*}[!ht]
		\centering
		\begin{subfigure}[m]{0.45\textwidth}
			\centering
			\includegraphics[width=1.0\textwidth]{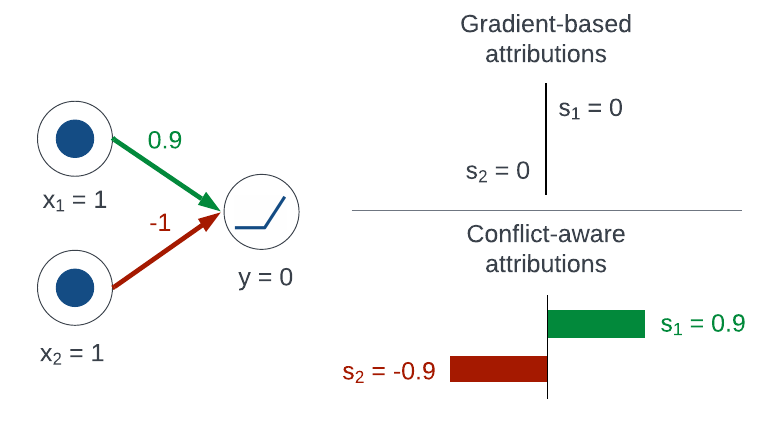}
			\caption{Feature attributions in the presence of a feature conflict}
			\label{fig:example-cancelation-heart-attack}
		\end{subfigure}
		\begin{subfigure}[m]{0.53\textwidth}
			\centering
			\includegraphics[width=0.6\textwidth]{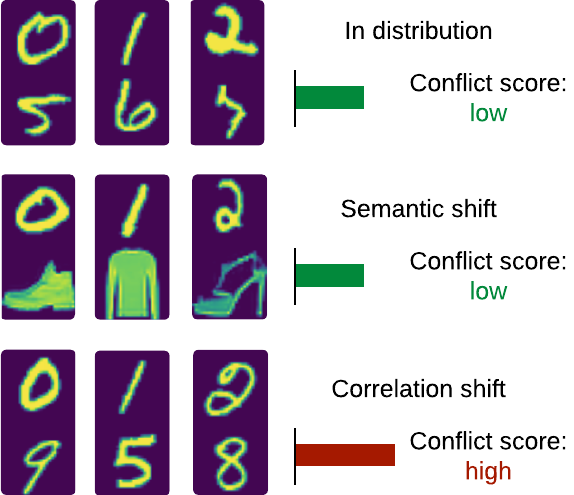}
			\caption{Behavior of model conflicts on OOD data}
			\label{fig:conflicts-ood-illustration}
		\end{subfigure}
		\caption{An illustration of our case studies. \textbf{(a)} While both $x_1$ (presence of chest pain) and $x_2$ (pain worsening with breathing and movement) play a crucial role towards prediction $y$ (risk of heart attack), their importance scores $s_1$ and $s_2$ are null with gradient-based attribution methods (top right). Instead, our proposed conflict-aware attributions (bottom right) reflect the underpinning conflict. \textbf{(b)} We identify a potentially useful connection between model conflicts and different kinds of distributional shift.}
		\label{fig:figure-1}
	\end{figure*}
	
	Several kinds of conflicts have been previously studied in the literature on ANNs, including conflicts between the labels of similar samples \citep{ledesma-on-removing-conflicts}, conflicts between different loss terms \citep{config-conflict-free-training} and, notably for our work, conflicts between canceling features \citep{shrikumar-deeplift}. However, the past treatment of such conflicts has been mostly anecdotal with no precise definition, analysis of their prevalence or a thorough investigation of their impact on ANN explainability.
	
	To motivate the importance of incorporating conflicts in the explanations of machine learning models, consider again the diagnostic AI system tasked with predicting heart attack likelihood and a simple neural unit that implements the associated behavior, visualised in Figure \ref{fig:example-cancelation-heart-attack}. Although the presence of chest pain and the observation that the pain worsens with breathing are both important for arriving at the final prediction, gradient-based methods applied to the given neuron return zero importance scores for both of these features, misleadingly concluding that they were completely irrelevant to the model's decision-making process. While this is a toy example, our analysis in this paper (Section \ref{sec:conflict-prevalence}) shows that the emergence of conflicts in models is not negligible, as, on average, $49\%$ of neurons in multilayer perceptrons (MLPs) are engaged in conflicts neglected by gradient-based attribution methods.
	
	In this paper, we introduce a rigorous theoretical framework for reasoning about conflicts in ANNs, and conduct an in-depth exploration of how these conflicts can be correctly identified and beneficially used for explaining these models. Our analysis consists of two case studies.
	
	In the first case study, we propose a new feature attribution method for ANNs, \emph{Conflict-Aware Feature-wise Explanations (CAFE)}, which is specifically designed to provide principled and accurate handling of conflicts while also tracing the effects of model bias terms. Through our theoretical analysis, we prove that CAFE satisfies a novel \emph{local conflict awareness} property, as well as several other desirable properties from the literature, such as linearity and completeness \citep{sundararajan-integrated-gradients}. In our experiments, we demonstrate that CAFE produces the most accurate scores on synthetic models with conflicting features and achieves competitive performance in explanation faithfulness when evaluated on MLPs and FT-Transformers \citep{gorishniy-revisiting-ft-transformer} trained on seven tabular datasets, including four models from the OpenXAI benchmark \citep{agarwal-openxai}.
	
	In our second case study, we conduct an early investigation aiming to explore the role of conflicts in out-of-distribution (OOD) scenarios. Through our experiments on tabular and image data, we find that the magnitude of the model conflicts varies depending on the type of distributional shift experienced by the model. Our results show the potential utility of conflicts for gaining better understanding of the sources and mechanisms of these shifts.
	
	To summarise, our paper makes the following contributions:
	\begin{enumerate}
		\item We develop a rigorous theory of conflicts between (both external and internal) features in ANNs.
		\item We develop a novel feature attribution method, CAFE, designed to accurately surface conflicts. We prove that CAFE satisfies several desirable properties and achieves competitive performance in standard evaluation metrics.
		\item To demonstrate the potential utility of surfacing conflicts for understanding machine learning phenomena, we explore the connections between the magnitude of these conflicts and different kinds of distributional shift.
	\end{enumerate}
	
	\section{Related Work}
	\label{sec:related-work}
	\textbf{Conflicts in Neural Models.} Several kinds of conflict have been previously explored in the context of machine learning models.  \citet{ledesma-on-removing-conflicts} have considered label conflicts between similar input instances, devising an algorithm for removing the conflicting samples with the aim of optimising model performance. The ConFIG method~\citep{config-conflict-free-training} aims to eliminate conflicts arising from the use of multiple objective functions during training. 
	Most relevant to our work, the DeepLIFT RevealCancel method \citep{shrikumar-deeplift} previously considered canceling features in the context of feature attributions, as discussed below. To the best of our knowledge, no works have previously focused on formalising the notion of conflicts between features.
	
	\textbf{Feature Attribution Methods.} Feature attribution methods quantify the importance of each input feature with respect to the output of a machine learning model. Amongst these methods, simple gradients \citep{simonyan-vanillagrad} or gradients multiplied with the inputs \citep{shrikumar-old-deeplift-gxi} capture ANNs' behavior in a small vicinity around the input, which may not represent the overall behavior when the models are irregular or saturated functions. As an alternative to raw gradients, several enhanced gradient-based attribution methods have been proposed, including LRP \citep{bach-on-pixel-wise-lrp, montavon-lrp-overview}, DeepLIFT Rescale \citep{shrikumar-deeplift} and Integrated Gradients \citep{sundararajan-integrated-gradients}. These methods may not reflect the effects of conflicting features, as illustrated in Figure~\ref{fig:example-cancelation-heart-attack}. DeepLIFT and Integrated Gradients compute explanations with respect to a specific \emph{reference input}, as does our proposed CAFE method. Perturbation methods, such as LIME \citep{ribeiro-lime}, Shapley Value Sampling \citep{strumbelj-shapley-value-sampling} and SHAP \citep{lundberg-shap}, estimate feature importance by modifying input instances and observing changes in model outputs. These methods, however, often require multiple model evaluations, making them computationally demanding.
	
	Two attribution methods are particularly relevant to our goals of accounting for conflicts and biases. DeepLIFT RevealCancel \citep{shrikumar-deeplift} uses an approximation of Shapley values for partially surfacing canceling features, but still underestimates their effects (see example in Section \ref{apd:real-world-conflict} in the Appendix), does not satisfy the formal notion of local conflict awareness, disregards model biases and has not been generalised to more complex models, such as transformers. Bias Back-propagation (BBp) \citep{wang-bias-matters} attributes the effects of biases to input features, differing from CAFE, which computes separate attribution scores for input features and biases, disentangling the contributions of the two. Additionally, unlike CAFE, BBp is only applicable to ANNs with piece-wise linear activation functions.
	
	We discuss additional related work on explanations and their evaluation in Appendix \ref{sec:additional-related-work}.
	
	\section{Notation and Preliminaries}
	\label{sec:preliminaries}
	Our aim is to analyze conflicts in a model $\model: \vectdim{\ldim{0}} \rightarrow \vectdim{\ldim{\nlayers}}$ with $\nlayers$ layers. We view $\model$ as composed of layers $\layer{1}, \ldots, \layer{\nlayers}$, where $\layer{1}$ is the \emph{input layer} and $\layer{\nlayers}$ is the \emph{output layer}. When referring to the vector of activation values of the neurons in layer $\layer{\currlayer}$ ($1 \!\leq \!\currlayer \!\leq \!\nlayers$), we use the notation $\va^{(\currlayer)} $. Apart from standard model inputs, we will also consider reference inputs $\refinputvec \in \vectdim{\ldim{0}}$ and activations $\refactvec{\currlayer} \in \vectdim{\ldim{\currlayer}}$, playing the role of domain-specific ``neutral" vectors. Commonly, this will be zero vectors.
	
	In order to simplify some of our definitions, we assume that linear transformations and applications of activation functions are performed by distinct layers.\footnote{For example, if $\layer{1}$ is the input layer of a model, $\layer{2}$ will typically be a linear layer while $\layer{3}$ will be a layer applying an activation function. Note that our example in Section \ref{sec:introduction} did not follow this convention, for simplicity.}
	A \emph{linear layer} $\layer{\currlayer + 1}: \vectdim{\ldim{\currlayer}} \rightarrow \vectdim{\ldim{\currlayer + 1}}$ computes the operation $\actvec{\currlayer + 1} = \weightmat^{(\currlayer + 1) \top} \actvec{\currlayer} + \biasvec{n+1}$ where $\actvec{n} \in \mathbb{R}^{\ldim{\currlayer}}$ is the output of the previous layer, $\actvec{\currlayer+1} \in \vectdim{\ldim{\currlayer + 1}}$ is the layer output, $\weightmat^{(n+1)} \in \matdim{\ldim{\currlayer}}{\ldim{\currlayer + 1}}$ is the weight matrix and $\biasvec{n+1} \in \vectdim{\ldim{\currlayer + 1}}$ is the bias vector. An \emph{activation layer} $\layer{\currlayer+1}: \vectdim{\ldim{\currlayer}} \rightarrow \vectdim{\ldim{\currlayer + 1}}$ computes the operation $\actvec{\currlayer + 1} = \simpleactfun(\actvec{\currlayer})$ for some activation function $\simpleactfun$. The input and output dimensions of an activation layer, $\ldim{\currlayer}$ and $\ldim{\currlayer + 1}$ respectively, are always identical.
	
	In some of our definitions, we will consider modified versions of a particular vector. These will be represented by the notation $\inputvec | \inputelem{\idxone} := v$, which signifies a vector $\inputvec$ with the value at index $\idxone$ assigned to a new value $v$. The assignments in this notation can be chained in a list to denote the modification of values at multiple indices. To describe setting all activations with indices in set $S$ to their reference values, we will use a shorthand $\remove(S) = \left[\actelem{\currlayer - 1}{\idxone} := \refactelem{\currlayer - 1}{\idxone}\right]_{\idxone \in S}$. For the purposes of clipping values at zero so that they are non-negative, we define a function $\clip(\vx) = \max(\vx, 0)$ that computes an element-wise maximum with 0 for any $\vx$ (for example, $\clip([3, -2]) = [3, 0]$). We also define a unary operator $\bot \vx$ signifying that $\vx$ should be treated as a constant when differentiating, similar to the behaviour of \texttt{detach} and \texttt{stop\_gradient} methods in machine learning libraries. Finally, we will use $\odot$ to denote the Hadamard product.
	
	\section{Theory of Conflicts}
	\label{sec:theory-of-conflicts}
	\subsection{Definition of Conflicts}
	In this section, we develop a theoretical framework for reasoning about conflicts in ANNs. Ultimately, our goal is to clearly define when two sets of features are in conflict and quantify the effect each conflict can have on the output of a model. Note that, in the interest of simplicity, we do not explicitly consider model biases in our definitions of conflicts and assume they can be incorporated as additional features feeding into each linear layer. We start by formally defining when two sets of features act against each other and when they are in conflict:
	
	\begin{definition}[Counteractive Features]
		\label{def:counteractive-features}
		Given a model $\model$ with $N$ layers, a composition of its layers $\layer{\currlayer: m} = \layer{m} \circ \ldots \circ \layer{\currlayer}$ where $1 < n \leq m \leq N$, and sets of features $C$ and $C'$, we say that $C$ \emph{counteracts} $C'$  with respect to an activation $\actelem{m}{\idxthree} = \layer{\currlayer:m}(\actvec{\currlayer - 1})_\idxthree$ and a reference activation $\refactvec{\currlayer - 1}$, denoted as $\counter(\model[\currlayer:m]_\idxthree, C, C',\allowbreak \actvec{\currlayer - 1}, \refactvec{\currlayer - 1})$, if:
		\begin{adjustbox}{width=\columnwidth}
			\begin{minipage}{\columnwidth}
				\begin{align*}
					\frac{
						\layer{\currlayer:m}\left(\actvec{\currlayer - 1}\right)_\idxthree
						- \layer{\currlayer:m}\left(\actvec{\currlayer - 1} \big| \remove(C)\right)_\idxthree
					}{
						\layer{\currlayer:m}\left(\actvec{\currlayer - 1} \big| \remove(C)\right)_\idxthree - \layer{\currlayer:m}\left(\actvec{\currlayer - 1} \big| \remove(C \cup C')\right)_\idxthree
					} < 0
				\end{align*}
			\end{minipage}
		\end{adjustbox}
		with $\remove(S)$ setting all the features in $S$ to domain-appropriate reference values.
	\end{definition}
	
	\begin{definition}[Feature Conflicts]
		\label{def:feature-conflicts}
		Given a model $\model$ with $N$ layers, a composition of its layers $\layer{\currlayer: m} = \layer{m} \circ \ldots \circ \layer{\currlayer}$ where $1 < n \leq m \leq N$, and sets of features $C$ and $C'$, we say that $C$ and $C'$ are in \emph{conflict} with respect to an activation $\actelem{m}{\idxthree} = \layer{\currlayer:m}(\actvec{\currlayer - 1})_\idxthree$ and a reference activation $\refactvec{\currlayer - 1}$, denoted as $\conf(\model[\currlayer:m]_\idxthree, C,\allowbreak C',\allowbreak \actvec{\currlayer - 1}, \refactvec{\currlayer - 1})$, if $C$ counteracts $C'$ or $C'$ counteracts $C$ (with respect to the same activations).
	\end{definition}
	
	Intuitively, the definitions say that two sets of features are in conflict if the change in the layers' output caused by the features in the first set can be (at least partially) reversed by the features in the second set. This aligns with the idea that conflicting features should have opposing effects on the output. We illustrate the application of the two definitions on the example below:
	
	\begin{example}[Counteractive Features and Feature Conflicts]
		\label{ex:counteractive-features-conflicts}
		Consider the example from Figure \ref{fig:example-cancelation-heart-attack} with $ \layer{\currlayer:m} = \relu(0.9 \cdot \inputelem{1} - \inputelem{2})$, $\inputvec = [1, 1]$ and $\refinputvec = [0, 0]$. Intuitively, $\{\inputelem{2}\}$ counteracts $\{\inputelem{1}\}$, as the output would be $0.9$ in the absence of $\inputelem{2}$ (note that the opposite is not the case, as setting $\inputelem{1}$ to $0$ would not affect the output). If we substitute $C = \{\inputelem{2}\}$ and $C' = \{\inputelem{1}\}$ into Definition \ref{def:counteractive-features}, we indeed get $\frac{L^{(n:m)}([1,1])-L^{(n:m)}([1,0])}{L^{(n:m)}([1,0])-L^{(n:m)}([0,0])}=\frac{-0.9}{0.9}=-1<0$, meaning that $\{\inputelem{2}\}$ counteracts $\{\inputelem{1}\}$ and the two sets of features are in conflict.
	\end{example}
	
	We will consider three important instantiations of the above definitions: 
	\begin{itemize}
		\item When $\currlayer = 2$, $m = \nlayers$, and $\layer{\currlayer:m} = \model$\footnote{This is because \layer{1} is the input layer, which does not perform any operation.}, Definitions~\ref{def:counteractive-features} and \ref{def:feature-conflicts} describe \emph{input-output}  counteractive features and conflicts, respectively,  with respect to model output $\actelem{\nlayers}{\idxthree} = \model(\inputvec)_\idxthree$ and a reference input $\refinputvec = \refactvec{\currlayer - 1}$.
		\item When $m = n + 1$, $\layer{\currlayer}$ is a linear layer and $\layer{m} = \layer{n + 1}$ is an activation layer,  Definitions~\ref{def:counteractive-features} and \ref{def:feature-conflicts} describe \emph{local internal} counteractive features and conflicts (i.e., phenomena localized in a set of individual neurons), respectively, with respect to activation $\actelem{n+1}{\idxthree} = \layer{n+1}(\layer{n}(\actvec{\currlayer - 1}))_\idxthree$ and a reference activation $\refactvec{\currlayer - 1}$.
		\item Finally,  Definitions~\ref{def:counteractive-features} and \ref{def:feature-conflicts} describe \emph{minimal}  counteractive features and conflicts, respectively, when there are no subsets $S \subseteq C$, $S' \subseteq C'$, with either $S \subset C$ or $S' \subset C'$, that also satisfy the same definitions.
	\end{itemize}
	
	Apart from identifying when two sets of features are in conflict, it is also useful to be able to quantify the magnitude of this conflict, as conflicts only resulting in little change in the model's output are comparatively less important. To this end, we introduce the notions of counteraction magnitude and feature conflict magnitude:
	
	\begin{definition}[Counteraction Magnitude]
		\label{def:counteraction-magnitude}
		Given a model $\model$ with $N$ layers, a composition of its layers $\layer{\currlayer: m} = \layer{m} \circ \ldots \circ \layer{\currlayer}$ where $1 < n \leq m \leq N$, and sets of features $C$ and $C'$, we can define the \emph{counteraction magnitude} of $C$ against $C'$ with respect to an activation $\actelem{m}{\idxthree} = \layer{\currlayer:m}(\actvec{\currlayer - 1})_\idxthree$ and a reference activation $\refactvec{\currlayer - 1}$ as:
		\begin{adjustbox}{width=\columnwidth}
			\begin{minipage}{\columnwidth}
				\begin{align*}
					&\countermag(\model[\currlayer:m]_\idxthree, C, C', \actvec{\currlayer - 1}, \refactvec{\currlayer - 1}) \\
					&= \min\biggl(
					\left|\layer{\currlayer:m}\left(\actvec{\currlayer - 1}\right)_\idxthree - \layer{\currlayer:m}\left(\actvec{\currlayer - 1} \big| \remove(C)\right)_\idxthree\right|, \\
					&\phantom{= \min\biggl(:}\left|\layer{\currlayer:m}\left(\actvec{\currlayer - 1} \big| \remove(C)\right)_\idxthree - \layer{\currlayer:m}\left(\actvec{\currlayer - 1} \big| \remove(C \cup C')\right)_\idxthree\right|
					\biggr)
				\end{align*}
			\end{minipage}
		\end{adjustbox}
		if $\counter(\model[\currlayer:m]_\idxthree, C, C', \actvec{\currlayer - 1}, \refactvec{\currlayer - 1})$ and $0$ otherwise, with $\remove$ as in Definition $\ref{def:counteractive-features}$.
	\end{definition}
	
	\begin{definition}[Feature Conflict Magnitude]
		\label{def:feature-conflict-magnitude}
		Given a model $\model$ with $N$ layers, a composition of its layers $\layer{\currlayer: m} = \layer{m} \circ \ldots \circ \layer{\currlayer}$ where $1 < n \leq m \leq N$, and the feature sets $C$ and $C'$, we can define the magnitude of their conflict with respect to an activation $\actelem{m}{\idxthree} = \layer{\currlayer:m}(\actvec{\currlayer - 1})_\idxthree$ and a reference activation $\refactvec{\currlayer - 1}$ as:
		\begin{align*}
			&\cmag(\model[\currlayer:m]_\idxthree, C, C', \actvec{\currlayer - 1}, \refactvec{\currlayer - 1}) \\
			&= \max\Bigl(\countermag(\model[\currlayer:m]_\idxthree, C, C', \actvec{\currlayer - 1}, \refactvec{\currlayer - 1}), \\
			&\phantom{= \max\bigl(:}\countermag(\model[\currlayer:m]_\idxthree, C', C, \actvec{\currlayer - 1}, \refactvec{\currlayer - 1})\Bigr)
		\end{align*}
	\end{definition}
	
	Intuitively, the counteraction magnitude quantifies the extent to which features in one set can reverse the effects of features in another set, while feature conflict magnitude measures the degree to which two sets of conflicting features reverse each other.
	
	\begin{example}[Counteraction Magnitude and Feature Conflict Magnitude]
		Consider our running example of the RELU neuron from Figure \ref{fig:example-cancelation-heart-attack}. For this neuron, we obtain $\cmag(\model[\currlayer:m]_\idxthree, \{\inputelem{2}\}, \{\inputelem{1}\},\allowbreak [1, 1],\allowbreak [0, 0]) = \countermag(\model[\currlayer:m]_\idxthree, \{\inputelem{2}\}, \{\inputelem{1}\},\allowbreak [1, 1],\allowbreak [0, 0]) = \min(|0 - 0.9|, |0.9 - 0|) = 0.9$.
	\end{example}
	
	\subsection{Properties of Conflicts}
	Definitions~\ref{def:counteractive-features} and \ref{def:feature-conflicts} and their instances of interest satisfy several properties:
	
	\begin{proposition}
		\label{prop:conflict-symmetry}
		The feature conflict relation is symmetric and anti-reflexive.
	\end{proposition}
	
	\begin{proposition}
		\label{prop:conflict-eq-ref}
		A feature $\actelem{\currlayer}{\idxone}$ for which $\refactelem{\currlayer}{\idxone} = \actelem{\currlayer}{\idxone}$ cannot be an element of any minimal counteractive feature set or any feature set in minimal conflict.
	\end{proposition}
	
	\begin{proposition}
		\label{prop:conflict-linear}
		For a linear model $\model(\inputvec) = \sum^{\ldim{\inputvec}}_\idxone w_\idxone \inputelem{\idxone} + b$ and an input $\inputvec$, the feature sets $C$ and $C'$ are in a minimal input-output conflict with respect to $\refinputvec$ iff:
		\begin{enumerate}
			\item $C = \{\idxone\}, C' = \{\idxtwo\}$ for some $\idxone$, $\idxtwo$, and
			\item $w_\idxone (\inputelem{\idxone} - \refinputelem{\idxone}) < 0 < w_\idxtwo (\inputelem{\idxtwo} - \refinputelem{\idxtwo})$ or $w_\idxone (\inputelem{\idxone} - \refinputelem{\idxone}) > 0 > w_\idxtwo (\inputelem{\idxtwo} - \refinputelem{\idxtwo})$
		\end{enumerate}
	\end{proposition}
	
	Additional properties applicable to ReLU neurons and the proofs of the above propositions are provided in Appendix \ref{sec:conflict-properties-proofs}. While similar (albeit more complex) properties could also be formulated for other types of ANNs, such as those with GELU activations, we defer the investigation of these variations to future work.
	
	\section{Implications of Conflicts for Explainability}
	Since conflict magnitudes directly reflect the potential impact of these conflicts on model's outputs (or outputs of its layers), a natural question to consider is whether feature attribution methods for ANNs reliably capture these conflicts. Failure to do so may mislead users about the true effects of the features and obscure the disagreements between them. To investigate this, we introduce the notions of underreported and hidden conflicts:
	
	\begin{definition}[Underreported / Hidden Conflicts]
		\label{def:underreported-hidden-conflict}
		Consider a model $\model$ with $N$ layers, a composition of its layers $\layer{\currlayer: m} = \layer{m} \circ \ldots \circ \layer{\currlayer}$ where $1 < n \leq m \leq N$, an activation $\actelem{m}{\idxthree} = \layer{\currlayer:m}(\actvec{\currlayer - 1})_\idxthree$, a reference activation $\refactvec{\currlayer - 1}$, a feature set $C'$ and a set $S^C$ containing all sets $C$ such that $C$ counteracts $C'$ in a minimal conflict with respect to $\actelem{m}{\idxthree}$ and $\refactvec{\currlayer - 1}$. Given some feature attribution method $\expmethod$ and its attribution scores $\scoresvec$ for $\layer{\currlayer:m}(\actvec{\currlayer - 1})_\idxthree$ computed with respect to a reference input $\refactvec{\currlayer - 1}$ (if used by the given attribution method), we say that the conflicts involving $C'$ are \emph{underreported} by $\expmethod$ if either $|\sum_{\idxtwo \in C'}s_{\idxtwo}| < \argmax_{C \in S^C} \cmag(\model[\currlayer:m]_\idxthree, C, C', \actvec{\currlayer - 1}, \refactvec{\currlayer - 1})$ or in case there $\exists C \in S^C, \sgn(\sum_{\idxone \in C}s_{\idxone}) \neq -\sgn(\sum_{\idxtwo \in C'}s_{\idxtwo})$. Similarly, we say that the conflicts involving $C'$ are \emph{hidden} if either $|\sum_{\idxtwo \in C'}s_{\idxtwo}| = 0$ or $\exists C \in S^C, \sgn(\sum_{\idxone \in C}s_{\idxone}) \neq -\sgn(\sum_{\idxtwo \in C'}s_{\idxtwo})$.
	\end{definition}
	
	Intuitively, conflicts are underreported if feature attribution scores fail to fully reflect the extent to which a feature set $C'$ could affect the output of the model if it was not counteracted by one or more sets of conflicting features. A conflict is hidden if attribution scores fail to assign any overall importance to a feature set counteracted by other features or if they misleadingly suggest that two feature sets are in agreement when they are actually in conflict.
	
	Unfortunately, it would be unrealistic to expect feature attribution methods to perfectly capture all the input-output conflicts, as this can often be infeasible to do with linear attribution scores. However, in certain cases, these methods may at least be able to capture local conflicts at the level of individual neurons, making these conflicts more likely to be correctly reflected in the final attribution scores. We formalize this property in the following definition:
	
	\begin{definition}[Local Conflict Awareness]
		\label{def:local-conflict-awareness}
		We say that a feature attribution method $\expmethod$ is locally conflict-aware on a certain class of models if it does not underreport or hide minimal local internal conflicts for any model in this class.
	\end{definition}
	
	Despite the desirability of this property, gradient-based attribution methods fail to satisfy it for ANNs with both ReLU and GELU activations, as exemplified in Figure~\ref{fig:example-cancelation-heart-attack}. While perturbation-based explanation techniques might better capture conflicts due to their extensive sampling of model outputs, they are considerably more computationally expensive due to their dependence on a large number of model evaluations. In the following section, we will draw inspiration from the above formalization of conflicts to develop a novel attribution method that would be more capable in reflecting conflicts while still being computable in a single forward-backward model pass.
	
	\section{Explaining Models with Conflict-Aware Explanations}
	\label{sec:explaining-nns-with-cafe}
	CAFE aims to quantify how much each input feature contributes to the model's output and (optionally, controlled by a \emph{conflict sensitivity constant} $c$, as we will see later) the feature contributions that got canceled due to conflicts. Thus, unlike other methods, CAFE returns two separate scores for each feature -- capturing its overall positive and negative effects. This is crucial for uncovering conflicts between features.
	In addition to scores for the inputs, CAFE also returns aggregated scores for the bias terms, indicating how much they affect the output compared to the input features. This may highlight cases where predictions are primarily driven by biases rather than the input.
	
	Formally, CAFE computes \emph{positive feature attribution scores} $\scoresvec^\poss \in \vectdim{\ldim{0}}$, \emph{negative feature attribution scores} $\scoresvec^\negs \in \vectdim{\ldim{0}}$, a \emph{positive bias attribution score} $\scorebias^\poss \in \mathbb{R}$ and a \emph{negative bias attribution score} $\scorebias^\negs \in \mathbb{R}$ for $\model(\inputvec)$, given some \emph{input} $\inputvec \in \vectdim{\ldim{0}}$, a \emph{reference input} $\refinputvec \in \vectdim{\ldim{0}}$ and a specified \emph{target neuron} identified by an index $\targetneuron$ to the final layer $\layer{\nlayers}$ of $\model$ (or a selected hidden layer when explaining intermediate representations). When dealing with classification models, we disregard the final sigmoid/softmax layer, as this has been argued to result in more intuitive model explanations \citep{shrikumar-deeplift}. Note that the values of all the scores are non-negative, as they only capture the magnitude of the corresponding positive/negative contributions (see Figure \ref{tab:example-explosion-gelu-scores} for an example).
	
	CAFE scores are computed using a modified forward-backward pass through the explained model. In the forward pass, CAFE computes the \emph{total positive and negative scores} at each layer $\scoresvecn{\currlayer}{\poss} \in \vectdim{\ldim{\currlayer}}, \scoresvecn{\currlayer}{\negs} \in \vectdim{\ldim{\currlayer}}$ where $\currlayer$ indicates the layer number. The backward pass then distributes the scores from $\scoresvecnelem{\nlayers}{\poss}{\targetneuron}$ and $\scoresvecnelem{\nlayers}{\negs}{\targetneuron}$ among the input features and biases to compute the final attributions. For an overall picture, we first introduce the backward pass procedure for computing the final scores before giving the forward rules.
	
	\subsection{Backward Pass}
	The final CAFE \emph{attribution scores} can be computed as follows:
	
	\begin{definition}[Attribution Scores]
		\label{def:backward-attribution-scores}
		Given $\vfoundation = \inputvec - \refinputvec$, the final layer forward scores $\scoresvecn{\nlayers}{\poss}$ and $\scoresvecn{\nlayers}{\negs}$ and the target index $\targetneuron$, the attribution scores can be computed as:
		{\allowdisplaybreaks
			\begin{gather*}
				\scoresvec^\poss = \frac{\partial \scoresvecnelem{\nlayers}{\poss}{\targetneuron}}{\partial \vfoundation} \odot \vfoundation \quad
				\scoresvec^\negs = \frac{\partial \scoresvecnelem{\nlayers}{\negs}{\targetneuron}}{\partial \vfoundation} \odot \vfoundation \quad \scoresvec^\combs = \scoresvec^\poss - \scoresvec^\negs \\
				\scorebias^\poss = \sum^{\nlayers}_{\currlayer} \sum^{\ldim{\currlayer}}_{\idxone} \left[ \frac{\partial \scoresvecnelem{\nlayers}{\poss}{\targetneuron}}{\partial \biasvec{\currlayer}} \odot \biasvec{\currlayer} \right]_i \\
				\scorebias^\negs = \sum^{\nlayers}_{n} \sum^{\ldim{\currlayer}}_{\idxone} \left[ \frac{\partial \scoresvecnelem{\nlayers}{\negs}{\targetneuron}}{\partial \biasvec{\currlayer}} \odot \biasvec{\currlayer} \right]_i \quad \scorebias^\combs = \scorebias^\poss - \scorebias^\negs \\
				\sconf = \min \left( \sum^{\ldim{0}}_{\idxone} \scoresvecelem{\poss}{\idxone}, \sum^{\ldim{0}}_{\idxone} \scoresvecelem{\negs}{\idxone} \right)
			\end{gather*}
		}
	\end{definition}
	
	In addition to the positive and negative scores for the features and biases, we also define the combined scores $\scoresvec^\combs$ and $\scorebias^\combs$, which reflect the overall estimated effect of the input features or biases on the output. The $\scoresvec^\combs$ scores are analogical to the standard attribution scores returned by other methods, and we use them when comparing CAFE against baselines. These combined scores provide a higher-level explanation of the model behaviour and abstract away some of its implementation details (such as the dual positive and negative effects of a single feature), while still highlighting feature conflicts. We also define the conflict score $\sconf$, a notion unique to CAFE, which quantifies the degree of conflict between positively and negatively contributing features. Intuitively, when the attribution scores are overwhelmingly negative or positive, the $\sconf$ value will be low, but if there are both strong positive and strong negative attributions, the $\sconf$ value will be larger. In our case study in Section \ref{sec:synthetic-details}, we demonstrate that this conflict score exhibits intuitive and potentially useful properties under distributional shift.
	
	\subsection{Input Layer Rule}
	This rule states that the input layer scores are simply the absolute differences between the input features and the corresponding reference input features, as captured below:
	
	\begin{definition}[Input Layer Scores]
		\label{def:input-layer-rule}
		The input layer $\layer{1}$ scores for input $\inputvec$ and reference input $\refinputvec$ are %defined 
		as follows:
		\begin{align*}
			\vfoundation = \inputvec - \refinputvec \quad
			\scoresvecn{1}{\poss} = \clip(\vfoundation) \quad
			\scoresvecn{1}{\negs} = \clip(-\vfoundation)
		\end{align*}
	\end{definition}
	
	\subsection{Linear Layer Rule}
	The propagation of scores through a linear layer is similar to performing a standard forward pass — we simply multiply the attribution scores by the weight matrix and add the bias. However, to distinguish between the positive and negative contributions at each neuron, we separately consider the positive and negative weights and bias terms (e.g., previously negative scores multiplied by negative weights become positive):
	
	\begin{definition}[Linear Layer Scores]
		\label{def:linear-layer-rule}
		The scores for a linear layer $\layer{\currlayer + 1}$ are:
		\begin{align*}
			\scoresvecn{\currlayer + 1}{\poss} &= \scoresvecn{\currlayer}{\poss} \clip(\weightmat^{(\currlayer + 1)}) \\ &\phantom{= }
			+ \scoresvecn{\currlayer}{\negs} \clip(-\weightmat^{(\currlayer + 1)}) + \clip(\biasvec{\currlayer + 1}) \displaybreak[0] \\
			\scoresvecn{\currlayer + 1}{\negs} &= \scoresvecn{\currlayer}{\poss} \clip(-\weightmat^{(\currlayer + 1)}) \\ &\phantom{= }
			+ \scoresvecn{\currlayer}{\negs} \clip(\weightmat^{(\currlayer + 1)}) + \clip(-\biasvec{\currlayer + 1})
		\end{align*}
	\end{definition}
	
	\subsection{Activation Rule}
	\label{sec:cafe-activation-rule}
	
	\begin{figure*}[!ht]
		\centering
		\begin{subfigure}[m]{0.49\textwidth}
			\centering
			\includegraphics[width=1.0\textwidth]{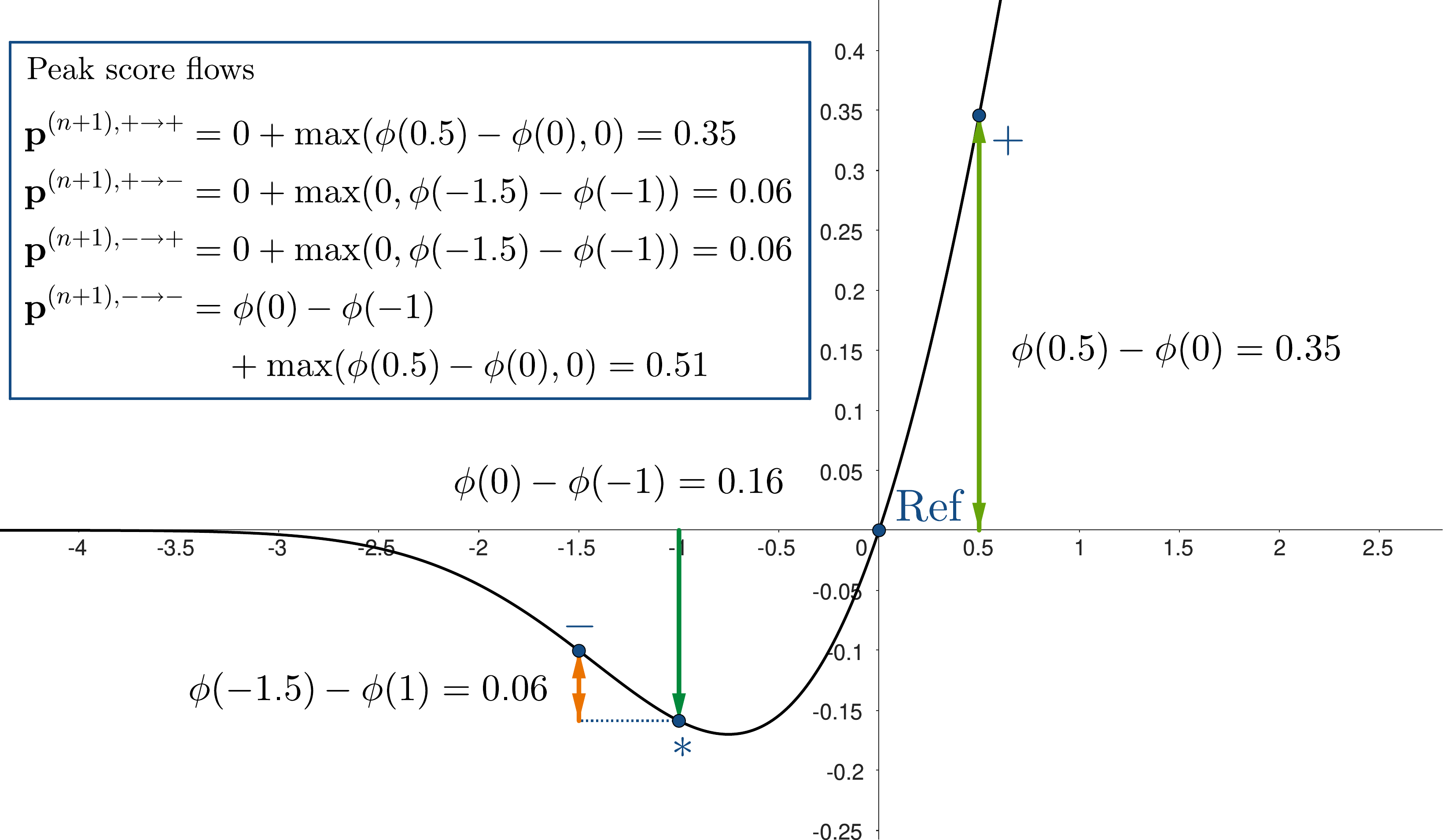}
			\caption{Illustration of peak score flows, $\textbf{p}^{(\currlayer + 1)}$}
			\label{fig:gelu-peaks-detail}
		\end{subfigure}
		\begin{subfigure}[m]{0.49\textwidth}
			\centering
			\includegraphics[width=1.0\textwidth]{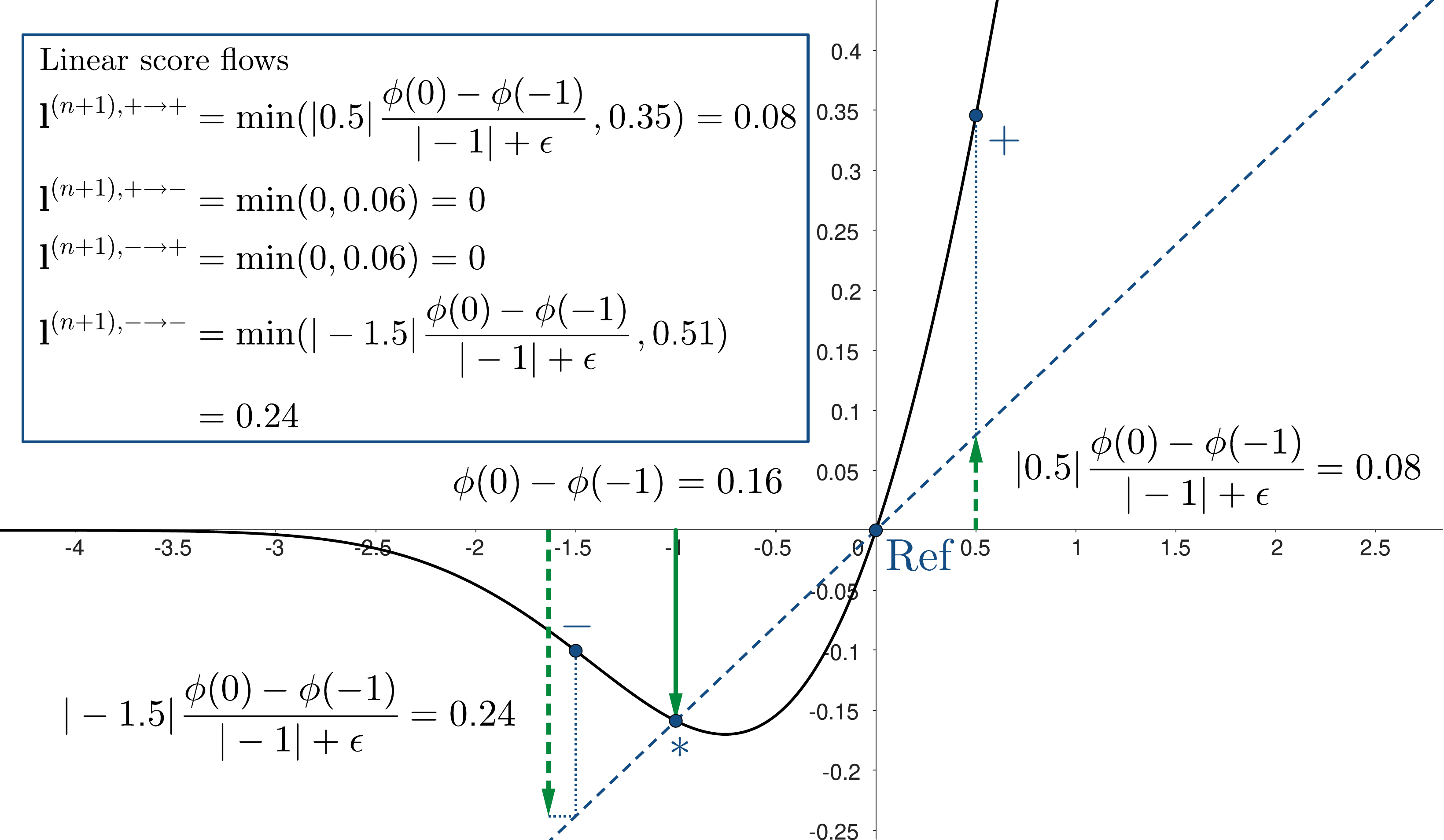}
			\caption{Illustration of linear score flows, $\textbf{l}^{(\currlayer + 1) }$}
			\label{fig:gelu-linear-detail}
		\end{subfigure}
		
		\caption{
			%Visualization illustrating 
			Illustration of the computation of score flows for a GELU unit with reference activation $\refactvec{\currlayer} = 0$, positive inputs $\oinputs{\currlayer + 1}{\poss} = 0.5$, negative inputs $\oinputs{\currlayer + 1}{\negs} = -1.5$ and combined inputs $\oinputs{\currlayer + 1}{\combs} = -1$. See examples \ref{ex:peak-score-flows} and \ref{ex:linear-score-flows} for details.
		}
		\label{fig:example-gelu-detail}
	\end{figure*}
	
	\begin{figure*}[!ht]
		\begin{minipage}{\textwidth}
			\begin{definition}[Peak Score Flows from Positive Inputs to Positive Outputs]
				\label{def:peak-effects}
				The peak score flows from positive inputs to positive outputs at the activation layer $\layer{\currlayer + 1}$ are defined as follows:
				\normalfont
				\begin{align*}
					\vpeak{\currlayer + 1}{\poss}{\poss} &= \indvector{\currlayer}{\poss} \odot \clip(\simpleactfun(\oinputs{\currlayer + 1}{\combs}) - \simpleactfun(\refactvec{\currlayer})) +\max(\clip(\simpleactfun(\oinputs{\currlayer + 1}{\poss}) - \simpleactfun(\indvector{\currlayer}{\poss} \odot \oinputs{\currlayer + 1}{\combs} + \indvector{\currlayer}{\negs} \odot \refactvec{\currlayer})), \\
					&\phantom{=ssss}\clip(\simpleactfun(\indvector{\currlayer}{\poss} \odot \refactvec{\currlayer} + \indvector{\currlayer}{\negs} \odot \oinputs{\currlayer + 1}{\combs}) - \simpleactfun(\oinputs{\currlayer + 1}{\negs})))
				\end{align*}
			\end{definition}
		\end{minipage}
	\end{figure*}
	
	\begin{figure*}[!ht]
		\begin{minipage}{\textwidth}
			\begin{definition}[Linear Score Flows from Positive Inputs to Positive Outputs]
				\label{def:linear-effects}
				The linear score flows from positive inputs to positive outputs at the activation layer $\layer{\currlayer + 1}$ are defined as follows:
				\normalfont
				\begin{align*}
					\vlinear{\currlayer + 1}{\poss}{\poss} &= \min \left( \scoresvecn{\currlayer}{\poss} \odot \frac{\clip \left( \text{sgn}(\scoresvecn{\currlayer}{\combs}) \odot (\phi(\oinputs{\currlayer + 1}{\combs}) - \phi(\refactvec{\currlayer})) \right)}{|\scoresvecn{\currlayer}{\combs}| + \epsilon}, \vpeak{\currlayer + 1}{\poss}{\poss} \right)
				\end{align*}
				where sgn is the element-wise sign function and $\epsilon$ is a small positive stabiliser enforcing the behaviour $\frac{0}{0} = 0$.
			\end{definition}
		\end{minipage}
	\end{figure*}
	
	Defining the rule for propagating scores through non-linear activations is a considerably greater challenge, as the effects of the non-linearities cannot be precisely captured by linear scores, forcing us to approximate. Ultimately, our goal is to quantify how much the positive/negative inputs to the activation layer affected its positive/negative outputs. We do so by computing the so-called \emph{score flows}. To give users a fine-grained control over the handling of conflicting features, we define two variants of score flows, which can be mixed together as desired by appropriately choosing the value of the conflict sensitivity constant $\cancelconst{\currlayer + 1}$.
	
	Before we present further rules, let us introduce a few more preliminaries. In the following definitions, we will consider a version of $\model$, denoted  $\model_\reference$, with all biases set to $0$. We will refer to $\model_\reference$'s activation values at the layer $\layer{\currlayer}$ when applied to $\refinputvec$ as $\refactvec{\currlayer}$. It will also be useful to consider the combined positive and negative input scores $\scoresvecn{\currlayer}{\combs} = \scoresvecn{\currlayer}{\poss} - \scoresvecn{\currlayer}{\negs}$, the overall positive inputs $\oinputs{\currlayer + 1}{\poss} = \refactvec{\currlayer} + \scoresvecn{\currlayer}{\poss}$, the overall negative inputs $\oinputs{\currlayer + 1}{\negs} = \refactvec{\currlayer} - \scoresvecn{\currlayer}{\negs}$ and the combined inputs $\oinputs{\currlayer + 1}{\combs} = \refactvec{\currlayer} + \scoresvecn{\currlayer}{\combs}$. Finally, we will use $\indvector{\currlayer}{\poss} = \mathbb{1}[\scoresvecn{\currlayer}{\combs} \geq 0]$ and $\indvector{\currlayer}{\negs} = \mathbb{1}[\scoresvecn{\currlayer}{\combs} < 0]$ as indicator vectors identifying neurons with non-negative and negative combined input scores, respectively.
	
	\textbf{Peak Score Flows.} These quantities aim to capture the impact of the positive/negative inputs on the positive/negative outputs of an activation layer while also incorporating the canceled effects of conflicting features. To this end, they are based on changes in the activation layer outputs as a result of the combined inputs ($\oinputs{\currlayer +1}{\combs}$) as well as the changes in the outputs when the positive ($\oinputs{\currlayer +1}{\poss}$) and negative ($\oinputs{\currlayer +1}{\negs}$) inputs are considered separately. The formula for computing the \emph{peak score flows from positive inputs to positive outputs} is given in Definition \ref{def:peak-effects}.
	
	The formulas for other combinations are deferred to Definition \ref{def:peak-effects-full} in Appendix \ref{apd:score-flow-definitions} due to a lack of space. The definition is directly informed by the requirements to (i) adjust the intervals over which the differences in outputs are computed depending on whether the combined scores are positive or negative, (ii) capture canceled effects of conflicting features, (iii) maintain the completeness property, and (iv) handle arbitrary and potentially non-monotonic activation functions. While these practical considerations make the peak score flows definition more complex compared to our abstract definitions of conflicts, it relies on the same general principle. In particular, the peak score flows can be seen as locally considering the effects of the potentially conflicting sets of positive inputs (grouped into $\oinputs{\currlayer +1}{\poss}$) and negative inputs (grouped into $\oinputs{\currlayer +1}{\negs}$). For efficiency purposes, the rule only considers groups of positive and negative features rather than arbitrary feature sets.
	
	\begin{example}[Peak Score Flows]
		\label{ex:peak-score-flows}
		The application of the peak score flows definition is illustrated in Figure \ref{fig:gelu-peaks-detail}. The figure captures the computation of peak score flows for a single GELU neuron with reference activation $\refactvec{\currlayer} = 0$, positive inputs $\oinputs{\currlayer + 1}{\poss} = 0.5$, negative inputs $\oinputs{\currlayer + 1}{\negs} = -1.5$ and combined inputs $\oinputs{\currlayer + 1}{\combs} = 0.5 - 1.5 = -1$. We can observe that peak score flows incorporate not only the change in the activation function output as a result of the combined positive and negative inputs (i.e., the decrease by $0.16$), but also the canceled effects of the positive inputs (which could cause increase by $0.35$) and the negative inputs (which could cause increase by $0.06$).
	\end{example}
	Note that the effects of conflicting features need to be added to the peak score flows for both opposite signs in order to maintain the completeness property. The maximum operator is used for aggregating the effects of conflicting features along the same-signed slopes of the activation function, as these effects can never manifest at the same time.
	
	\textbf{Linear Score Flows.} Differently from the peak score flows, \emph{linear score flows} only consider the uncanceled, currently observable effects from the inputs, potentially ignoring the effects of conflicting features. The formula for computing \emph{linear score flows from positive inputs to positive outputs} is given in Definition \ref{def:linear-effects}, with the other variants detailed in Definition \ref{def:linear-effects-full}. 
	
	Linear flows estimate the effects of activation layer inputs on the outputs by constructing a linear approximation of an activation function. This approach is similar in spirit to other gradient-based methods utilising a reference input \citep{sundararajan-integrated-gradients, shrikumar-deeplift}. However, in the presence of unstable or shattered model gradients, this approach may result in unreasonably high attribution scores. To prevent this issue, the linear score flow values are capped by the corresponding peak score flows, which already incorporate the maximum possible changes in the activation layer outputs for extreme values.
	
	\begin{example}[Linear Score Flows]
		\label{ex:linear-score-flows}
		Linear score flows are illustrated in Figure \ref{fig:gelu-linear-detail}, considering the same GELU unit as in Example \ref{ex:peak-score-flows}. We can observe that the linear score flows are directly based on a linear approximation of the activation function with the reference (Ref) and current (\combs{}) activation values as endpoints.
	\end{example}
	
	\textbf{Score Multipliers.} To proceed with propagating the scores through the activation layer, the peak score flows and linear score flows are combined in \emph{score multipliers}. The relative weights of the two components are customisable by the user-provided conflict sensitivity constant $\cancelconst{\currlayer}$. This enables users to decide how much to reflect the effects of conflicting features in the final attribution scores. Values of $\cancelconst{\currlayer}$ closer to $0$ typically result in more focused attribution scores while the values closer to $1$ encourage greater sensitivity to conflicts between the individual features. The multipliers are also normalised by the corresponding positive/negative scores from the previous layer. This ensures that the total score flows reflected in the peak and linear flows are redistributed between the individual features proportionally to their share on these scores. The formal definition is given below:
	
	\begin{definition}[Score Multipliers]
		\label{def:activation-multipliers}
		The score multipliers for the activation layer $\layer{\currlayer}$ are:
		\begin{align*}
			\vmultiplier{\currlayer}{\bullet}{\circ} = \frac{(1 - \cancelconst{\currlayer}) \vlinear{\currlayer}{\bullet}{\circ} + \cancelconst{\currlayer} \vpeak{\currlayer}{\bullet}{\circ}}{\scoresvecn{\currlayer - 1}{\bullet} + \epsilon}
		\end{align*}
		\noindent  where $\cancelconst{\currlayer}$ is the \emph{conflict sensitivity constant}, $0 \leq \cancelconst{\currlayer} \leq 1$, $\epsilon$ is a small positive stabiliser enforcing $\frac{0}{0} = 0$, and $\bullet$ and $\circ$ are wildcards for $+$ and $-$ signs.
	\end{definition}
	
	\textbf{Activation Layer Scores.} Finally, given the score multipliers, we can directly compute the \emph{activation layer scores}:
	
	\begin{definition}[Activation Layer Scores]
		\label{def:activation-scores}
		The scores for an activation layer $\layer{\currlayer + 1}$ are:
		\begin{align*}
			\scoresvecn{\currlayer + 1}{\bullet} = &\bot \vmultiplier{\currlayer + 1}{\poss}{\bullet} \odot \scoresvecn{\currlayer}{\poss} \\
			&+ \bot \vmultiplier{\currlayer + 1}{\negs}{\bullet} \odot \scoresvecn{\currlayer}{\negs}
		\end{align*}
		\noindent where $\bullet$ is a wildcard for $+$ and $-$ signs.
	\end{definition}
	
	Note that we treat the score multipliers as constants during differentiation, which effectively linearises the activation layer for the backward pass. 
	
	\subsection{The Choice of Conflict Sensitivity Constant}
	\label{sec:choice-of-c}
	As we detailed above, the behaviour of CAFE when attributing conflicts can be controlled by setting the value of the conflict sensitivity constant $\cancelconst{\currlayer}$. Typically, the choice of $\cancelconst{\currlayer}$ should be guided by the preference regarding the degree to which conflicts should be captured in the final scores. Setting $\cancelconst{\currlayer} = 0$ is appropriate to disregard conflicts, while $\cancelconst{\currlayer} = 1$ maximally factors in the canceled effects of conflicting features, potentially overshadowing the non-canceled feature effects. In the absence of any specific preferences for conflict handling, a value of $\cancelconst{\currlayer} = 0.5$ is a sensible middle ground that identifies conflicts without making them dominate the scores. As we will show in Section \ref{sec:real-data-experiments}, this variant of CAFE also generally achieves the best faithfulness and balanced robustness on real data. Alternatively, the value of $\cancelconst{\currlayer}$ can also be directly tuned to optimize for a particular explanation metric or a set of metrics on a chosen validation dataset.
	
	\section{Evaluation}
	\label{sec:experiments}
	In our evaluation, we take a multi-faceted approach including the consideration of CAFE's theoretical properties, computational %complexity
	efficiency, and experimental evaluation on synthetic and real-data models.\footnote{Our code is available at \url{https://github.com/adamdejl/hidden-conflicts}}

	\subsection{Theoretical Analysis}
	\label{sec:theoretical-evaluation}
	To demonstrate that CAFE satisfies its primary design criterion of being more capable at surfacing feature conflicts, we show that it is locally conflict aware for ReLU models, in line with Definition \ref{def:local-conflict-awareness}.
	
	\begin{theorem}
		\label{theo:cafe-conflict-awareness}
		CAFE with $\cancelconst{} = 1$ computed with respect to a reference input of zero is locally conflict aware on multilayer perceptrons with ReLU activation functions.
	\end{theorem}
	
	We also prove that CAFE satisfies three desirable properties from the literature (which have been adjusted to cater for separating positive and negative scores and producing bias attributions -- see Appendix \ref{sec:cafe-properties-proofs}). \emph{Missingness}, previously considered for SHAP \citep{lundberg-shap}, requires that missing features are always assigned a zero attribution score. \emph{Linearity}, one of the axioms for Integrated Gradients \citep{sundararajan-integrated-gradients}, requires that the attribution scores preserve any linear behaviour, that is, for a model formed as a linear combination of two other models, the attribution scores should be the result of applying the same linear combination to the scores for the two constituent models. Finally, \emph{completeness} requires that the attribution scores exactly account for all the changes to the model output caused by the input features.
	
	\begin{theorem}
		\label{theo:literature}
		CAFE satisfies missingness, linearity and completeness for any choice of conflict sensitivity constant $\cancelconst{\currlayer}$.
	\end{theorem}
	
	Formal definitions of the properties and the associated proofs are provided in Appendix \ref{sec:cafe-properties-proofs}.
	
	The asymptotic time complexity of CAFE is equivalent to the cost of the forward and backward functions associated with the explained model. In our experiments, we found that the runtimes of CAFE are comparable to those of other gradient-based methods and are significantly better compared to methods requiring extensive sampling. The detailed runtimes are reported in Appendix \ref{sec:comp-complexity-runtimes}.
	
	\subsection{Synthetic Data Experiments}
	\label{sec:synthetic-details}
	\textbf{Data with Conflicting Features.} In these experiments, we empirically test the ability of CAFE and various baselines to %correctly 
	identify the effects of conflicting features. To this end, we construct several synthetic datasets using a controlled data generation process, which enables us to establish the expected ``ground-truth" attribution scores. We also experiment with procedurally generated models with known reasoning.
	
	Details and the full results are given in Appendix \ref{apd:synthetic-details}. On a high level, variants of CAFE using larger cancellation sensitivity constants outperformed all the considered baselines. CAFE ($c = 1.0$) achieved the best performance in all experiments. Overall, the results suggest that CAFE variants with higher conflict sensitivity is highly capable of attributing conflicting features, even when compared to computationally expensive sampling methods.
	
	\subsection{Real Data Experiments}
	\label{sec:real-data-experiments}
	\textbf{Datasets and Models.} We used four datasets and pre-trained MLPs from the OpenXAI benchmark \citep{agarwal-openxai} — COMPAS \citep{mattu-machine-bias}, Home Equity Line of Credit (HELOC) \citep{heloc-data}, Adult Income \citep{adult-income} and German Credit \citep{german-credit}. Additionally, we trained further MLP models on the Titanic \citep{titanic-frank} and Covertype \citep{covertype-blackard} datasets, as well as a subset of the MIMIC-IV medical database \citep{johnson-mimic-iv}. We also conducted evaluation on a set of FT-Transformer models trained on the same datasets. For each dataset and model architecture combination, we trained five differently initialised models and averaged the results, except for the experiments using pre-trained models from OpenXAI. Experimental details are reported in Appendix \ref{sec:real-details}.
	
	\begin{table*}[!tb]
		\caption{Infidelity of attribution methods on MLPs trained on various datasets for smaller (S) and larger (L) perturbations. Best results in bold, second-best results are underlined. \faHourglassHalf{} marks methods more than 10-times slower than CAFE on average.}
		\label{tab:real-data-infidelity}
		\footnotesize
		\centering
		\begin{center}
			\begin{tabular}{ c c c c c c c c c c c c c c c }
				\toprule
				\multirow{4}{*}{\textbf{Method}} & \multicolumn{14}{c}{\textbf{Attribution Infidelity ($\downarrow$)}} \\
				\cmidrule(r){2-15}
				& \multicolumn{2}{c}{\textbf{COMPAS}} & \multicolumn{2}{c}{\textbf{HELOC}} & \multicolumn{2}{c}{\textbf{Adult}} & \multicolumn{2}{c}{\textbf{German}} & \multicolumn{2}{c}{\textbf{Titanic}} & \multicolumn{2}{c}{\textbf{MIMIC-IV}} & \multicolumn{2}{c}{\textbf{Covertype}} \\
				\cmidrule(r){2-3} \cmidrule(r){4-5} \cmidrule(r){6-7} \cmidrule(r){8-9} \cmidrule(r){10-11} \cmidrule(r){12-13} \cmidrule(r){14-15}
				& \textbf{S} & \textbf{L} & \textbf{S} & \textbf{L} & \textbf{S} & \textbf{L} & \textbf{S} & \textbf{L} & \textbf{S} & \textbf{L} & \textbf{S} & \textbf{L} & \textbf{S} & \textbf{L} \\
				\midrule
				Gradient $\cdot$ Input & 12.85 & 32.43 & 14.52 & 38.45 & 27.554 & 51.848 & \textbf{0.024} & \textbf{0.052} & 4.93 & 11.85 & 2.33 & 5.04 & 625.75 & 1703.62 \\
				LRP & 12.85 & 32.43 & 14.52 & 38.45 & 27.554 & 51.848 & \textbf{0.024} & \textbf{0.052} & 4.93 & 11.85 & 2.33 & 5.04 & 616.78 & 1682.71 \\
				DeepLIFT Rescale & 12.01 & 30.24 & 13.81 & 36.30 & 27.578 & 51.927 & \textbf{0.024} & \textbf{0.052} & \textbf{4.82} & \underline{11.45} & 2.24 & 4.86 & 616.11 & 1680.12 \\
				GradientSHAP & 12.07 & 30.41 & 14.08 & 36.89 & 27.588 & 51.953 & \textbf{0.024} & \textbf{0.052} & \underline{4.88} & 11.56 & 2.30 & 4.97 & 620.86 & 1688.85 \\
				Integrated Gradients & 12.01 & 30.23 & 13.81 & 36.30 & 27.578 & 51.927 & \textbf{0.024} & \textbf{0.052} & \textbf{4.82} & 11.46 & 2.25 & 4.88 & 618.00 & 1681.93 \\
				SmoothGrad & 28.51 & 68.12 & 16.50 & 41.69 & 28.176 & 52.690 & 0.052 & 0.112 & 6.17 & 13.94 & 3.17 & 6.31 & 639.88 & 1739.94 \\
				KernelSHAP \faHourglassHalf{} & 15.24 & 37.50 & 16.75 & 42.55 & 31.566 & 59.303 & 0.056 & 0.115 & 5.30 & 12.29 & 2.77 & 5.68 & 655.83 & 1779.54 \\
				Shapley Value Sampling \faHourglassHalf{} & \textbf{11.92} & 29.98 & 12.39 & 31.80 & 27.598 & 51.960 & \underline{0.048} & \underline{0.097} & 5.17 & 12.02 & \underline{2.22} & \underline{4.82} & \textbf{577.80} & \textbf{1589.68} \\
				LIME \faHourglassHalf{} & 18.24 & 43.30 & 16.19 & 41.23 & 28.716 & 54.041 & 0.061 & 0.126 & 5.31 & 12.32 & 2.67 & 5.53 & 647.22 & 1759.57 \\
				\midrule
				CAFE ($c = 0.0$) & 12.85 & 32.43 & 14.52 & 38.45 & 27.554 & 51.848 & \textbf{0.024} & \textbf{0.052} & 4.93 & 11.85 & 2.33 & 5.04 & 616.77 & 1682.70 \\
				CAFE ($c = 0.5$) & \underline{11.94} & \underline{29.85} & \textbf{11.89} & \underline{30.23} & \textbf{27.549} & \textbf{51.836} & \textbf{0.024} & \textbf{0.052} & \textbf{4.82} & \textbf{11.29} & \textbf{2.16} & \textbf{4.50} & \underline{605.52} & \underline{1661.20} \\
				CAFE ($c = 1.0$) & 12.06 & \textbf{29.75} & \underline{12.13} & \textbf{29.84} & \underline{27.552} & \underline{51.839} & \textbf{0.024} & \textbf{0.052} & 5.26 & 11.96 & 2.50 & 4.97 & 617.01 & 1687.21 \\
				\bottomrule
			\end{tabular}
		\end{center}
	\end{table*}
	
	\begin{table*}[!tb]
		\caption{Infidelity of attribution methods on FT-Transformers trained on various datasets for smaller (S) and larger (L) perturbations. Best results in bold, second-best results are underlined. \faHourglassHalf{} marks methods more than 10-times slower than CAFE on average.}
		\label{tab:ft-real-data-infidelity}
		\footnotesize
		\centering
		\begin{center}
			\begin{tabular}{ c c c c c c c c c c c c c c c }
				\toprule
				\multirow{4}{*}{\textbf{Method}} & \multicolumn{14}{c}{\textbf{Attribution Infidelity ($\downarrow$)}} \\
				\cmidrule(r){2-15}
				& \multicolumn{2}{c}{\textbf{COMPAS}} & \multicolumn{2}{c}{\textbf{HELOC}} & \multicolumn{2}{c}{\textbf{Adult}} & \multicolumn{2}{c}{\textbf{German}} & \multicolumn{2}{c}{\textbf{Titanic}} & \multicolumn{2}{c}{\textbf{MIMIC-IV}} & \multicolumn{2}{c}{\textbf{Covertype}} \\
				\cmidrule(r){2-3} \cmidrule(r){4-5} \cmidrule(r){6-7} \cmidrule(r){8-9} \cmidrule(r){10-11} \cmidrule(r){12-13} \cmidrule(r){14-15}
				& \textbf{S} & \textbf{L} & \textbf{S} & \textbf{L} & \textbf{S} & \textbf{L} & \textbf{S} & \textbf{L} & \textbf{S} & \textbf{L} & \textbf{S} & \textbf{L} & \textbf{S} & \textbf{L} \\
				\midrule
				Gradient $\cdot$ Input & 1.085 & 1.286 & 0.783 & 0.920 & 5.72 & 6.10 & 0.98 & 1.47 & 4.21 & 7.48 & 6.35 & 9.62 & 44.47 & 64.85 \\
				DeepLIFT Rescale & 1.080 & 1.285 & 0.783 & 0.920 & 5.68 & 6.06 & 0.97 & 1.44 & 4.20 & 7.53 & \underline{5.90} & \underline{8.78} & 44.33 & 64.69 \\
				GradientSHAP & 1.004 & 1.201 & 0.767 & 0.905 & 5.63 & 5.99 & 0.97 & 1.45 & 4.17 & 7.38 & 6.45 & 9.67 & 44.35 & 64.74 \\
				Integrated Gradients & 0.996 & \underline{1.191} & 0.765 & 0.903 & \underline{5.61} & \underline{5.97} & 0.96 & 1.45 & 4.17 & 7.39 & 6.40 & 9.60 & 44.17 & 64.54 \\
				SmoothGrad & 1.693 & 1.871 & 0.932 & 1.033 & 5.85 & 6.23 & 1.05 & 1.53 & 4.18 & 7.34 & 6.66 & 9.96 & 44.56 & 64.72 \\
				Shapley Value Sampling \faHourglassHalf{} & 0.995 & \underline{1.191} & 0.765 & 0.901 & \textbf{5.60} & \textbf{5.96} & 0.98 & 1.45 & 4.32 & 7.71 & 6.16 & 9.22 & \textbf{43.46} & \textbf{63.73} \\
				\midrule
				CAFE ($c = 0.0$) & 1.005 & 1.201 & \textbf{0.760} & \underline{0.899} & 5.66 & 6.04 & \textbf{0.88} & \underline{1.31} & \underline{3.93} & 6.99 & \textbf{5.78} & \textbf{8.65} & \underline{43.87} & 64.12 \\
				CAFE ($c = 0.5$) & \textbf{0.982} & \textbf{1.178} & \textbf{0.760} & \textbf{0.898} & 5.68 & 6.05 & \underline{0.89} & \textbf{1.30} & \textbf{3.89} & \textbf{6.88} & 6.06 & 8.86 & 43.90 & \underline{63.93} \\
				CAFE ($c = 1.0$) & \underline{0.983} & \textbf{1.178} & \underline{0.761} & \textbf{0.898} & 5.70 & 6.07 & 0.91 & 1.34 & \underline{3.93} & \underline{6.92} & 6.25 & 9.14 & 44.00 & 63.98 \\
				\bottomrule
			\end{tabular}
		\end{center}
	\end{table*}
	
	\textbf{Baselines.} Overall, we consider nine baseline feature attribution methods — Gradient $\cdot$ Input \citep{shrikumar-old-deeplift-gxi}, LRP \citep{bach-on-pixel-wise-lrp, montavon-lrp-overview}, DeepLIFT Rescale \citep{shrikumar-deeplift}, GradientSHAP \citep{lundberg-shap}, Integrated Gradients \citep{sundararajan-integrated-gradients}, SmoothGrad \citep{smilkov-smoothgrad}, KernelSHAP \citep{lundberg-shap}, Shapley Value Sampling \citep{strumbelj-shapley-value-sampling} and LIME \citep{ribeiro-lime}. For some of the experiments, we excluded the KernelSHAP and LIME baselines, as they were too computationally expensive, and LRP, as the used implementation did not support FT-Transformers.
	
	\textbf{Evaluation Metrics.} We consider faithfulness (operationalised through infidelity \citep{infidelity-sensitivity-explanations}) as our primary performance indicator on real data, as it amounts to the explanation accurately describing the behaviour of the model. We also consider structural infidelity~\citep{ayoobi-sparx}, \emph{max-sensitivity} \citep{infidelity-sensitivity-explanations} and \emph{complexity} \citep{bhatt-faithfulness-corr} as supplemental metrics. A more detailed justification of our evaluation strategy and the full experiment results are described in Appendix \ref{sec:real-details}.
	
	\textbf{Results --- Infidelity.} The infidelity metric \citep{infidelity-sensitivity-explanations} allows us to evaluate how well attribution scores predict the behaviour of the explained model on significantly perturbed inputs, with lower values indicating better results. We used Captum to compute both baseline explanation scores (using a zero reference input where applicable) and infidelity values, considering joint CAFE scores as for synthetic data. For MLP models (Table \ref{tab:real-data-infidelity}), CAFE ($c = 0.5$) outperformed or matched all comparably fast attribution methods. Remarkably, CAFE ($c = 0.5$) also sometimes outperformed considerably slower and computationally more expensive perturbation-based methods (marked with \faHourglassHalf{} in the result tables). For FT-Transformers (Table \ref{tab:ft-real-data-infidelity}), CAFE ($c = 0.5$) outperformed comparably fast methods on five of the seven datasets. Similarly as for the MLPs, it also often outperformed the computationally more expensive Shapley Value Sampling.
	
	We hypothesize that CAFE's high performance in fidelity may be due to its superior ability to predict the changes in the model output when perturbing features engaged in a conflict. As we will discuss in Section \ref{sec:conflict-prevalence}, a substantial share of internal model neurons is involved in conflicts, suggesting their importance for the mechanisms of neural networks.
	
	In contrast with the synthetic data experiments, CAFE ($c = 0.5$) outperformed CAFE ($c = 1.0$). This is expected, as the conflicts between real data features are likely to be more complex, and scores capturing them more noisy. Method runtimes and examples of CAFE attributions are provided in Appendices \ref{sec:comp-complexity-runtimes} and \ref{sec:cafe-attriubtion-examples}.
	
	\textbf{Results --- Structural Infidelity.} Apart from considering the input-output faithfulness of CAFE, we also evaluated it in terms of structural infidelity \citep{ayoobi-sparx}, considering its ability to provide faithful attributions for a randomly sampled subset of internal model neurons. The results are reported in Table \ref{tab:real-data-struct-infidelity} for MLPs and in Table \ref{tab:ft-real-data-struct-infidelity} for FT-Transformers. Similarly as for the input-output infidelity, CAFE ($c = 0.5$) again achieved the overall best result in terms of relative method ranking on both types of models.
	
	\begin{figure*}[!tb]
		\centering
		\includegraphics[width=0.73\textwidth]{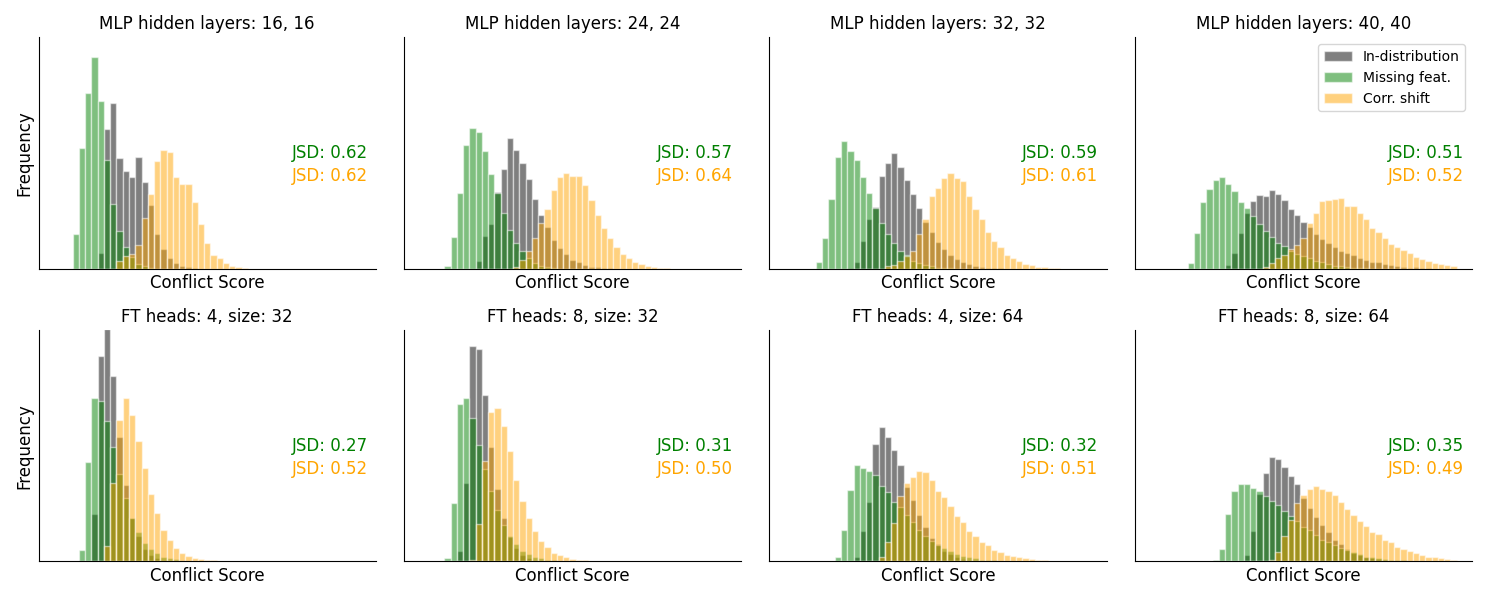}
		\caption{CAFE identifies an increase in conflicts induced by a correlation shift and a decrease in conflicts induced by a missing feature. We report the Jensen-Shannon distance (JSD) between the conflict score ($\sconf$) distributions of the shifted and training datasets.}
		\label{fig:ood}
	\end{figure*}
	
	\begin{figure*}[!tb]
		\centering
		\includegraphics[width=0.73\textwidth]{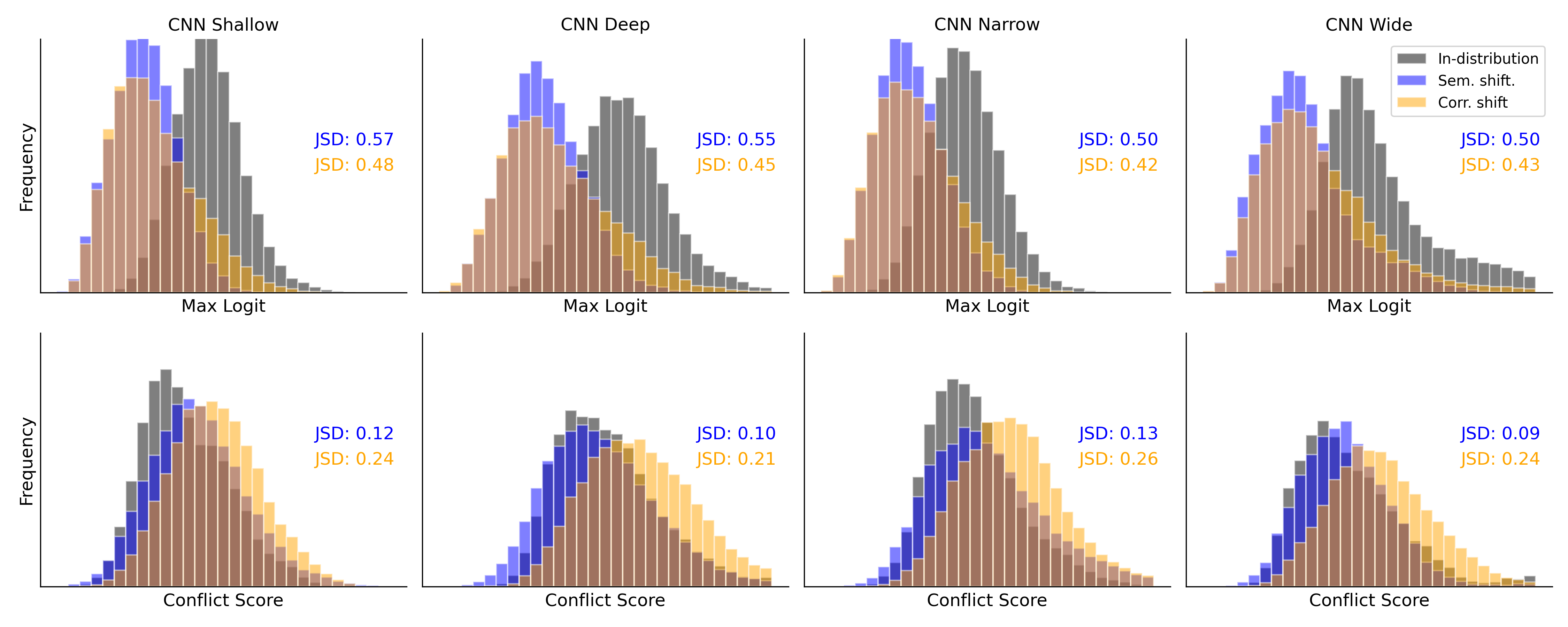}
		\caption{Out-of-distribution shifts are reflected in similarly lower MaxLogits (top). CAFE identifies a greater increase in feature conflicts induced by the correlation shift (bottom).}
		\label{fig:ood_mnist}
	\end{figure*}
	
	\textbf{Results --- Sensitivity.} We also evaluated the robustness of CAFE using the sensitivity metric \citep{infidelity-sensitivity-explanations}, though this metric may be at odds with fidelity. The results are shown in Table \ref{tab:real-data-sensitivity} for the MLPs and in Table \ref{tab:ft-real-data-sensitivity} for the FT-Transformers. CAFE ($c = 1.0$) exhibited the best sensitivity on MLP models, likely due to addressing the attribution score discontinuity issue outlined in Example \ref{ex:attribution-score-discontinuity}, while CAFE ($c = 0.5$) performed roughly similarly to the most robust baselines. For FT-Transformer models, both CAFE ($c = 0.5$) and CAFE ($c = 1.0$) outperformed comparably fast methods on four out of seven datasets and achieved competitive results on the remaining ones. Overall, we argue that CAFE provides balanced performance as measured by infidelity and sensitivity, especially since users can adapt their choice of $\cancelconst{}$ according to their preferred metric.
	
	\subsection{Conflict Prevalence}
	\label{sec:conflict-prevalence}
	To estimate the real-world prevalence and potential importance of conflicts, we computed the average share of inactive neurons with non-zero inputs for each of the models used in our real-data evaluation when applied to samples from its corresponding test set. On average, we find that $49\%$ neurons in MLP models and $30\%$ of neurons in FT-Transformer models are engaged in a conflict. We provide the details in Appendix \ref{apd:conflict-prevalence}. Additionally, to further demonstrate the phenomenon of conflicts, we give an example of a real model circuit engaged in a conflict in Appendix \ref{apd:real-world-conflict}.
	
	\section{Conflicts and Distributional Shifts}
	
	\textbf{Impact on Input Feature Conflicts.} Separately, CAFE's distinct ability to track positive and negative scores for each feature allows us to inspect the model for conflicts, which could potentially serve as an additional diagnostic tool. We provide some intuition on the behaviour of our conflict score $\sconf$ (see Definition \ref{def:backward-attribution-scores}) by synthetically generating training data with two predictive features and simulating a correlation shift and a missing feature to illustrate that the conflict score is sensitive to such shifts. Intuitively, we would expect a model trained on the training distribution to exhibit an increase in the conflict score when evaluated on the correlation-shifted dataset (as the previously correlated variables can now be in conflict) and a decrease in the conflict score when evaluated on the dataset with a missing feature (as the missing feature can no longer cause a conflict). Details on the data generation and the used experimental setup are provided in Appendix~\ref{apd:synthetic-details}. Figure~\ref{fig:ood} reports the distributions of conflict scores over $10000$ samples taken from each dataset and the Jensen-Shannon distances between these distributions. The results confirm our original expectations regarding conflict score behaviour.
	
	\textbf{Impact on Latent Feature Conflicts.}
	Outside of input feature conflicts, CAFE can also be used to understand the interaction and conflicts between latent features. Several works have previously noted that the norm of the latent feature vector % $\Vert z \Vert$
	is smaller for OOD samples (e.g., \citet{vaze2022openset,yu2023,park2023}). OOD-detection is a key task for safety-critical applications. \citet{dietterich2022108931} hypothesise that state-of-the-art OOD-detection methods using the maximum logit (MaxLogit) detect semantic shifts via the absence of familiar features as opposed to conflicting novel features. To test this hypothesis, we train convolutional neural networks (CNN) on a concatenation of MNIST where the top and bottom digits are both correlated with the label. We evaluate it on an OOD dataset with a partial semantic shift, which we induce by replacing the bottom digit with an image from Fashion-MNIST. Under the hypothesis, we would expect lower MaxLogits on the OOD dataset as well as no increase in the number of conflicts between latent features for these samples. We next consider a different OOD setting with a partial correlation shift. We replace the bottom digit with a random digit, thus removing the correlation between the bottom digit and the label. Since the semantics are similar, we expect a smaller decrease in MaxLogits. Since the CNN was trained on a dataset where both bottom and top were equally correlated with the label, we now additionally expect there to be conflicts in what the top and bottom features predict. Details on the data and models are provided in Appendix~\ref{apd:synthetic-details}.
	
	We plot the distribution of MaxLogits (top) and Conflict Scores (bottom) in Figure~\ref{fig:ood_mnist}. We observe that the correlation shift induces a similar decrease in MaxLogits compared to the semantic shift but, in contrast, tends to result in an increase in conflicts. CAFE's ability to capture conflicts could thus help better understand the source of out-of-distribution shifts.
	
	\section{Conclusion}
	We introduced a theoretical framework for reasoning about feature conflicts in ANNs and investigated its implications for their explainability. As part of our case studies, we introduced a novel feature attribution method, CAFE, providing a principled handling of conflicting features, while also separately quantifying the effects of the bias terms. We theoretically and experimentally validated CAFE's ability to surface conflicts, showed that it satisfies several desirable properties from the literature and demonstrated its competitive performance in standard metrics. Finally, we conducted a preliminary study investigating how conflicts surfaced by CAFE relate to different kinds of distributional shifts.
	
	There are several notable limitations of our work. Similar to other feature attribution methods, the attribution scores computed by CAFE are only approximations of the true behaviour of the explained model and may not provide a fully accurate account of the effects of the individual input features (see the discussion in FAQs \ref{apd:faq-considerations}, \ref{apd:faq-no-conflicts} and \ref{apd:faq-overestimates} for more details). Additionally, our case study on OOD is largely exploratory and further experiments would be needed to fully establish the utility of considering conflicts for diagnosing distributional shifts.
	
	There are several promising future directions. In particular, further exploring the properties and roles of conflicts in more diverse neural architectures and additional data modalities could provide valuable insights. User studies are also needed to fully assess conflict-aware feature attribution methods from a human-centric perspective and to determine whether surfacing feature conflicts and bias effects enhances users' understanding of the explained model.
	
	\subsubsection*{Acknowledgements}
	This research was partially supported by ERC under the EU's Horizon 2020 research and innovation programme (grant agreement no. 101020934, ADIX), by J.P. Morgan and the Royal Academy of Engineering under the Research Chairs and Senior Research Fellowships scheme (grant agreement no. RCSRF2021\textbackslash11\textbackslash45), by UKRI through the CDT in AI for Healthcare \url{https://ai4health.io/} (grant no. EP/S023283/1), by NIHR Imperial BRC (grant no. RDB01 79560) and by Brain Tumour Research Centre of Excellence at Imperial.
	
	\bibliography{refs}
	
	%%%%%%%%%%%%%%%%%%%%%%%%%%%%%%%%%%%%%%%%%%%%%%%%%%%%%%%%%%%%%%%%%%%%%%%%%%%%%%%
	%%%%%%%%%%%%%%%%%%%%%%%%%%%%%%%%%%%%%%%%%%%%%%%%%%%%%%%%%%%%%%%%%%%%%%%%%%%%%%%
	% APPENDIX
	%%%%%%%%%%%%%%%%%%%%%%%%%%%%%%%%%%%%%%%%%%%%%%%%%%%%%%%%%%%%%%%%%%%%%%%%%%%%%%%
	%%%%%%%%%%%%%%%%%%%%%%%%%%%%%%%%%%%%%%%%%%%%%%%%%%%%%%%%%%%%%%%%%%%%%%%%%%%%%%%
	\appendix
	\onecolumn
	\section{Appendices Outline}
	This appendix material is organised as follows. The FAQs in Appendix \ref{sec:faq} provide answers to several questions that may be relevant to practitioners and researchers who wish to use or extend CAFE in their work. In Appendix \ref{sec:additional-related-work}, we provide additional details about the related work. Appendix \ref{sec:conflict-properties-proofs} contains the proofs of the introduced properties of feature conflicts. In Appendix \ref{apd:cafe-motivating-examples}, we provide some additional examples motivating the design of CAFE. The full versions of the score flow definitions are stated in Appendix \ref{apd:score-flow-definitions}. A brief glossary of some of the key notions relating to CAFE along with additional notes on these notions is given in Appendix \ref{sec:notes-cafe-notions}. Adaptations required to apply CAFE to FT-Transformer models are described in Appendix \ref{apd:adapting-cafe-ft-transformer}. In Appendix \ref{sec:cafe-properties-proofs}, we provide details on the properties of CAFE as well as the proofs for Theorems \ref{theo:cafe-conflict-awareness} and \ref{theo:literature} from the main text. Appendix \ref{sec:experiment-details} states the details regarding our experiments, while Appendix \ref{sec:cafe-attriubtion-examples} shows several examples of CAFE attribution scores.
	
	\section{FAQ}
	\label{sec:faq}
	In this section, we provide answers to several questions that may be relevant to practitioners and researchers wishing to use or build upon CAFE in their work.
	
	\subsection{What activation functions and model classes are supported by CAFE?}
	In our experiments, we evaluated CAFE on networks using the ReLU and GELU activations, as we consider these to be one of the most common choices for recent neural architectures. Nevertheless, CAFE does not make any particular assumptions regarding the used activation function, and is thus readily applicable to networks using other common activation functions, such as tanh, sigmoid, swish or softplus. In theory, CAFE may fail to produce reasonable attribution scores for highly irregular activation functions with high numbers of local maxima and minima. However, such functions would likely be unsuitable for use in neural networks, so we do not see this to be a practical limitation.
	
	As implemented and used in our work, CAFE is directly applicable to multilayer perceptrons and FT-Transformers. We also conduct preliminary experiments with applying CAFE to convolutional neural networks, and believe that it would be relatively easy to extend to other neural architectures.
	
	\subsection{What value should I use for the cancelation sensitivity constants?}
	Please see Section \ref{sec:choice-of-c} of the main text for the relevant discussion.
	
	\subsection{What value should I use for the reference input?}
	Similarly to the cancelation sensitivity constant, the appropriate reference input is likely to be context-dependent. For continuous tabular data, vectors of zeroes or mean/median values are common choices. We believe the zero values to be a slightly more principled choice, as it forces the reference model (i.e., the model without biases) to output zero in regression tasks or a uniform distribution in classification tasks. Given this, the completeness property ensures that all changes from this base output are explained by the attribution scores. Additionally, the zero reference is the only option that guarantees that the model does not use any of the reference values in its computations, potentially masking their effect on the prediction during the computation of the attribution scores. Nevertheless, the use of zero reference vectors could also be criticised, as these will often be outside of the data manifold, making the corresponding attributions potentially less grounded in the behaviour of the model on the data distribution. From this point of view, the use of mean or median values may be more appropriate. For categorical data, we believe that a reasonable choice of reference is again zero, although the mode or a domain-specific base category may also be reasonable choices.
	
	\subsection{Are there any benefits to using CAFE if I do not care about surfacing conflicts between features?}
	Firstly, we would like to highlight that consideration of conflicts is helpful for producing explanations with a higher fidelity, as evidenced in our experiments. Thus, we believe that using CAFE with non-zero conflict sensitivity constants may produce generally more useful attribution scores, even if surfacing conflicts is not the explicit goal. Nevertheless, even using CAFE (0.0) offers several potential advantages over other methods — in particular, quantifying the effects of the bias, safeguarding against attribution score explosion and distinguishing between positive and negative attribution scores.
	
	\subsection{What are the considerations to take when explaining models using CAFE?}
	\label{apd:faq-considerations}
	Note that the attribution scores computed by CAFE (as well as other attribution methods) are only approximations of the true behaviour of the explained model and may not provide a fully accurate account of the effects of the individual input features. In particular, ``reasonable-looking" attribution scores for a particular example are not necessarily a guarantee that the explained model functions correctly or makes correct decisions. Past research has also suggested that cognitive biases of explanation recipients may lead them to incorrect judgements about the behaviour of the model even if the generated scores are accurate \cite{visual-explanations-useful, ghassemi-false-hope}. Thus, while feature attribution scores may help uncover and diagnose issues, they cannot exclude their presence and should be used critically in combination with other model evaluation and verification techniques.
	
	Also, while CAFE aims to accurately identify conflicts and safeguard against attribution score explosion, there are still some cases in which its scores may not fully capture feature conflicts or may overestimate the effects of the input features. See the answers to the below questions for details.
	
	\subsection{Are there any cases in which CAFE cannot fully capture conflicts between features?}
	\label{apd:faq-no-conflicts}
	While CAFE aims to surface most of the conflicts between features, there are some cases in which it is unable to do so, even when using the conflict sensitivity constant of $1$. In particular, CAFE may not always capture hypothetical effects of positive or negative features that are masked by the non-monotonicity of the considered activation function. For example, when considering a GELU neuron with the inputs $\inputelem{1} = -2$ and $\inputelem{2} = -5$, CAFE will conclude that both features have no effect on the output, since $\GELU(-7) - \GELU(0) = 0$. This ignores the fact that the GELU function decreases and then increases again on the interval, and that the $\inputelem{1}$ input could change the neuron output to $-0.05$ in the absence of $\inputelem{2}$. Effects canceled in this way could theoretically be captured by considering all activation function extrema within the intervals examined by CAFE. The computed effects could then be added to both positive and negative attribution scores so as to maintain completeness. Nevertheless, we do not believe that doing so would make the attribution scores particularly more insightful. Additionally, having to consider arbitrary activation function extrema would make our method considerably more complicated.
	
	\subsection{Are there any cases in which CAFE overestimates the effects of the input features?}
	\label{apd:faq-overestimates}
	Yes, despite the safeguards against the attribution score explosion problem, CAFE can overestimate the possible effects of certain features. For example, consider a model $\model(x) = \relu(-\relu(1-x))$. It is easy to see that the output of this model will always be zero regardless of the value of $x$, even if the bias is removed. Nevertheless, for an input $x = 1$, CAFE (1.0) computes attribution score of $+1$ for $x$ and $-1$ for the bias. This is because the CAFE activation rule can only consider the local behaviour at the individual neurons rather than the global behaviour of the model.
	
	\subsection{Are there any relations between CAFE and other feature attribution methods?}
	Since CAFE (0.0) effectively computes average gradients of the non-linearities (with additional adjustments safeguarding against attribution score explosion), its combined positive and negative feature scores are equivalent to Gradient $\cdot$ Input, LRP and DeepLIFT when computed with respect to a zero reference input for a network with ReLU activations and no additive biases. The linear attribution flow clipping does not affect CAFE attributions computed for ReLU networks, as the slope of the activation function on its active range is constant.
	
	\section{Details on Related Work}
	\label{sec:additional-related-work}
	\subsection{Other Relevant Explanation Methods.}
	Related to the aim of unearthing conflicts, existing work on deliberative explanations~ \cite{wang-deliberative-explanations} emphasized the importance of capturing insecurities in NNs, though it focused on producing sets of potentially ambiguous input regions instead of attribution scores, making it orthogonal to our approach. Also related to internal model deliberations are contrastive explanations with pertinent negatives \cite{dhurandhar-pertinent-negatives}, which highlight the missing parts of inputs that could cause the model to predict different classes, making them closer in spirit to counterfactual explanations. Contrastive LRP \cite{gu-contrastive-lrp} aims to make explanations for different classes more discriminative, but does not address conflicts between features. Finally, argumentative approaches like SpArX \cite{ayoobi-sparx} and ProtoArgNet \cite{ayoobi-protoargnet} aim at tracking the full reasoning of ANNs through representing them as graphs of attacking and supporting arguments. However, the produced explanations are considerably more complex than attribution scores and do not scale to recent deep architectures, such as transformers.
	
	\subsection{Explanation Evaluation}
	Evaluating AI model explanations, including feature attributions, remains an open research area, with a variety of discordant viewpoints on the most suitable methods for evaluation \cite{evaluating-ml-explanations, chen-what-makes-good-explanation, rahnama-blame-problem, nauta-anecdotal-evidence-quantitative-eval, le-benchmarking-xai}.
	
	Several performance metrics have been considered in the literature (e.g., see \cite{sokol2020explainability,nauta-anecdotal-evidence-quantitative-eval} for some overviews). Amongst these, \emph{fidelity} (also referred to as correctness, faithfulness or descriptive accuracy~\cite{nauta-anecdotal-evidence-quantitative-eval}) is widely regarded as crucial~\cite{infidelity-sensitivity-explanations}, as it amounts to explanations being truthful to the explained model. This metric is commonly operationalised through the inverse notions of infidelity~\cite{infidelity-sensitivity-explanations} and structural infidelity~\cite{ayoobi-sparx}. Other commonly considered metrics include the \emph{max-sensitivity} \cite{infidelity-sensitivity-explanations}, which assesses the robustness of explanations, and \emph{complexity} \cite{bhatt-faithfulness-corr}, which gauges the difficulty of understanding the explanations. We use these metrics for evaluating CAFE in our real-data experiments.
	
	In addition to using performance metrics on real data, controlled synthetic experiments have also been identified as a useful method for evaluating explanation methods \cite{nauta-anecdotal-evidence-quantitative-eval} -- we use these for assessing CAFE's capability to surface conflicts. Finally, CAFE is designed to satisfy several theoretical properties from the literature, in particular, \emph{completeness} \cite{sundararajan-integrated-gradients} (also known as ``summation-to-delta" \cite{shrikumar-deeplift} or ``sensitivity-N" \cite{ancona-towards-attribution}), which requires that the attribution scores sum to the change in the model output for the explained input, as well as \emph{missingness} \cite{lundberg-shap} and \emph{linearity} \cite{sundararajan-integrated-gradients}.
	
	\subsection{Additional Notes on Gradient-Based Methods}
	One of the simplest methods for computing the feature attribution scores is to take the gradient of the function modeled by the considered model with respect to its input \cite{simonyan-vanillagrad}. This gradient may additionally be multiplied with the inputs, which has been argued to produce more intuitive feature attribution scores \cite{shrikumar-old-deeplift-gxi}. Unfortunately, these simple approaches only indicate the behaviour of the model on an infinitesimally small region around the considered model input. If the function of the considered model is highly irregular or saturates for the given input (e.g., due to the score for the predicted class already being sufficiently high), its behaviour in a limited neighbourhood may not be indicative of the overall behaviour of the model.
	
	As an alternative to using raw gradients, the LRP method \cite{bach-on-pixel-wise-lrp, montavon-lrp-overview} introduced a set of rules for propagating relevance scores from the output of a neural network towards its inputs while maintaining the conservation property, analogical to the completeness property considered for methods using a reference input. It was later shown that the base variant of LRP is equivalent to Gradient $\cdot$ Input when applied to networks with ReLU activations \cite{shrikumar-old-deeplift-gxi}. Additionally, since LRP was only designed for networks using ReLU and tanh activation functions, it was demonstrated to behave erratically for activation functions for which $f(0) \neq 0$ \cite{ancona-towards-attribution}.
	
	DeepLIFT Rescale \cite{shrikumar-deeplift} and Integrated Gradients \cite{sundararajan-integrated-gradients} are two related feature attribution methods that aim to address the issues of instability and saturation of the model function by considering its gradients over a path from the currently considered input to a chosen reference point. The two methods differ in their approach to computing the scores — while Integrated Gradients estimate the mean gradient between the reference and the current input by sampling points along the path between them, DeepLIFT approximates the gradient by computing activation differences at the different neurons of the considered model. Both methods satisfy the completeness or summation-to-delta property, which stipulates that the produced attribution scores need to sum to the difference between the model output for the current input and the output for the reference input. In practice, both of these methods have been found to produce highly similar scores for most models, with DeepLIFT often being significantly faster to compute. However, DeepLIFT was also found to behave erratically on networks using multiplicative interactions for which Integrated Gradients still returned reasonable scores \cite{ancona-towards-attribution}.
	
	As argued in the main text of our paper, gradient-based methods are generally unable to accurately capture conflicts between features and also often exhibit other limitations, such as the attribution score explosion problem.
	
	\section{Conflict Properties and Proofs}
	\label{sec:conflict-properties-proofs}
	In this section, we provide additional properties of conflicts for ReLU neurons as well as the proofs of all the properties that we introduced in Section \ref{sec:theory-of-conflicts}.
	
	\begin{proposition}
		\label{prop:conflict-relu-positive}
		For a ReLU neuron with $\actelem{n+1}{\idxthree} = \relu(\actelem{\currlayer}{\idxthree}) = \relu(\sum^{\ldim{\currlayer - 1}}_\idxone \weightelem{\currlayer}{\idxone}{\idxthree} \actelem{\currlayer - 1}{\idxone})$ and $\actelem{n}{\idxthree} > 0$ the feature sets $C$ and $C'$ are in a minimal local internal conflict with respect to $\refactvec{\currlayer - 1}$ iff:
		\begin{enumerate}
			\item $C = \{\idxone\}$, $C' = \{\idxtwo\}$ for some $i$, $j$, and
			\item $\weightelem{\currlayer}{\idxone}{\idxthree} (\actelem{\currlayer - 1}{\idxone} - \refactelem{\currlayer - 1}{\idxone}) < 0 < \weightelem{\currlayer}{\idxtwo}{\idxthree} (\actelem{\currlayer - 1}{\idxtwo} - \refactelem{\currlayer - 1}{\idxtwo})$ or $\weightelem{\currlayer}{\idxone}{\idxthree} (\actelem{\currlayer - 1}{\idxone} - \refactelem{\currlayer - 1}{\idxone}) > 0 > \weightelem{\currlayer}{\idxtwo}{\idxthree} (\actelem{\currlayer - 1}{\idxtwo} - \refactelem{\currlayer - 1}{\idxtwo})$
		\end{enumerate}
	\end{proposition}
	
	\begin{proposition}
		\label{prop:conflict-relu-negative}
		For a ReLU neuron with $\actelem{n+1}{\idxthree} = \relu(\actelem{\currlayer}{\idxthree}) = \relu(\sum^{\ldim{\currlayer - 1}}_\idxone \weightelem{\currlayer}{\idxone}{\idxthree} \actelem{\currlayer - 1}{\idxone})$ and $\actelem{n}{\idxthree} \leq 0$ the feature set $C$ counteracts feature set $C'$ in a minimal local internal conflict with respect to $\refactvec{\currlayer - 1}$ iff:
		\begin{enumerate}    
			\item $C' = \{\idxtwo\}$ for some $\idxtwo$, and
			\item $\forall \idxone \in C,  \weightelem{\currlayer}{\idxone}{\idxthree} (\actelem{\currlayer - 1}{\idxone} - \refactelem{\currlayer - 1}{\idxone}) < 0$ and $\weightelem{\currlayer}{\idxtwo}{\idxthree} (\actelem{\currlayer - 1}{\idxtwo} - \refactelem{\currlayer - 1}{\idxtwo}) > 0$, and
			\item $|\actelem{\currlayer}{\idxthree}| < \sum_{\idxone \in C} \left|\weightelem{\currlayer}{\idxone}{\idxthree} (\actelem{\currlayer - 1}{\idxone} - \refactelem{\currlayer - 1}{\idxone})\right|$, and
			\item there is no $S \subset C$ for which $|\actelem{\currlayer}{\idxthree}| < \sum_{\idxone \in S} \Bigl|\weightelem{\currlayer}{\idxone}{\idxthree} (\actelem{\currlayer - 1}{\idxone} - \refactelem{\currlayer - 1}{\idxone})\Bigr|$
		\end{enumerate}
	\end{proposition}
	
	\begin{myproof}[Proposition \ref{prop:conflict-symmetry}]
		The symmetry of the conflict relation can be directly seen from Definition \ref{def:feature-conflicts}, which states that the sets of features $C$ and $C'$ are in conflict if $C$ counteracts $C'$ or $C'$ counteracts $C$. To see that the relation is anti-reflexive, consider the case when $C = C'$. Then, $\actvec{\currlayer - 1} \big| \remove(C) = \actvec{\currlayer - 1} \big| \remove(C') = \actvec{\currlayer - 1} \big| \remove(C \cup C')$, and thus $\layer{\currlayer:m}\left(\actvec{\currlayer - 1} \big| \remove(C)\right)_\idxthree = \layer{\currlayer:m}\left(\actvec{\currlayer - 1} \big| \remove(C')\right)_\idxthree = \layer{\currlayer:m}\left(\actvec{\currlayer - 1} \big| \remove(C \cup C')\right)_\idxthree$. Then, it is also the case that $\layer{\currlayer:m}\left(\actvec{\currlayer - 1} \big| \remove(C)\right)_\idxthree - \layer{\currlayer:m}\left(\actvec{\currlayer - 1} \big| \remove(C \cup C')\right)_\idxthree = \layer{\currlayer:m}\left(\actvec{\currlayer - 1} \big| \remove(C')\right)_\idxthree - \layer{\currlayer:m}\left(\actvec{\currlayer - 1} \big| \remove(C \cup C')\right)_\idxthree = 0$, and the condition from Definition \ref{def:counteractive-features} can never be satisfied. \aptLtoX[graphic=no,type=html]{&#x25A1;}{{$\square$}}
	\end{myproof}
	
	\begin{myproof}[Proposition \ref{prop:conflict-eq-ref}]
		Assume towards a contradiction that there is a feature $\actelem{\currlayer}{\idxone} = \refactelem{\currlayer}{\idxone}$ and a pair of two feature sets in a minimal conflicts, $C$ and $C'$, such that $\idxone \in C$ or $\idxone \in C'$. Without loss of generality, assume that $C$ counteracts $C'$ (the other case is symmetric). Consider the following two cases:
		\begin{enumerate}
			\item[(i)] $\idxone \in C$. Then, $\actvec{\currlayer - 1} \big| \remove(C) =  \actvec{\currlayer - 1} \big| \remove(C \setminus \{\idxone\})$ and $\actvec{\currlayer - 1} \big| \remove(C \cup C') =  \actvec{\currlayer - 1} \big| \remove((C \setminus \{\idxone\}) \cup C')$, which means that there is a set $S = C \setminus \{\idxone\} \subset C$ with $S$ counteracting $C'$, which is a contradiction to minimality.
			\item[(ii)] $\idxone \in C'$. Then, $\actvec{\currlayer - 1} \big| \remove(C \cup C') =  \actvec{\currlayer - 1} \big| \remove(C \cup (C' \setminus \{\idxone\}))$, which means that there is a set $S' = C' \setminus \{\idxone\} \subset C'$ with $C$ counteracting $S'$, which is a contradiction to minimality.
		\end{enumerate}
		Thus, $\actelem{\currlayer}{\idxone} = \refactelem{\currlayer}{\idxone}$ can never be an element of any minimal counteractive feature set or a feature set in minimal conflict. \aptLtoX[graphic=no,type=html]{&#x25A1;}{{$\square$}}
	\end{myproof}
	
	\begin{myproof}[Proposition \ref{prop:conflict-linear}]
		To establish the equivalence, we first show that there is necessarily a conflict if the listed conditions on $C$ and $C'$ are satisfied, before showing the implication in the opposite direction.
		
		Take an arbitrary linear model $\model(\inputvec) = \sum^{\ldim{\inputvec}}_\idxone w_\idxone \inputelem{\idxone} + b$, input $\inputvec$, a reference input $\refinputvec$, and two feature sets $C$ and $C'$ for which $C = \{\idxone\}, C' = \{\idxtwo\}$ for some $\idxone$, $\idxtwo$, and $w_\idxone (\inputelem{\idxone} - \refinputelem{\idxone}) < 0 < w_\idxtwo (\inputelem{\idxtwo} - \refinputelem{\idxtwo})$ or $w_\idxone (\inputelem{\idxone} - \refinputelem{\idxone}) > 0 > w_\idxtwo (\inputelem{\idxtwo} - \refinputelem{\idxtwo})$. Without loss of generality, assume that $w_\idxone (\inputelem{\idxone} - \refinputelem{\idxone}) < 0 < w_\idxtwo (\inputelem{\idxtwo} - \refinputelem{\idxtwo})$ (the other case is symmetric). Then,
		\begin{align*}
			&\frac{
				\model\left(\inputvec\right)
				- \model\left(\inputvec \big| \remove(C)\right)
			}{
				\model\left(\inputvec \big| \remove(C)\right) - \model\left(\inputvec \big| \remove(C \cup C')\right)
			} = \frac{
				\model\left(\inputvec\right)
				- \model\left(\inputvec \big| \inputelem{\idxone} := \refinputelem{\idxone} \right)
			}{
				\model\left(\inputvec \big| \inputelem{\idxone} := \refinputelem{\idxone}\right) - \model\left(\inputvec \big| \inputelem{\idxone} := \refinputelem{\idxone}, \inputelem{\idxtwo} := \refinputelem{\idxtwo}\right)
			} \\
			&= \frac{
				\sum^{\ldim{\inputvec}}_{\idxthree \neq \idxone} w_\idxthree \inputelem{\idxthree} + w_\idxone \inputelem{\idxone} + b - \sum^{\ldim{\inputvec}}_{\idxthree \neq \idxone} w_\idxthree \inputelem{\idxthree} - w_\idxone \refinputelem{\idxone} - b
			}{
				\sum^{\ldim{\inputvec}}_{\idxthree \neq \idxone, \idxthree \neq \idxtwo} w_\idxthree \inputelem{\idxthree} + w_\idxone \refinputelem{\idxone} + w_\idxtwo \inputelem{\idxtwo} + b - \sum^{\ldim{\inputvec}}_{\idxthree \neq \idxone, \idxthree \neq \idxtwo} w_\idxthree \inputelem{\idxthree} - w_\idxone \refinputelem{\idxone} - w_\idxtwo \refinputelem{\idxtwo} - b
			} = \frac{w_\idxone (\inputelem{\idxone} - \refinputelem{\idxone})}{w_\idxtwo (\inputelem{\idxtwo} - \refinputelem{\idxtwo})} < 0
		\end{align*}
		and thus $C$ and $C'$ are in a minimal input-output conflict with respect to $\refinputvec$ (the conflict is necessarily minimal, as an empty set can never be in conflict with another set), as required.
		
		To see that the implication also holds in the opposite direction, consider an arbitrary linear model $\model(\inputvec) = \sum^{\ldim{\inputvec}}_\idxone w_\idxone \inputelem{\idxone} + b$, input $\inputvec$, and two feature sets $C$ and $C'$ that are in a minimal input-output conflict with respect to $\refinputvec$. We consider the following exhaustive cases:
		\begin{enumerate}
			\item[(i)] $C = \emptyset$ or $C' = \emptyset$. This contradicts our assumptions, as an empty set can never be in conflict with another set, making this case impossible.
			\item[(ii)] $C \neq \emptyset$, $C' \neq \emptyset$, $\forall \idxone \in C, w_\idxone (\inputelem{\idxone} - \refinputelem{\idxone}) < 0$ and $\forall \idxtwo \in C', w_\idxtwo (\inputelem{\idxtwo} - \refinputelem{\idxtwo}) < 0$. Without loss of generality, we assume that $C$ counteracts $C'$ with respect to $\refinputvec$, as the two cases are symmetric. Then,
			\begin{align*}
				&\frac{
					\model\left(\inputvec\right)
					- \model\left(\inputvec \big| \remove(C)\right)
				}{
					\model\left(\inputvec \big| \remove(C)\right) - \model\left(\inputvec \big| \remove(C \cup C')\right)
				} = \frac{
					\model\left(\inputvec\right)
					- \model\left(\inputvec \big| \left[\inputelem{\idxone} := \refinputelem{\idxone}\right]_{\idxone \in C} \right)
				}{
					\model\left(\inputvec \big| \left[\inputelem{\idxone} := \refinputelem{\idxone}\right]_{\idxone \in C}\right) - \model\left(\inputvec \big| \left[\inputelem{\idxone} := \refinputelem{\idxone}\right]_{\idxone \in C \cup C'}\right)
				} \\
				&= \frac{
					\sum^{\ldim{\inputvec}}_{\idxthree \notin C} w_\idxthree \inputelem{\idxthree} + \sum_{\idxone \in C} w_\idxone \inputelem{\idxone} + b - \sum^{\ldim{\inputvec}}_{\idxthree \notin C} w_\idxthree \inputelem{\idxthree} - \sum_{\idxone \in C} w_\idxone \refinputelem{\idxone} - b
				}{
					\sum^{\ldim{\inputvec}}_{\idxthree \notin C \cup C'} w_\idxthree \inputelem{\idxthree} + \sum_{\idxone \in C} w_\idxone \refinputelem{\idxone} + \sum_{\idxtwo \in C'} w_\idxtwo \inputelem{\idxtwo} + b - \sum^{\ldim{\inputvec}}_{\idxthree \notin C \cup C'} w_\idxthree \inputelem{\idxthree} - \sum_{\idxone \in C} w_\idxone \refinputelem{\idxone} - \sum_{\idxtwo \in C'} w_\idxtwo \refinputelem{\idxtwo} - b
				} \\
				&= \frac{\sum_{\idxone \in C} w_\idxone (\inputelem{\idxone} - \refinputelem{\idxone})}{\sum_{\idxtwo \in C'} w_\idxtwo (\inputelem{\idxtwo} - \refinputelem{\idxtwo})} > 0
			\end{align*}
			with the last inequality holding due to $\sum_{\idxone \in C} w_\idxone (\inputelem{\idxone} - \refinputelem{\idxone}) < 0$ and $\sum_{\idxtwo \in C'} w_\idxtwo (\inputelem{\idxtwo} - \refinputelem{\idxtwo}) < 0$. This is a contradiction, since by Definition \ref{def:counteractive-features}, $C$ doesn't counteract $C'$ with respect to $\refinputvec$ if the above inequality holds, and thus this case cannot arise given the assumptions.
			\item[(iii)] $C \neq \emptyset$, $C' \neq \emptyset$, $\forall \idxone \in C, w_\idxone (\inputelem{\idxone} - \refinputelem{\idxone}) > 0$ and $\forall \idxtwo \in C', w_\idxtwo (\inputelem{\idxtwo} - \refinputelem{\idxtwo}) > 0$. This case also cannot arise by analogous reasoning as for the previous case, with the final inequality holding due to $\sum_{\idxone \in C} w_\idxone (\inputelem{\idxone} - \refinputelem{\idxone}) > 0$ and $\sum_{\idxtwo \in C'} w_\idxtwo (\inputelem{\idxtwo} - \refinputelem{\idxtwo}) > 0$.
			\item[(iv)] $\exists \idxone \in C, w_\idxone (\inputelem{\idxone} - \refinputelem{\idxone}) > 0$ and $\exists \idxtwo \in C', w_\idxtwo (\inputelem{\idxtwo} - \refinputelem{\idxtwo}) < 0$. Consider sets $S = \{\idxone\}$ and $S' = \{\idxtwo\}$ with $\idxone$ and $\idxtwo$ being the witnesses to the case assumptions. Then, by what we have shown above, $S$ and $S'$ are in a minimal input-output conflict with respect to $\refinputvec$. But then, it must necessarily be the case that $C = S$ and $C' = S'$, otherwise the conflict between them could not be minimal, which would violate our assumptions. It immediately follows that $C = \{\idxone\}$, $C' = \{\idxtwo\}$ and $w_\idxone (\inputelem{\idxone} - \refinputelem{\idxone}) > 0 > w_\idxtwo (\inputelem{\idxtwo} - \refinputelem{\idxtwo})$, as required.
			\item[(v)] $\exists \idxone \in C, w_\idxone (\inputelem{\idxone} - \refinputelem{\idxone}) < 0$ and $\exists \idxtwo \in C', w_\idxtwo (\inputelem{\idxtwo} - \refinputelem{\idxtwo}) > 0$. By analogous reasoning as for the previous case, we get that $C = \{\idxone\}$, $C' = \{\idxtwo\}$ and $w_\idxone (\inputelem{\idxone} - \refinputelem{\idxone}) < 0 < w_\idxtwo (\inputelem{\idxtwo} - \refinputelem{\idxtwo})$, as required.
		\end{enumerate}
		This concludes the proof. \aptLtoX[graphic=no,type=html]{&#x25A1;}{{$\square$}}
	\end{myproof}
	
	\begin{myproof}[Proposition \ref{prop:conflict-relu-positive}]
		To establish the equivalence, we first show that there is necessarily a conflict if the listed conditions are satisfied, before showing the implication in the opposite direction.
		
		Take an arbitrary ReLU neuron with $\actelem{n+1}{\idxthree} = \relu(\actelem{\currlayer}{\idxthree}) = \relu(\sum^{\ldim{\currlayer - 1}}_\idxone \weightelem{\currlayer}{\idxone}{\idxthree} \actelem{\currlayer - 1}{\idxone})$, $\actelem{n}{\idxthree} > 0$, a reference activation vector $\refactvec{\currlayer - 1}$, and two feature sets $C$ and $C'$ for which $C = \{\idxone\}$, $C' = \{\idxtwo\}$ for some $i$, $j$, and $\weightelem{\currlayer}{\idxone}{\idxthree} (\actelem{\currlayer - 1}{\idxone} - \refactelem{\currlayer - 1}{\idxone}) < 0 < \weightelem{\currlayer}{\idxtwo}{\idxthree} (\actelem{\currlayer - 1}{\idxtwo} - \refactelem{\currlayer - 1}{\idxtwo})$ or $\weightelem{\currlayer}{\idxone}{\idxthree} (\actelem{\currlayer - 1}{\idxone} - \refactelem{\currlayer - 1}{\idxone}) > 0 > \weightelem{\currlayer}{\idxtwo}{\idxthree} (\actelem{\currlayer - 1}{\idxtwo} - \refactelem{\currlayer - 1}{\idxtwo})$. Without loss of generality, assume that $\weightelem{\currlayer}{\idxone}{\idxthree} (\actelem{\currlayer - 1}{\idxone} - \refactelem{\currlayer - 1}{\idxone}) < 0 < \weightelem{\currlayer}{\idxtwo}{\idxthree} (\actelem{\currlayer - 1}{\idxtwo} - \refactelem{\currlayer - 1}{\idxtwo})$, with the other case being symmetric. Then,
		\begin{align*}
			&\layer{\currlayer:\currlayer + 1}\left(\actvec{\currlayer - 1}\right)_\idxthree - \layer{\currlayer:\currlayer + 1}\left(\actvec{\currlayer - 1} \big| \remove(C)\right)_\idxthree
			= \layer{\currlayer:\currlayer + 1}\left(\actvec{\currlayer - 1}\right)_\idxthree - \layer{\currlayer:\currlayer + 1}\left(\actvec{\currlayer - 1} \big| \actelem{\currlayer - 1}{\idxone} := \refactelem{\currlayer}{\idxone} \right)_\idxthree \\
			&= \relu\left(\sum^{\ldim{\currlayer - 1}}_{\idxfour \neq \idxone} \weightelem{\currlayer}{\idxfour}{\idxthree} \actelem{\currlayer - 1}{\idxfour} + \weightelem{\currlayer}{\idxone}{\idxthree} \actelem{\currlayer - 1}{\idxone}\right)
			- \relu\left(\sum^{\ldim{\currlayer - 1}}_{\idxfour \neq \idxone} \weightelem{\currlayer}{\idxfour}{\idxthree} \actelem{\currlayer - 1}{\idxfour} + \weightelem{\currlayer}{\idxone}{\idxthree} \refactelem{\currlayer - 1}{\idxone}\right) < 0
		\end{align*}
		The final inequality directly follows from the assumptions $\weightelem{\currlayer}{\idxone}{\idxthree} \actelem{\currlayer - 1}{\idxone} < \weightelem{\currlayer}{\idxone}{\idxthree} \refactelem{\currlayer - 1}{\idxone}$ and $\actelem{n}{\idxthree} = \sum^{\ldim{\currlayer - 1}}_{\idxfour \neq \idxone} \weightelem{\currlayer}{\idxfour}{\idxthree} \actelem{\currlayer - 1}{\idxfour} + \weightelem{\currlayer}{\idxone}{\idxthree} \actelem{\currlayer - 1}{\idxone} > 0$, as well as $\relu$ being strictly increasing for positive inputs. Similarly,
		\begin{align*}
			&\layer{\currlayer:\currlayer + 1}\left(\actvec{\currlayer - 1} \big| \remove(C)\right)_\idxthree - \layer{\currlayer:\currlayer + 1}\left(\actvec{\currlayer - 1} \big| \remove(C \cup C')\right)_\idxthree \\
			&= \layer{\currlayer:\currlayer + 1}\left(\actvec{\currlayer - 1} \big| \actelem{\currlayer - 1}{\idxone} := \refactelem{\currlayer}{\idxone} \right)_\idxthree - \layer{\currlayer:\currlayer + 1}\left(\actvec{\currlayer - 1} \big| \actelem{\currlayer - 1}{\idxone} := \refactelem{\currlayer}{\idxone},  \actelem{\currlayer - 1}{\idxtwo} := \refactelem{\currlayer}{\idxtwo}\right)_\idxthree \\
			&= \relu\left(\sum^{\ldim{\currlayer - 1}}_{\idxfour \neq \idxone, \idxfour \neq \idxtwo} \weightelem{\currlayer}{\idxfour}{\idxthree} \actelem{\currlayer - 1}{\idxfour} + \weightelem{\currlayer}{\idxone}{\idxthree} \refactelem{\currlayer - 1}{\idxone} + \weightelem{\currlayer}{\idxtwo}{\idxthree} \actelem{\currlayer - 1}{\idxtwo}\right) \\
			&\phantom{=(}- \relu\left(\sum^{\ldim{\currlayer - 1}}_{\idxfour \neq \idxone, \idxfour \neq \idxtwo} \weightelem{\currlayer}{\idxfour}{\idxthree} \actelem{\currlayer - 1}{\idxfour} + \weightelem{\currlayer}{\idxone}{\idxthree} \refactelem{\currlayer - 1}{\idxone} + \weightelem{\currlayer}{\idxtwo}{\idxthree} \refactelem{\currlayer - 1}{\idxtwo}\right) \\
			&> 0
		\end{align*}
		The final inequality directly follows from $\weightelem{\currlayer}{\idxtwo}{\idxthree} \actelem{\currlayer - 1}{\idxone} > \weightelem{\currlayer}{\idxtwo}{\idxthree} \refactelem{\currlayer - 1}{\idxtwo}$, $\sum^{\ldim{\currlayer - 1}}_{\idxfour \neq \idxone, \idxfour \neq \idxtwo} \weightelem{\currlayer}{\idxfour}{\idxthree} \actelem{\currlayer - 1}{\idxfour} + \weightelem{\currlayer}{\idxone}{\idxthree} \refactelem{\currlayer - 1}{\idxone} + \weightelem{\currlayer}{\idxtwo}{\idxthree} \actelem{\currlayer - 1}{\idxtwo} > \actelem{n}{\idxthree} > 0$, as well as ReLU being increasing for all inputs and strictly increasing for positive inputs.
		
		From the above, we can immediately see that:
		\begin{align*}
			\frac{
				\layer{\currlayer:\currlayer + 1}\left(\actvec{\currlayer - 1}\right)_\idxthree
				- \layer{\currlayer:\currlayer + 1}\left(\actvec{\currlayer - 1} \big| \remove(C)\right)_\idxthree
			}{
				\layer{\currlayer:\currlayer + 1}\left(\actvec{\currlayer - 1} \big| \remove(C)\right)_\idxthree - \layer{\currlayer:\currlayer + 1}\left(\actvec{\currlayer - 1} \big| \remove(C \cup C')\right)_\idxthree
			} < 0
		\end{align*}
		and thus $C$ and $C'$ are in a minimal local internal conflict with respect to $\refactvec{\currlayer - 1}$, as required.
		
		To see that the implication also holds in the opposite direction, consider an arbitrary ReLU neuron with $\actelem{n+1}{\idxthree} = \relu(\actelem{\currlayer}{\idxthree}) = \relu(\allowbreak \sum^{\ldim{\currlayer - 1}}_\idxone \weightelem{\currlayer}{\idxone}{\idxthree} \actelem{\currlayer - 1}{\idxone})$, $\actelem{n}{\idxthree} > 0$, and two feature sets $C$ and $C'$  that are in a minimal local internal conflict with respect to $\refactvec{\currlayer - 1}$. We consider the following exhaustive cases:
		\begin{enumerate}
			\item[(i)] $C = \emptyset$ or $C' = \emptyset$. This contradicts our assumptions, as an empty set can never be in conflict with another set, making this case impossible.
			\item[(ii)] $C \neq \emptyset$, $C' \neq \emptyset$, $\forall \idxone \in C, \weightelem{\currlayer}{\idxone}{\idxthree} (\actelem{\currlayer - 1}{\idxone} - \refactelem{\currlayer - 1}{\idxone}) < 0$ and $\forall \idxtwo \in C', \weightelem{\currlayer}{\idxtwo}{\idxthree} (\actelem{\currlayer - 1}{\idxtwo} - \refactelem{\currlayer - 1}{\idxtwo}) < 0$. Without loss of generality, we assume that $C$ counteracts $C'$ with respect to $\refactvec{\currlayer - 1}$, as the two cases are symmetric. Then,
			\begin{align*}
				&\layer{\currlayer:\currlayer + 1}\left(\actvec{\currlayer - 1}\right)_\idxthree - \layer{\currlayer:\currlayer + 1}\left(\actvec{\currlayer - 1} \big| \remove(C)\right)_\idxthree
				= \layer{\currlayer:\currlayer + 1}\left(\actvec{\currlayer - 1}\right)_\idxthree - \layer{\currlayer:\currlayer + 1}\left(\actvec{\currlayer - 1} \big| \left[\actelem{\currlayer - 1}{\idxone} := \refactelem{\currlayer}{\idxone}\right]_{\idxone \in C} \right)_\idxthree \\
				&= \relu\left(\sum^{\ldim{\currlayer - 1}}_{\idxfour \notin C} \weightelem{\currlayer}{\idxfour}{\idxthree} \actelem{\currlayer - 1}{\idxfour} + \sum^{\ldim{\currlayer - 1}}_{\idxone \in C} \weightelem{\currlayer}{\idxone}{\idxthree} \actelem{\currlayer - 1}{\idxone}\right)
				- \relu\left(\sum^{\ldim{\currlayer - 1}}_{\idxfour \notin C} \weightelem{\currlayer}{\idxfour}{\idxthree} \actelem{\currlayer - 1}{\idxfour} + \sum^{\ldim{\currlayer - 1}}_{\idxfour \in C} \weightelem{\currlayer}{\idxone}{\idxthree} \refactelem{\currlayer - 1}{\idxone}\right) < 0
			\end{align*}
			with the last inequality holding due to $\sum^{\ldim{\currlayer - 1}}_{\idxone \in C} \weightelem{\currlayer}{\idxone}{\idxthree} \actelem{\currlayer - 1}{\idxone} < \sum^{\ldim{\currlayer - 1}}_{\idxone \in C} \weightelem{\currlayer}{\idxone}{\idxthree} \refactelem{\currlayer - 1}{\idxone}$ (which directly follows from the case assumptions), $\actelem{\currlayer}{\idxthree} = \sum^{\ldim{\currlayer - 1}}_{\idxfour \notin C} \weightelem{\currlayer}{\idxfour}{\idxthree} \actelem{\currlayer - 1}{\idxfour} + \sum^{\ldim{\currlayer - 1}}_{\idxone \in C} \weightelem{\currlayer}{\idxone}{\idxthree} \actelem{\currlayer - 1}{\idxone} > 0$, as well as ReLU being strictly increasing for positive inputs. Similarly,
			\begin{align*}
				&\layer{\currlayer:\currlayer + 1}\left(\actvec{\currlayer - 1} \big| \remove(C)\right)_\idxthree - \layer{\currlayer:\currlayer + 1}\left(\actvec{\currlayer - 1} \big| \remove(C \cup C')\right)_\idxthree \\
				&= \layer{\currlayer:\currlayer + 1}\left(\actvec{\currlayer - 1} \big| \left[\actelem{\currlayer - 1}{\idxone} := \refactelem{\currlayer}{\idxone}\right]_{\idxone \in C} \right)_\idxthree - \layer{\currlayer:\currlayer + 1}\left(\actvec{\currlayer - 1} \big| \left[\actelem{\currlayer - 1}{\idxone} := \refactelem{\currlayer}{\idxone}\right]_{\idxone \in C \cup C'}\right)_\idxthree \\
				&= \relu\left(\sum^{\ldim{\currlayer - 1}}_{\idxfour \notin C \cup C'} \weightelem{\currlayer}{\idxfour}{\idxthree} \actelem{\currlayer - 1}{\idxfour} + \sum^{\ldim{\currlayer - 1}}_{\idxone \in C} \weightelem{\currlayer}{\idxone}{\idxthree} \refactelem{\currlayer - 1}{\idxone} + \sum^{\ldim{\currlayer - 1}}_{\idxtwo \in C'} \weightelem{\currlayer}{\idxtwo}{\idxthree} \actelem{\currlayer - 1}{\idxtwo}\right) \\
				&\phantom{=(}- \relu\left(\sum^{\ldim{\currlayer - 1}}_{\idxfour \notin C \cup C'} \weightelem{\currlayer}{\idxfour}{\idxthree} \actelem{\currlayer - 1}{\idxfour} + \sum^{\ldim{\currlayer - 1}}_{\idxone \in C} \weightelem{\currlayer}{\idxone}{\idxthree} \refactelem{\currlayer - 1}{\idxone} + \sum^{\ldim{\currlayer - 1}}_{\idxtwo \in C'} \weightelem{\currlayer}{\idxtwo}{\idxthree} \refactelem{\currlayer - 1}{\idxtwo}\right) \\
				&< 0
			\end{align*}
			with the last inequality holding due to $\sum^{\ldim{\currlayer - 1}}_{\idxtwo \in C'} \weightelem{\currlayer}{\idxtwo}{\idxthree} \actelem{\currlayer - 1}{\idxtwo} < \sum^{\ldim{\currlayer - 1}}_{\idxtwo \in C'} \weightelem{\currlayer}{\idxtwo}{\idxthree} \refactelem{\currlayer - 1}{\idxtwo}$ (which directly follows from the case assumptions), $\sum^{\ldim{\currlayer - 1}}_{\idxfour \notin C \cup C'} \weightelem{\currlayer}{\idxfour}{\idxthree} \actelem{\currlayer - 1}{\idxfour} + \sum^{\ldim{\currlayer - 1}}_{\idxone \in C} \weightelem{\currlayer}{\idxone}{\idxthree} \refactelem{\currlayer - 1}{\idxone} + \sum^{\ldim{\currlayer - 1}}_{\idxtwo \in C'} \weightelem{\currlayer}{\idxtwo}{\idxthree} \actelem{\currlayer - 1}{\idxtwo} > \actelem{\currlayer}{\idxthree} > 0$, as well as ReLU being strictly increasing for positive inputs. The results above immediately give us that:
			\begin{align*}
				\frac{
					\layer{\currlayer:\currlayer + 1}\left(\actvec{\currlayer - 1}\right)_\idxthree
					- \layer{\currlayer:\currlayer + 1}\left(\actvec{\currlayer - 1} \big| \remove(C)\right)_\idxthree
				}{
					\layer{\currlayer:\currlayer + 1}\left(\actvec{\currlayer - 1} \big| \remove(C)\right)_\idxthree - \layer{\currlayer:\currlayer + 1}\left(\actvec{\currlayer - 1} \big| \remove(C \cup C')\right)_\idxthree
				} > 0
			\end{align*}
			This is a contradiction, since by Definition \ref{def:counteractive-features}, $C$ cannot counteract $C'$ with respect to $\refactvec{\currlayer - 1}$ if the above inequality holds, and thus this case cannot arise given the assumptions.       
			\item[(iii)] $C \neq \emptyset$, $C' \neq \emptyset$, $\forall \idxone \in C, \weightelem{\currlayer}{\idxone}{\idxthree} (\actelem{\currlayer - 1}{\idxone} - \refactelem{\currlayer - 1}{\idxone}) > 0$ and $\forall \idxtwo \in C', \weightelem{\currlayer}{\idxtwo}{\idxthree} (\actelem{\currlayer - 1}{\idxtwo} - \refactelem{\currlayer - 1}{\idxtwo}) > 0$. This case also cannot arise by similar reasoning as for the previous case, with $\layer{\currlayer:\currlayer + 1}\left(\actvec{\currlayer - 1}\right)_\idxthree - \layer{\currlayer:\currlayer + 1}\left(\actvec{\currlayer - 1} \big| \remove(C)\right)_\idxthree > 0$, $\layer{\currlayer:\currlayer + 1}\left(\actvec{\currlayer - 1} \big| \remove(C)\right)_\idxthree - \layer{\currlayer:\currlayer + 1}\left(\actvec{\currlayer - 1} \big| \remove(C \cup C')\right)_\idxthree \geq 0$ and the ratio from Definition \ref{def:counteractive-features} either being positive or undefined (with $0$ in the denominator). These inequalities hold due to $\sum^{\ldim{\currlayer - 1}}_{\idxone \in C} \weightelem{\currlayer}{\idxone}{\idxthree} \actelem{\currlayer - 1}{\idxone} > \sum^{\ldim{\currlayer - 1}}_{\idxone \in C} \weightelem{\currlayer}{\idxone}{\idxthree} \refactelem{\currlayer - 1}{\idxone}$, $\sum^{\ldim{\currlayer - 1}}_{\idxtwo \in C'} \weightelem{\currlayer}{\idxtwo}{\idxthree} \actelem{\currlayer - 1}{\idxtwo} > \sum^{\ldim{\currlayer - 1}}_{\idxtwo \in C'} \weightelem{\currlayer}{\idxtwo}{\idxthree} \refactelem{\currlayer - 1}{\idxtwo}$, $\refactelem{\currlayer}{\idxthree} > 0$ as well as ReLU being increasing for all inputs and strictly increasing for positive inputs.
			\item[(iv)] $\exists \idxone \in C, \weightelem{\currlayer}{\idxone}{\idxthree} (\actelem{\currlayer - 1}{\idxone} - \refactelem{\currlayer - 1}{\idxone}) > 0$ and $\exists \idxtwo \in C', \weightelem{\currlayer}{\idxtwo}{\idxthree} (\actelem{\currlayer - 1}{\idxtwo} - \refactelem{\currlayer - 1}{\idxtwo}) < 0$. Consider sets $S = \{\idxone\}$ and $S' = \{\idxtwo\}$ with $\idxone$ and $\idxtwo$ being the witnesses to the case assumptions. Then, by what we have shown above, $S$ and $S'$ are in a minimal input-output conflict with respect to $\refactvec{\currlayer - 1}$. But then, it must necessarily be the case that $C = S$ and $C' = S'$, otherwise the conflict between them could not be minimal, which would violate our assumptions. It immediately follows that $C = \{\idxone\}$, $C' = \{\idxtwo\}$ and $\weightelem{\currlayer}{\idxone}{\idxthree} (\actelem{\currlayer - 1}{\idxone} - \refactelem{\currlayer - 1}{\idxone}) > 0 > \weightelem{\currlayer}{\idxtwo}{\idxthree} (\actelem{\currlayer - 1}{\idxtwo} - \refactelem{\currlayer - 1}{\idxtwo})$, as required.
			\item[(v)] $\exists \idxone \in C, \weightelem{\currlayer}{\idxone}{\idxthree} (\actelem{\currlayer - 1}{\idxone} - \refactelem{\currlayer - 1}{\idxone}) < 0$ and $\exists \idxtwo \in C',\weightelem{\currlayer}{\idxtwo}{\idxthree} (\actelem{\currlayer - 1}{\idxtwo} - \refactelem{\currlayer - 1}{\idxtwo}) > 0$. By analogous reasoning as for the previous case, we get that $C = \{\idxone\}$, $C' = \{\idxtwo\}$ and $\weightelem{\currlayer}{\idxone}{\idxthree} (\actelem{\currlayer - 1}{\idxone} - \refactelem{\currlayer - 1}{\idxone}) < 0 < \weightelem{\currlayer}{\idxtwo}{\idxthree} (\actelem{\currlayer - 1}{\idxtwo} - \refactelem{\currlayer - 1}{\idxtwo})$, as required.
		\end{enumerate}
		This concludes the proof. \aptLtoX[graphic=no,type=html]{&#x25A1;}{{$\square$}}
	\end{myproof}
	
	\begin{myproof}[Proposition \ref{prop:conflict-relu-negative}]
		To establish the equivalence, we first show that a feature set $C$ counteracts feature set $C'$ if the listed conditions are satisfied, before showing the implication in the opposite direction.
		
		Take an arbitrary ReLU neuron with $\actelem{n+1}{\idxthree} = \relu(\actelem{\currlayer}{\idxthree}) = \relu(\sum^{\ldim{\currlayer - 1}}_\idxone \weightelem{\currlayer}{\idxone}{\idxthree} \actelem{\currlayer - 1}{\idxone})$, $\actelem{n}{\idxthree} \leq 0$, a reference activation vector $\refactvec{\currlayer - 1}$, and two feature sets $C$ and $C'$ for which $C' = \{\idxtwo\}$ for some $\idxtwo$, $\forall \idxone \in C,  \weightelem{\currlayer}{\idxone}{\idxthree} (\actelem{\currlayer - 1}{\idxone} - \refactelem{\currlayer - 1}{\idxone}) < 0$, $\weightelem{\currlayer}{\idxtwo}{\idxthree} (\actelem{\currlayer - 1}{\idxtwo} - \refactelem{\currlayer - 1}{\idxtwo}) > 0$, $|\actelem{\currlayer}{\idxthree}| < \sum_{\idxone \in C} \left|\weightelem{\currlayer}{\idxone}{\idxthree} (\actelem{\currlayer - 1}{\idxone} - \refactelem{\currlayer - 1}{\idxone})\right|$, and there is no $S \subset C$ for which $|\actelem{\currlayer}{\idxthree}| < \sum_{\idxone \in S} \left|\weightelem{\currlayer}{\idxone}{\idxthree} (\actelem{\currlayer - 1}{\idxone} - \refactelem{\currlayer - 1}{\idxone})\right|$. Then,
		\begin{align*}
			&\layer{\currlayer:\currlayer + 1}\left(\actvec{\currlayer - 1}\right)_\idxthree - \layer{\currlayer:\currlayer + 1}\left(\actvec{\currlayer - 1} \big| \remove(C)\right)_\idxthree
			= \layer{\currlayer:\currlayer + 1}\left(\actvec{\currlayer - 1}\right)_\idxthree - \layer{\currlayer:\currlayer + 1}\left(\actvec{\currlayer - 1} \big| \left[\actelem{\currlayer - 1}{\idxone} := \refactelem{\currlayer}{\idxone}\right]_{\idxone \in C} \right)_\idxthree \\
			&= \relu\left(\actelem{n}{\idxthree}\right)
			- \relu\left(\actelem{n}{\idxthree} - \sum^{\ldim{\currlayer - 1}}_{\idxone \in C} \weightelem{\currlayer}{\idxone}{\idxthree} \actelem{\currlayer - 1}{\idxone} + \sum^{\ldim{\currlayer - 1}}_{\idxone \in C} \weightelem{\currlayer}{\idxone}{\idxthree} \refactelem{\currlayer - 1}{\idxone}\right) \\
			&= 0
			- \relu\left(\actelem{n}{\idxthree} - \sum^{\ldim{\currlayer - 1}}_{\idxone \in C} \weightelem{\currlayer}{\idxone}{\idxthree} (\actelem{\currlayer - 1}{\idxone} - \refactelem{\currlayer - 1}{\idxone})\right) < 0
		\end{align*}
		The equality on the last line follows from the assumption that $\actelem{\currlayer}{\idxthree} \leq 0$ and $\relu(x) = 0$ for any $x \leq 0$. To see that the final inequality holds, consider that $\forall \idxone \in C,  \weightelem{\currlayer}{\idxone}{\idxthree} (\actelem{\currlayer - 1}{\idxone} - \refactelem{\currlayer - 1}{\idxone}) < 0$ and that $|\actelem{\currlayer}{\idxthree}| < \sum_{\idxone \in C} \left|\weightelem{\currlayer}{\idxone}{\idxthree} (\actelem{\currlayer - 1}{\idxone} - \refactelem{\currlayer - 1}{\idxone})\right|$, from which it follows that $\actelem{n}{\idxthree} - \sum^{\ldim{\currlayer - 1}}_{\idxone \in C} \weightelem{\currlayer}{\idxone}{\idxthree} (\actelem{\currlayer - 1}{\idxone} - \refactelem{\currlayer - 1}{\idxone}) > 0$. Given that $\relu(x) = x$ for $x \geq 0$, this necessarily means that $\relu\left(\actelem{n}{\idxthree} - \sum^{\ldim{\currlayer - 1}}_{\idxone \in C} \weightelem{\currlayer}{\idxone}{\idxthree} (\actelem{\currlayer - 1}{\idxone} - \refactelem{\currlayer - 1}{\idxone})\right) > 0$, from which the final inequality follows immediately. Similarly,
		{\allowdisplaybreaks
			\begin{align*}
				&\layer{\currlayer:\currlayer + 1}\left(\actvec{\currlayer - 1} \big| \remove(C)\right)_\idxthree - \layer{\currlayer:\currlayer + 1}\left(\actvec{\currlayer - 1} \big| \remove(C \cup C')\right)_\idxthree \\
				&= \layer{\currlayer:\currlayer + 1}\left(\actvec{\currlayer - 1} \big| \left[\actelem{\currlayer - 1}{\idxone} := \refactelem{\currlayer}{\idxone}\right]_{\idxone \in C} \right)_\idxthree - \layer{\currlayer:\currlayer + 1}\left(\actvec{\currlayer - 1} \big| \left[\actelem{\currlayer - 1}{\idxone} := \refactelem{\currlayer}{\idxone}\right]_{\idxone \in C}, \actelem{\currlayer - 1}{\idxtwo} := \refactelem{\currlayer}{\idxtwo} \right)_\idxthree \\
				&= \relu\left(\actelem{n}{\idxthree} - \sum^{\ldim{\currlayer - 1}}_{\idxone \in C} \weightelem{\currlayer}{\idxone}{\idxthree} (\actelem{\currlayer - 1}{\idxone} - \refactelem{\currlayer - 1}{\idxone})\right) \\
				&\phantom{=(}- \relu\left(\actelem{n}{\idxthree} - \sum^{\ldim{\currlayer - 1}}_{\idxone \in C} \weightelem{\currlayer}{\idxone}{\idxthree} (\actelem{\currlayer - 1}{\idxone} - \refactelem{\currlayer - 1}{\idxone}) - \weightelem{\currlayer}{\idxtwo}{\idxthree} (\actelem{\currlayer - 1}{\idxtwo} - \refactelem{\currlayer - 1}{\idxtwo})\right) \\
				&> 0
			\end{align*}
		}
		The final inequality follows from $\actelem{n}{\idxthree} - \sum^{\ldim{\currlayer - 1}}_{\idxone \in C} \weightelem{\currlayer}{\idxone}{\idxthree} (\actelem{\currlayer - 1}{\idxone} - \refactelem{\currlayer - 1}{\idxone}) > 0$, $\weightelem{\currlayer}{\idxtwo}{\idxthree} (\actelem{\currlayer - 1}{\idxtwo} - \refactelem{\currlayer - 1}{\idxtwo}) > 0$ as well as ReLU being increasing for all inputs and strictly increasing for positive inputs.
		
		From the above, we can immediately see that:
		\begin{align*}
			\frac{
				\layer{\currlayer:\currlayer + 1}\left(\actvec{\currlayer - 1}\right)_\idxthree
				- \layer{\currlayer:\currlayer + 1}\left(\actvec{\currlayer - 1} \big| \remove(C)\right)_\idxthree
			}{
				\layer{\currlayer:\currlayer + 1}\left(\actvec{\currlayer - 1} \big| \remove(C)\right)_\idxthree - \layer{\currlayer:\currlayer + 1}\left(\actvec{\currlayer - 1} \big| \remove(C \cup C')\right)_\idxthree
			} < 0
		\end{align*}
		and thus $C$ counteracts $C'$ in a local internal conflict with respect to $\refactvec{\currlayer - 1}$. This conflict must necessarily be minimal, as an empty set can never be in conflict with another set, and since there is no $S \subset C$ for which $|\actelem{\currlayer}{\idxthree}| < \sum_{\idxone \in S} \left|\weightelem{\currlayer}{\idxone}{\idxthree} (\actelem{\currlayer - 1}{\idxone} - \refactelem{\currlayer - 1}{\idxone})\right|$ by assumption.
		
		To see that the implication also holds in the opposite direction, consider an arbitrary ReLU neuron with $\actelem{n+1}{\idxthree} = \relu(\actelem{\currlayer}{\idxthree}) = \relu(\allowbreak \sum^{\ldim{\currlayer - 1}}_\idxone \weightelem{\currlayer}{\idxone}{\idxthree} \actelem{\currlayer - 1}{\idxone})$, $\actelem{n}{\idxthree} \leq 0$, and two feature sets $C$ and $C'$, with $C$ counteracting $C'$ in a minimal local internal conflict with respect to $\refactvec{\currlayer - 1}$. We consider the following exhaustive cases:
		\begin{enumerate}
			\item[(i)] $C = \emptyset$ or $C' = \emptyset$. This contradicts our assumptions, as an empty set can never counteract another set, making this case impossible.
			\item[(ii)] $C \neq \emptyset$, $C' \neq \emptyset$, $\forall \idxone \in C, \weightelem{\currlayer}{\idxone}{\idxthree} (\actelem{\currlayer - 1}{\idxone} - \refactelem{\currlayer - 1}{\idxone}) < 0$ and $\forall \idxtwo \in C', \weightelem{\currlayer}{\idxtwo}{\idxthree} (\actelem{\currlayer - 1}{\idxtwo} - \refactelem{\currlayer - 1}{\idxtwo}) < 0$. Then,
			\begin{align*}
				&\layer{\currlayer:\currlayer + 1}\left(\actvec{\currlayer - 1}\right)_\idxthree - \layer{\currlayer:\currlayer + 1}\left(\actvec{\currlayer - 1} \big| \remove(C)\right)_\idxthree
				= \layer{\currlayer:\currlayer + 1}\left(\actvec{\currlayer - 1}\right)_\idxthree - \layer{\currlayer:\currlayer + 1}\left(\actvec{\currlayer - 1} \big| \left[\actelem{\currlayer - 1}{\idxone} := \refactelem{\currlayer}{\idxone}\right]_{\idxone \in C} \right)_\idxthree \\
				&= \relu\left(\actelem{\currlayer}{\idxthree}\right)
				- \relu\left(\actelem{n}{\idxthree} - \sum^{\ldim{\currlayer - 1}}_{\idxone \in C} \weightelem{\currlayer}{\idxone}{\idxthree} (\actelem{\currlayer - 1}{\idxone} - \refactelem{\currlayer - 1}{\idxone})\right) \leq 0
			\end{align*}
			with the last inequality holding due to $\sum^{\ldim{\currlayer - 1}}_{\idxone \in C} \weightelem{\currlayer}{\idxone}{\idxthree} (\actelem{\currlayer - 1}{\idxone} - \refactelem{\currlayer - 1}{\idxone}) < 0$, as well as ReLU being increasing for all inputs. Similarly,
			{\allowdisplaybreaks
				\begin{align*}
					&\layer{\currlayer:\currlayer + 1}\left(\actvec{\currlayer - 1} \big| \remove(C)\right)_\idxthree - \layer{\currlayer:\currlayer + 1}\left(\actvec{\currlayer - 1} \big| \remove(C \cup C')\right)_\idxthree \\
					&= \layer{\currlayer:\currlayer + 1}\left(\actvec{\currlayer - 1} \big| \left[\actelem{\currlayer - 1}{\idxone} := \refactelem{\currlayer}{\idxone}\right]_{\idxone \in C} \right)_\idxthree - \layer{\currlayer:\currlayer + 1}\left(\actvec{\currlayer - 1} \big| \left[\actelem{\currlayer - 1}{\idxone} := \refactelem{\currlayer}{\idxone}\right]_{\idxone \in C \cup C'}\right)_\idxthree \\
					&= \relu\left(\actelem{n}{\idxthree} - \sum^{\ldim{\currlayer - 1}}_{\idxone \in C} \weightelem{\currlayer}{\idxone}{\idxthree} (\actelem{\currlayer - 1}{\idxone} - \refactelem{\currlayer - 1}{\idxone})\right) \\
					&\phantom{=(}- \relu\left(\actelem{n}{\idxthree} - \sum^{\ldim{\currlayer - 1}}_{\idxone \in C} \weightelem{\currlayer}{\idxone}{\idxthree} (\actelem{\currlayer - 1}{\idxone} - \refactelem{\currlayer - 1}{\idxone}) - \sum^{\ldim{\currlayer - 1}}_{\idxtwo \in C'} \weightelem{\currlayer}{\idxtwo}{\idxthree} (\actelem{\currlayer - 1}{\idxtwo} - \refactelem{\currlayer - 1}{\idxtwo})\right) \\
					&\leq 0
				\end{align*}
			}
			with the last inequality holding due to $\sum^{\ldim{\currlayer - 1}}_{\idxtwo \in C} \weightelem{\currlayer}{\idxtwo}{\idxthree} (\actelem{\currlayer - 1}{\idxtwo} - \refactelem{\currlayer - 1}{\idxtwo}) < 0$, as well as ReLU being increasing for all inputs. The results above immediately give us that either
			\begin{align*}
				\frac{
					\layer{\currlayer:\currlayer + 1}\left(\actvec{\currlayer - 1}\right)_\idxthree
					- \layer{\currlayer:\currlayer + 1}\left(\actvec{\currlayer - 1} \big| \remove(C)\right)_\idxthree
				}{
					\layer{\currlayer:\currlayer + 1}\left(\actvec{\currlayer - 1} \big| \remove(C)\right)_\idxthree - \layer{\currlayer:\currlayer + 1}\left(\actvec{\currlayer - 1} \big| \remove(C \cup C')\right)_\idxthree
				} \geq 0
			\end{align*}
			or the ratio is undefined (with $0$ in the denominator). This is a contradiction, since by Definition \ref{def:counteractive-features}, $C$ cannot counteract $C'$ with respect to $\refactvec{\currlayer - 1}$ unless the ratio is negative, and thus this case cannot arise given the assumptions.
			\item[(iii)] $C \neq \emptyset$, $C' \neq \emptyset$, $\forall \idxone \in C, \weightelem{\currlayer}{\idxone}{\idxthree} (\actelem{\currlayer - 1}{\idxone} - \refactelem{\currlayer - 1}{\idxone}) > 0$. This case also cannot arise by similar reasoning as for the previous case, with $\layer{\currlayer:\currlayer + 1}\left(\actvec{\currlayer - 1}\right)_\idxthree - \layer{\currlayer:\currlayer + 1}\left(\actvec{\currlayer - 1} \big| \remove(C)\right)_\idxthree = 0$ and the ratio from Definition \ref{def:counteractive-features} either being zero or undefined (if there is $0$ in the denominator). The equality $\layer{\currlayer:\currlayer + 1}\left(\actvec{\currlayer - 1}\right)_\idxthree - \layer{\currlayer:\currlayer + 1}\left(\actvec{\currlayer - 1} \big| \remove(C)\right)_\idxthree = 0$ holds due to $\actelem{n}{\idxthree} \leq 0$, $\sum^{\ldim{\currlayer - 1}}_{\idxtwo \in C} \weightelem{\currlayer}{\idxtwo}{\idxthree} (\actelem{\currlayer - 1}{\idxtwo} - \refactelem{\currlayer - 1}{\idxtwo}) > 0$ and $\relu(x) = 0$ for $x \leq 0$.
			
			\item[(iv)] $\exists \idxone \in C, \weightelem{\currlayer}{\idxone}{\idxthree} (\actelem{\currlayer - 1}{\idxone} - \refactelem{\currlayer - 1}{\idxone}) < 0$ and $\exists \idxtwo \in C',\weightelem{\currlayer}{\idxtwo}{\idxthree} (\actelem{\currlayer - 1}{\idxtwo} - \refactelem{\currlayer - 1}{\idxtwo}) > 0$. We will first show that $\forall \idxone \in C, \weightelem{\currlayer}{\idxone}{\idxthree} (\actelem{\currlayer - 1}{\idxone} - \refactelem{\currlayer - 1}{\idxone}) < 0$. Assume towards a contradiction that $\exists \idxfive \in C, \weightelem{\currlayer}{\idxfive}{\idxthree} (\actelem{\currlayer - 1}{\idxfive} - \refactelem{\currlayer - 1}{\idxfive}) \geq 0$, with $\idxfive$ being the witness to this assumption. Then,
			\begin{align*}
				&\layer{\currlayer:\currlayer + 1}\left(\actvec{\currlayer - 1}\right)_\idxthree - \layer{\currlayer:\currlayer + 1}\left(\actvec{\currlayer - 1} \big| \remove(C)\right)_\idxthree
				= \layer{\currlayer:\currlayer + 1}\left(\actvec{\currlayer - 1}\right)_\idxthree - \layer{\currlayer:\currlayer + 1}\left(\actvec{\currlayer - 1} \big| \left[\actelem{\currlayer - 1}{\idxone} := \refactelem{\currlayer}{\idxone}\right]_{\idxone \in C} \right)_\idxthree \\
				&= \relu\left(\actelem{\currlayer}{\idxthree}\right)
				- \relu\left(\actelem{n}{\idxthree} - \sum^{\ldim{\currlayer - 1}}_{\idxone \in C, \idxone \neq \idxfive} \weightelem{\currlayer}{\idxone}{\idxthree} (\actelem{\currlayer - 1}{\idxone} - \refactelem{\currlayer - 1}{\idxone}) - \weightelem{\currlayer}{\idxfive}{\idxthree} (\actelem{\currlayer - 1}{\idxfive} - \refactelem{\currlayer - 1}{\idxfive})\right) < 0
			\end{align*}
			Note that the above inequality must hold as, by assumption, $C$ counteracts $C'$. Thus, by Definition \ref{def:counteractive-features}, the difference above must be non-zero. Since $\relu(\refactelem{\currlayer}{\idxthree}) = 0$ and since $0$ is the minimum possible value of ReLU, the difference must necessarily be negative. If this is the case, then it must also hold that $\layer{\currlayer:\currlayer + 1}\left(\actvec{\currlayer - 1}\right)_\idxthree - \layer{\currlayer:\currlayer + 1}\left(\actvec{\currlayer - 1} \big| \remove(C \setminus \{\idxfive\})\right)_\idxthree < 0$ since $\weightelem{\currlayer}{\idxfive}{\idxthree} (\actelem{\currlayer - 1}{\idxfive} - \refactelem{\currlayer - 1}{\idxfive}) \geq 0$. Similarly,
			{\allowdisplaybreaks
				\begin{align*}
					&\layer{\currlayer:\currlayer + 1}\left(\actvec{\currlayer - 1} \big| \remove(C)\right)_\idxthree - \layer{\currlayer:\currlayer + 1}\left(\actvec{\currlayer - 1} \big| \remove(C \cup C')\right)_\idxthree \\
					&= \layer{\currlayer:\currlayer + 1}\left(\actvec{\currlayer - 1} \big| \left[\actelem{\currlayer - 1}{\idxone} := \refactelem{\currlayer}{\idxone}\right]_{\idxone \in C} \right)_\idxthree - \layer{\currlayer:\currlayer + 1}\left(\actvec{\currlayer - 1} \big| \left[\actelem{\currlayer - 1}{\idxone} := \refactelem{\currlayer}{\idxone}\right]_{\idxone \in C \cup C'}\right)_\idxthree \\
					&= \relu\left(\actelem{n}{\idxthree} - \sum^{\ldim{\currlayer - 1}}_{\idxone \in C, \idxone \neq \idxfive} \weightelem{\currlayer}{\idxone}{\idxthree} (\actelem{\currlayer - 1}{\idxone} - \refactelem{\currlayer - 1}{\idxone}) - \weightelem{\currlayer}{\idxfive}{\idxthree} (\actelem{\currlayer - 1}{\idxfive} - \refactelem{\currlayer - 1}{\idxfive})\right) \\
					&\phantom{=(}- \relu\left(\actelem{n}{\idxthree} - \sum^{\ldim{\currlayer - 1}}_{\idxone \in C, \idxone \neq \idxfive} \weightelem{\currlayer}{\idxone}{\idxthree} (\actelem{\currlayer - 1}{\idxone} - \refactelem{\currlayer - 1}{\idxone}) - \weightelem{\currlayer}{\idxfive}{\idxthree} (\actelem{\currlayer - 1}{\idxfive} - \refactelem{\currlayer - 1}{\idxfive}) - \sum^{\ldim{\currlayer - 1}}_{\idxtwo \in C'} \weightelem{\currlayer}{\idxtwo}{\idxthree} (\actelem{\currlayer - 1}{\idxtwo} - \refactelem{\currlayer - 1}{\idxtwo})\right) \\
					&> 0
				\end{align*}
			}
			The inequality above again follows from $C$ counteracting $C'$ and the Definition \ref{def:counteractive-features}. This also means that $\layer{\currlayer:\currlayer + 1}\left(\actvec{\currlayer - 1} \big| \remove(C \setminus \{\idxfive\})\right)_\idxthree - \layer{\currlayer:\currlayer + 1}\left(\actvec{\currlayer - 1} \big| \remove(C \setminus \{\idxfive\} \cup C' \setminus \{\idxfive\})\right)_\idxthree > 0$, as $\weightelem{\currlayer}{\idxfive}{\idxthree} (\actelem{\currlayer - 1}{\idxfive} - \refactelem{\currlayer - 1}{\idxfive}) \geq 0$ (note that $\relu(x - y) - \relu(x' - y) > 0$ with $y \geq 0$ necessarily implies $\relu(x) - \relu(x') > 0$ for any $x, x' \in \mathbb{R}$). By these results, we have:
			\begin{align*}
				\frac{
					\layer{\currlayer:\currlayer + 1}\left(\actvec{\currlayer - 1}\right)_\idxthree
					- \layer{\currlayer:\currlayer + 1}\left(\actvec{\currlayer - 1} \big| \remove(C \setminus \{\idxfive\})\right)_\idxthree
				}{
					\layer{\currlayer:\currlayer + 1}\left(\actvec{\currlayer - 1} \big| \remove(C \setminus \{\idxfive\})\right)_\idxthree - \layer{\currlayer:\currlayer + 1}\left(\actvec{\currlayer - 1} \big| \remove(C \setminus \{\idxfive\} \cup C' \setminus \{\idxfive\})\right)_\idxthree
				} < 0
			\end{align*}
			But then $C \setminus \{\idxfive\}$ counteracts $C' \setminus \{\idxfive\}$ with respect to $\refactvec{\currlayer - 1}$. Since $C \setminus \{\idxfive\} \subset C$ (as $\idxfive$ must be in C), this contradicts our original assumption that $C$ counteracts $C'$ in a minimal local internal conflict with respect to $\refactvec{\currlayer - 1}$. Thus, $\forall \idxone \in C, \weightelem{\currlayer}{\idxone}{\idxthree} (\actelem{\currlayer - 1}{\idxone} - \refactelem{\currlayer - 1}{\idxone}) < 0$, as required.
			
			Next, we show that $|\actelem{\currlayer}{\idxthree}| < \sum_{\idxone \in C} \left|\weightelem{\currlayer}{\idxone}{\idxthree} (\actelem{\currlayer - 1}{\idxone} - \refactelem{\currlayer - 1}{\idxone})\right|$. Assume towards a contradiction that $|\actelem{\currlayer}{\idxthree}| \geq \sum_{\idxone \in C} \left|\weightelem{\currlayer}{\idxone}{\idxthree} (\actelem{\currlayer - 1}{\idxone} - \refactelem{\currlayer - 1}{\idxone})\right|$. Then,
			\begin{align*}
				&\layer{\currlayer:\currlayer + 1}\left(\actvec{\currlayer - 1}\right)_\idxthree - \layer{\currlayer:\currlayer + 1}\left(\actvec{\currlayer - 1} \big| \remove(C)\right)_\idxthree
				= \layer{\currlayer:\currlayer + 1}\left(\actvec{\currlayer - 1}\right)_\idxthree - \layer{\currlayer:\currlayer + 1}\left(\actvec{\currlayer - 1} \big| \left[\actelem{\currlayer - 1}{\idxone} := \refactelem{\currlayer}{\idxone}\right]_{\idxone \in C} \right)_\idxthree \\
				&= \relu\left(\actelem{\currlayer}{\idxthree}\right)
				- \relu\left(\actelem{n}{\idxthree} - \sum^{\ldim{\currlayer - 1}}_{\idxone \in C} \weightelem{\currlayer}{\idxone}{\idxthree} (\actelem{\currlayer - 1}{\idxone} - \refactelem{\currlayer - 1}{\idxone})\right) = 0 - 0 = 0
			\end{align*}
			This is due to $\actelem{\currlayer}{\idxthree} \leq 0$ and $\actelem{n}{\idxthree} - \sum^{\ldim{\currlayer - 1}}_{\idxone \in C} \weightelem{\currlayer}{\idxone}{\idxthree} (\actelem{\currlayer - 1}{\idxone} - \refactelem{\currlayer - 1}{\idxone}) < 0$, with the latter following from $|\actelem{\currlayer}{\idxthree}| \geq \sum_{\idxone \in C} \Bigl|\weightelem{\currlayer}{\idxone}{\idxthree} (\actelem{\currlayer - 1}{\idxone} - \refactelem{\currlayer - 1}{\idxone})\Bigr|$ and $\forall \idxone \in C, \weightelem{\currlayer}{\idxone}{\idxthree} (\actelem{\currlayer - 1}{\idxone} - \refactelem{\currlayer - 1}{\idxone}) < 0$. This is a contradiction, as, according to Definition \ref{def:counteractive-features}, $C$ couldn't counteract $C'$ in that case. Therefore, $|\actelem{\currlayer}{\idxthree}| < \sum_{\idxone \in C} \left|\weightelem{\currlayer}{\idxone}{\idxthree} (\actelem{\currlayer - 1}{\idxone} - \refactelem{\currlayer - 1}{\idxone})\right|$.
			
			Now, consider a set $S' = \{\idxtwo\}$ with $\idxtwo$ being the witness to the second case assumption. By what we have shown when proving the first direction of the implication, $C$ must necessarily counteract $S'$ in a local internal conflict with respect to $\refactvec{\currlayer - 1}$. In order for this conflict to be minimal, there must be no $S \subset C$ for which $|\actelem{\currlayer}{\idxthree}| < \sum_{\idxone \in S} \left|\weightelem{\currlayer}{\idxone}{\idxthree} (\actelem{\currlayer - 1}{\idxone} - \refactelem{\currlayer - 1}{\idxone})\right|$ (otherwise such $S$ would counteract $S'$, breaking minimality) and $C' = S'$. Thus, we have shown that all the conditions from Proposition \ref{prop:conflict-relu-negative} must hold, as required.
		\end{enumerate}
		This concludes the proof. \aptLtoX[graphic=no,type=html]{&#x25A1;}{{$\square$}}
	\end{myproof}
	
	\section{CAFE Motivating Examples}
	\label{apd:cafe-motivating-examples}
	To further motivate the importance of surfacing conflicts in feature attribution methods and to exemplify some of the other issues addressed by CAFE we provide several additional examples.
	
\begin{figure}[tb]
		\centering
		\includegraphics[width=0.26\textwidth]{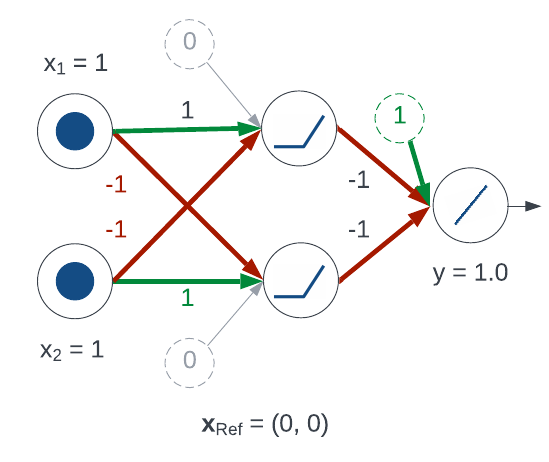}
		\caption{Visualisation of a simple neural network (NN) modelling the XNOR function, with two conflicting inputs ($x_1 = 1$, $x_2 = 1$) and a positive bias ($1$ in the green dashed circle) that are not reliably captured by standard feature attribution methods. While existing gradient-based methods return zero feature attributions, CAFE (1.0) computes positive and negative scores 1$^+$/1$^-$ for each input and 1$^+$/0$^-$ for the bias, which more closely reflect the internal reasoning process of the model. See Example \ref{ex:xnor-conflict} for details.}
		\label{fig:example-cancelation-xnor-nn}
	\end{figure}

	As we discussed in the main text, feature attribution methods often fail to unearth conflicts and do not surface the influence of model biases, as we demonstrate on the following example:
	\begin{example}[XNOR Conflict]
		\label{ex:xnor-conflict}
		Consider the neural network (NN) in Figure \ref{fig:example-cancelation-xnor-nn}, computing the binary XNOR function (which is $0$ if exactly one of the two binary inputs is $1$ and $1$ otherwise), and the input $\inputvec = (1, 1)$\footnote{In the figure, as in all examples and experiments, we also consider a reference input ($\refinputvec = (0, 0)$ in the figure), but disregard it for methods not using such an input (e.g., Gradient $\cdot$ Input, LRP).}. By inspection of the weights and neuron activations, it is clear that the input features are in conflict, pushing the pre-activations of both hidden neurons to zero values for which the ReLU function is inactive. This causes all gradient-based methods (in particular, G $\cdot$ I, LRP, DL-R and IG) to return zero attribution scores for both input features, thus failing to surface that the features were considered but then canceled out. Additionally, since the input features cancel each other symmetrically and since the NN output is driven by the $+1$ bias of the output neuron, even non-gradient-based methods that can partially handle conflicts (notably DeepLIFT RevealCancel) return zeros. The prior methods also do not return any attributions for the bias.
	\end{example}
	
	Another desirable property for feature attribution methods is robustness to small changes to the explained inputs, provided that they do not result in significant changes to the output. As we show on the example below, the inability to reliably surface conflicts also negatively affects attribution robustness.
	
\begin{figure*}[!tb]
		\centering
		\begin{subfigure}[m]{0.32\textwidth}
			\centering
			\includegraphics[width=0.87\textwidth]{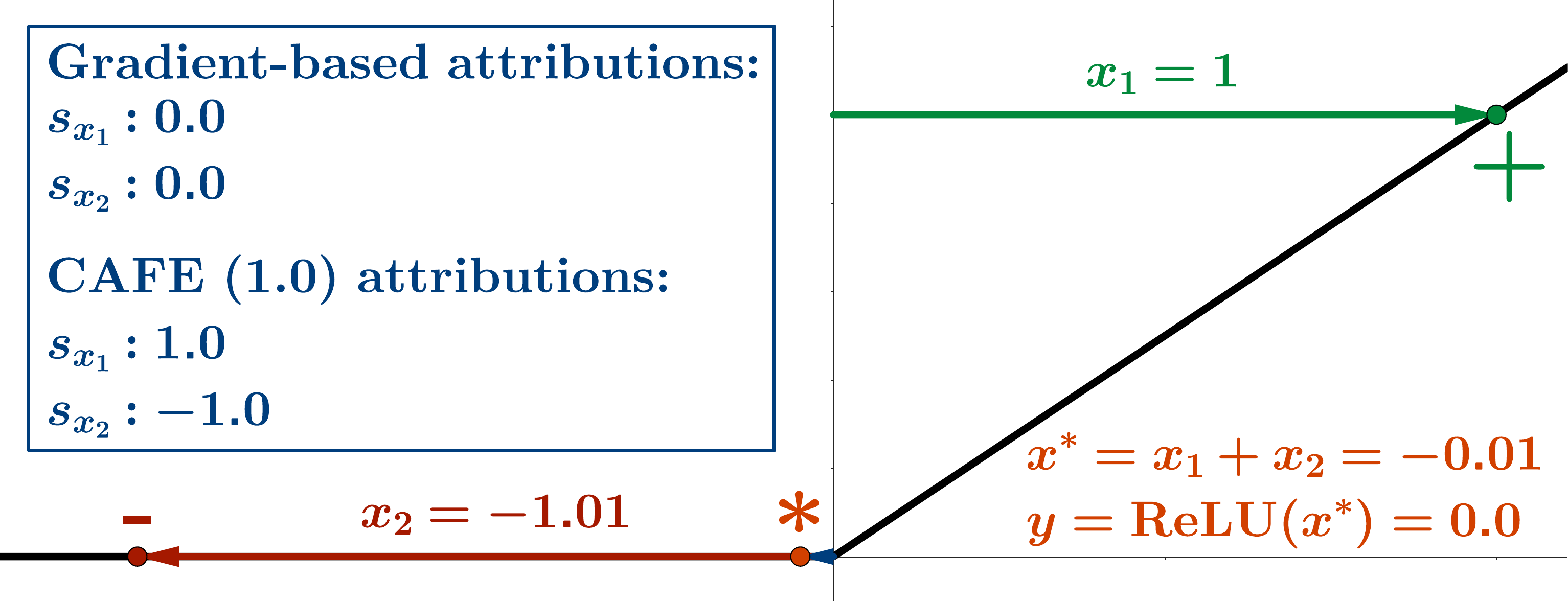}
			\\ \vspace{3pt}
			\includegraphics[width=0.87\textwidth]{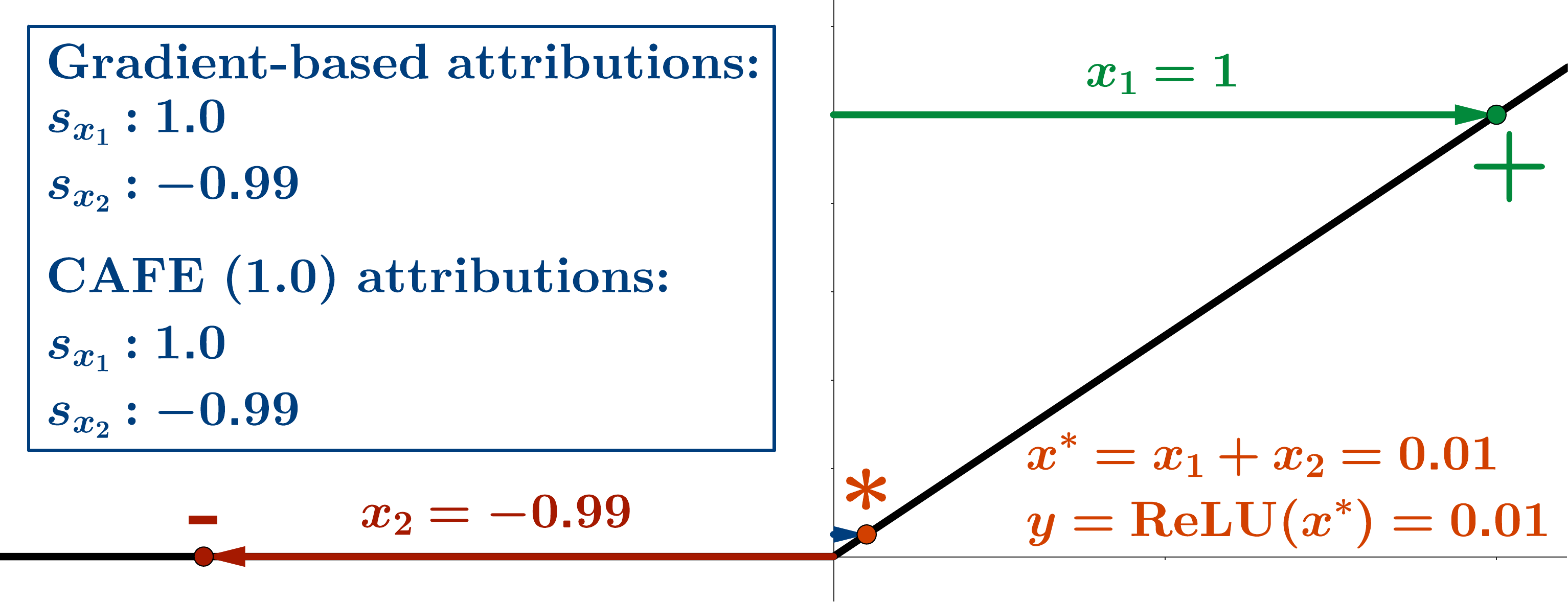}
			\caption{ReLU Attribution Discontinuity}
			\label{fig:example-relu-discontinuity}
		\end{subfigure}
		\hfill
		\begin{subfigure}[m]{0.17\linewidth}
			\centering
			\includegraphics[width=\textwidth]{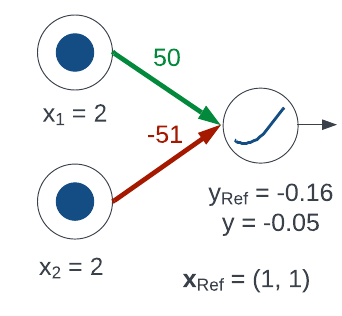}
			\caption{GELU NN}
			\label{fig:example-explosion-gelu}
			\vspace{7pt}
		\end{subfigure}
		\hfill
		\begin{subfigure}[m]{0.4\textwidth}
			\scriptsize
			\centering
			\begin{center}
				\begin{tabular}{ c c c }
					\toprule
					\multirow{2}{*}{\textbf{Method}} & \multicolumn{2}{c}{\textbf{Attribution Scores}} \\
					\cmidrule(r){2-3}
					& \textbf{$\inputelem{1}$} & \textbf{$\inputelem{2}$} \\
					\midrule
					Gradient $\cdot$ Input \warning & -8.52 & 8.69 \\
					LRP & 2.28 & -2.32 \\
					DeepLIFT Rescale \warning & -5.66 & 5.77 \\
					DeepLIFT RevealCancel & 24.56 & -24.44 \\
					Integrated Gradients \warning & -5.66 & 5.77 \\
					\midrule
					CAFE (0.0) & 0.0$^+$/0.05$^-$ & 0.16$^+$/0.0$^-$ \\
					CAFE (1.0) & 49.16$^+$/0.05$^-$ & 0.16$^+$/49.16$^-$ \\
					\bottomrule
				\end{tabular}
			\end{center}
			\caption{
				\label{fig:geluscores}Attribution scores returned by the different attribution methods for the GELU NN in Figure \ref{fig:example-explosion-gelu}.}
			\label{tab:example-explosion-gelu-scores}
		\end{subfigure}
		\caption{Illustration of explanation robustness issues addressed by CAFE. \textbf{(a)} depicts two minimally different inputs to a ReLU neuron that receive markedly different attribution scores from gradient-based methods, as detailed in Example \ref{ex:attribution-score-discontinuity}. Meanwhile, the analogical explanations computed by CAFE (1.0) are almost identical. \textbf{(b)} shows a GELU unit with two conflicting inputs ($x_1 = 2$, $x_2 = 2$) and \textbf{(c)} the associated feature attribution scores. Baseline methods marked with \warning{} experience the attribution score explosion issue, significantly overestimating the possible effects of input features on the output. LRP and DeepLIFT RevealCancel do not exhibit this issue here, but their scores still do not fully capture conflicting features. CAFE is the only method that can be directed to either isolate the behaviour of the input features on the active negative range of GELU (CAFE (0.0)) or capture their full effect on the output when considering their conflict (CAFE (1.0)). See Example \ref{ex:attribution-score-explosion} for details.}
		\label{fig:examples-issues}
	\end{figure*}

	\begin{example}[Attribution Score Discontinuity]
		\label{ex:attribution-score-discontinuity}
		Consider a neuron computing the function $y = \relu(x_1 + x_2)$, and the two scenarios depicted in Figure \ref{fig:example-relu-discontinuity}. Although the change in the inputs (from $x_2 = -1.01$ to $x_2 = -0.99$, with $x_1$ remaining unchanged) and the output (from $y = 0.0$ to $y = 0.01$) is minimal between the two scenarios, the feature attributions produced by gradient-based methods change dramatically.
	\end{example}
	
	Feature attribution methods should also gracefully handle variations in  model activations and, by extension, its gradients, especially since NNs and their gradients can be highly irregular \cite{balduzzi-shattered}. Instead, many existing feature attribution methods are prone to what we call ``attribution score explosion", which can cause the scores to become unreasonably low or high, beyond the model's actual output range. We demonstrate this issue on the following example:
	
	\begin{example}[Attribution Score Explosion]
		\label{ex:attribution-score-explosion}
		Consider the NN in Figure \ref{fig:example-explosion-gelu}, input $\inputvec = (2, 2)$ and reference input $\refinputvec = (1, 1)$. Figure~\ref{fig:geluscores} shows that all methods except LRP and DL-RC assign large negative scores to $\inputelem{1}$ and large positive scores to $\inputelem{2}$. While these scores somewhat capture the local behaviour of the GELU activation function, they significantly overestimate the effects of each feature. Since GELU flattens and tends to $0$ as its input becomes increasingly negative, the highest amount by which $\inputelem{2}$ as a negative feature can increase the output from the reference value $y_{\reference} = -0.16$ is $0.16$. The $8.69 >> 0.16$ and $5.77 >> 0.16$ positive scores returned by $\inputelem{2}$ by G $\cdot$ I, LRP and DL-R are thus unreasonably high. Meanwhile, although the scores from LRP and DL-RC show that $\inputelem{2}$ could have a negative effect on the output, they only illustrate fractions of this potential effect and also mask the actual small positive effect of $\inputelem{2}$ in the negative range of GELU. Instead, it may be useful to capture both positive and negative effects of features, and to be able to control the degree to which the possible, canceled features' effects are reflected in the attribution scores.
	\end{example}
	
	CAFE aims to rectify these issues by taking a principled approach to handling conflicts while also explicitly incorporating the effects of biases.
	
	\section{Score Flow Definitions}
	\label{apd:score-flow-definitions}
	The full definitions for the peak and linear score flows are provided in Definitions \ref{def:peak-effects-full} and \ref{def:linear-effects-full}.
	
	\begin{figure*}[!htp]
		\begin{minipage}{\textwidth}
			\begin{definition}[Peak Score Flows]
				\label{def:peak-effects-full}
				The peak score flows at the activation layer $\layer{\currlayer + 1}$ are defined as follows:
				\normalfont
				\begin{align*}
					\vpeak{\currlayer + 1}{\poss}{\poss} &= \indvector{\currlayer}{\poss} \odot \clip(\simpleactfun(\oinputs{\currlayer + 1}{\combs}) - \simpleactfun(\refactvec{\currlayer})) +\max(\clip(\simpleactfun(\oinputs{\currlayer + 1}{\poss}) - \simpleactfun(\indvector{\currlayer}{\poss} \odot \oinputs{\currlayer + 1}{\combs} + \indvector{\currlayer}{\negs} \odot \refactvec{\currlayer})), \\
					&\phantom{=ssss}\clip(\simpleactfun(\indvector{\currlayer}{\poss} \odot \refactvec{\currlayer} + \indvector{\currlayer}{\negs} \odot \oinputs{\currlayer + 1}{\combs}) - \simpleactfun(\oinputs{\currlayer + 1}{\negs}))) \displaybreak[0] \\
					\vpeak{\currlayer + 1}{\negs}{\poss} &= \indvector{\currlayer}{\negs} \odot \clip(\simpleactfun(\oinputs{\currlayer + 1}{\combs}) - \simpleactfun(\refactvec{\currlayer})) +\max(\clip(\simpleactfun(\indvector{\currlayer}{\poss} \odot \oinputs{\currlayer + 1}{\combs} + \indvector{\currlayer}{\negs} \odot \refactvec{\currlayer}) - \simpleactfun(\oinputs{\currlayer + 1}{\poss})), \\
					&\phantom{=ssss}\clip(\simpleactfun(\oinputs{\currlayer + 1}{\negs}) - \simpleactfun(\indvector{\currlayer}{\poss} \odot \refactvec{\currlayer} + \indvector{\currlayer}{\negs} \odot \oinputs{\currlayer + 1}{\combs}))) \displaybreak[0] \\
					\vpeak{\currlayer + 1}{\poss}{\negs} &= \indvector{\currlayer}{\poss} \odot \clip(\simpleactfun(\refactvec{\currlayer}) - \simpleactfun(\oinputs{\currlayer + 1}{\combs})) +\max(\clip(\simpleactfun(\indvector{\currlayer}{\poss} \odot \oinputs{\currlayer + 1}{\combs} + \indvector{\currlayer}{\negs} \odot \refactvec{\currlayer}) - \simpleactfun(\oinputs{\currlayer + 1}{\poss})), \\
					&\phantom{=ssss}\clip(\simpleactfun(\oinputs{\currlayer + 1}{\negs}) - \simpleactfun(\indvector{\currlayer}{\poss} \odot \refactvec{\currlayer} + \indvector{\currlayer}{\negs} \odot \oinputs{\currlayer + 1}{\combs}))) \displaybreak[0] \\
					\vpeak{\currlayer + 1}{\negs}{\negs} &= \indvector{\currlayer}{\negs} \odot \clip(\simpleactfun(\refactvec{\currlayer}) - \simpleactfun(\oinputs{\currlayer + 1}{\combs})) +\max(\clip(\simpleactfun(\oinputs{\currlayer + 1}{\poss}) - \simpleactfun(\indvector{\currlayer}{\poss} \odot \oinputs{\currlayer + 1}{\combs} + \indvector{\currlayer}{\negs} \odot \refactvec{\currlayer})), \\
					&\phantom{=ssss}\clip(\simpleactfun(\indvector{\currlayer}{\poss} \odot \refactvec{\currlayer} + \indvector{\currlayer}{\negs} \odot \oinputs{\currlayer + 1}{\combs}) - \simpleactfun(\oinputs{\currlayer + 1}{\negs}))) \displaybreak[0]
				\end{align*}
			\end{definition}
		\end{minipage}
	\end{figure*}

	\begin{figure*}[!htp]
		\begin{minipage}{\textwidth}
			\begin{definition}[Linear Score Flows]
				\label{def:linear-effects-full}
				The peak score flows at the activation layer $\layer{\currlayer + 1}$ are defined as follows:
				\normalfont
				\begin{align*}
					\vlinear{\currlayer + 1}{\poss}{\poss} &= \min \left( \scoresvecn{\currlayer}{\poss} \odot \frac{\clip \left( \text{sgn}(\scoresvecn{\currlayer}{\combs}) \odot (\phi(\oinputs{\currlayer + 1}{\combs}) - \phi(\refactvec{\currlayer})) \right)}{|\scoresvecn{\currlayer}{\combs}| + \epsilon}, \vpeak{\currlayer + 1}{\poss}{\poss} \right) \\
					\vlinear{\currlayer + 1}{\negs}{\poss} &= \min \left( \scoresvecn{\currlayer}{\negs} \odot \frac{\clip \left( \text{sgn}(\scoresvecn{\currlayer}{\combs}) \odot (\phi(\refactvec{\currlayer}) - \phi(\oinputs{\currlayer + 1}{\combs})) \right)}{|\scoresvecn{\currlayer}{\combs}| + \epsilon}, \vpeak{\currlayer + 1}{\negs}{\poss} \right) \\
					\vlinear{\currlayer + 1}{\poss}{\negs} &= \min \left( \scoresvecn{\currlayer}{\poss} \odot \frac{\clip \left( \text{sgn}(\scoresvecn{\currlayer}{\combs}) \odot (\phi(\refactvec{\currlayer}) - \phi(\oinputs{\currlayer + 1}{\combs})) \right)}{|\scoresvecn{\currlayer}{\combs}| + \epsilon}, \vpeak{\currlayer + 1}{\poss}{\negs} \right) \\
					\vlinear{\currlayer + 1}{\negs}{\negs} &= \min \left( \scoresvecn{\currlayer}{\negs} \odot \frac{\clip \left( \text{sgn}(\scoresvecn{\currlayer}{\combs}) \odot (\phi(\oinputs{\currlayer + 1}{\combs}) - \phi(\refactvec{\currlayer})) \right)}{|\scoresvecn{\currlayer}{\combs}| + \epsilon}, \vpeak{\currlayer + 1}{\negs}{\negs} \right) \\
				\end{align*}
				where sgn is the element-wise sign function and $\epsilon$ is a small positive stabiliser enforcing the behaviour $\frac{0}{0} = 0$.
			\end{definition}
		\end{minipage}
	\end{figure*}

	\section{Notes on CAFE Notions}
	\label{sec:notes-cafe-notions}
	In this section, we provide brief descriptions of the key notions that we introduced or used in the main paper along with additional notes.
	
	\paragraph{Reference Input} The reference input ($\refinputvec$) should be a context-dependent neutral value with respect to which the attribution scores should be computed. In the context of tabular data, the commonly used choices for a reference input include zeroes or mean/median values of the features. We discuss the choice of the reference input in more detail in the FAQs in Appendix \ref{sec:faq}.
	
	\paragraph{Reference (Pre-)Activations} The reference (pre-)activations are the (pre-)activation values of neurons obtained by applying the explained model with all biases ablated ($\model_{\reference}$) to the reference input ($\refinputvec$).
	
	\paragraph{Attribution Scores} Scores estimating the positive or negative effects of a feature or the network biases on a particular target neuron. Without additional context, we typically mean the attribution scores computed with respect to the sole output neuron of the network (or, in the case of multi-class classification, the output neuron corresponding to the class predicted with the highest likelihood).
	
	\paragraph{Overall Inputs} The different variants of the overall inputs capture the possible inputs to an activation layer when considering the reference inputs joined with (i) combined positive and negative contributions from the features and biases $\oinputs{\currlayer + 1}{\combs} = \refactvec{\currlayer} + \scoresvecn{\currlayer}{\combs}$, (ii) only the positive contributions $\oinputs{\currlayer + 1}{\poss} = \refactvec{\currlayer} + \scoresvecn{\currlayer}{\poss}$, and (iii) only the negative contributions $\oinputs{\currlayer + 1}{\negs} = \refactvec{\currlayer} + \scoresvecn{\currlayer}{\negs}$.
	
	\paragraph{Peak Score Flows} As explained in the main paper, the primary purpose of the peak score flows is to establish the maximum impact of neuron's positive/negative inputs on its positive/negative outputs. This includes both actual observed changes in the neuron's output with respect to the reference and canceled changes to the output that arise due to the conflicting features.
	
	\paragraph{Linear Score Flows} Linear score flows are a notion similar to peak score flows, with the difference that linear score flows are only supposed to capture the behaviour of the neuron on its currently ``active" interval, i.e., the interval between the reference pre-activation and the pre-activation corresponding to the combined inputs $\oinputs{\currlayer + 1}{\combs}$. The linear score flows are clipped if they exceed the value of peak score flows, preventing attribution score explosion.
	
	\paragraph{Attribution Multipliers} These are multipliers specifying the coefficients to use when propagating scores through activation layers. Depending on the value of the cancelation sensitivity constant for the given layer, they control the degree to which the canceled effects of conflicting features at the given layer should be reflected in the propagated scores. The attribution multipliers are treated as constants during the CAFE backward pass.
	
	\section{Adapting CAFE to FT-Transformer Models}
	\label{apd:adapting-cafe-ft-transformer}
	In this section, we provide the details on adapting CAFE to the components of FT-Transformer models, which requires handling propagation of the scores through the feature tokenization procedures, as well as the residual, layer normalization and multi-head attention layers. Feature tokenization can be handled in a way analogical to the standard input layer, with the scores for the embedding elements being summed together to obtain the attributions for the given feature. The special \texttt{[CLS]} embedding prepended during the tokenization is treated as a bias vector given its independence on the input features.
	
	Residual connections can be handled by propagating the scores through the associated block using the standard forward rules and summing them with the original scores at the block input. Multi-head attention layers can be effectively linearised using the findings from \cite{elhage-mathematical-circits} and treating the computed attention patterns as constant linear weights (this ensures that the attribution scores ignore the behaviour of the Query-Key transformer circuit, which is mainly responsible for determining how information is routed through the model, and instead focuses them on the behaviour of the Output-Value circuit, which determines the output of the attention layer). Layer normalization can be linearised in a similar way, by treating the sample mean and variance as constants.
	
	\section{CAFE Properties and Proofs}
	\label{sec:cafe-properties-proofs}
	In this section of the appendix, we provide detailed definitions and proofs of the theoretical properties introduced in Section \ref{sec:theoretical-evaluation} of the main text.
	
	\subsection{Missingness}
	In this section, we show that CAFE satisfies the ``missingness" property, which was previously considered by \cite{lundberg-shap} (though their version did not include reference inputs):
	
	\begin{property}[Missingness]
		A feature attribution satisfies the missingness property iff the attribution method always assigns a score of $0$ to features:
		\begin{enumerate}
			\item[(i)] that are $0$, if the feature attribution method does not use a reference input.
			\item[(ii)] that are equal to the reference input, if the feature attribution method uses a reference input.
		\end{enumerate}
	\end{property}
	
	\begin{proposition}
		\label{prop:cafe-missingness}
		CAFE satisfies the missingness property for any choice of the cancelation sensitivity constant $\cancelconst{}$.
	\end{proposition}
	
	\begin{myproof}
		Take an arbitrary index $\idxone$ and assume that $\inputelem{\idxone} = \refinputelem{\idxone}$. Then, by Definition \ref{def:input-layer-rule}, $\vfoundationelem{\idxone} = \inputelem{\idxone} - \refinputelem{\idxone} = 0$. By Definition \ref{def:backward-attribution-scores}, $\scoresvecelem{\poss}{\idxone} = \frac{\partial \scoresvecnelem{\nlayers}{\poss}{\targetneuron}}{\partial \vfoundationelem{\idxone}} \cdot \vfoundationelem{\idxone} = \frac{\partial \scoresvecnelem{\nlayers}{\poss}{\targetneuron}}{\partial \vfoundationelem{\idxone}} \cdot 0 = 0$ and $\scoresvecelem{\negs}{\idxone} = \frac{\partial \scoresvecnelem{\nlayers}{\negs}{\targetneuron}}{\partial \vfoundationelem{\idxone}} \cdot \vfoundationelem{\idxone} = \frac{\partial \scoresvecnelem{\nlayers}{\negs}{\targetneuron}}{\partial \vfoundationelem{\idxone}} \cdot 0 = 0$, as required. \aptLtoX[graphic=no,type=html]{&#x25A1;}{{$\square$}}
	\end{myproof}
	
	\subsection{Linearity}
	As our next formal property, we consider linearity, which has been previously proposed as a desirable axiom for the Integrated Gradients method \cite{sundararajan-integrated-gradients}:
	
	\begin{property}[Linearity]
		A feature attribution method is linear iff, for any two models $\model_1$ and $\model_2$, coefficients $a$ and $b$, a combined model $\model_3(\inputvec) \triangleq a \cdot \model_1(\inputvec) + b \cdot \model_2(\inputvec)$, current input $\inputvec$ and reference input $\refinputvec$ (if the attribution method takes a reference input), the feature attribution scores for $\model_3$ and input $\inputvec$ with respect to the reference input $\refinputvec$ are equal to $\scoresvec_3 = a \cdot \scoresvec_1 + b \cdot \scoresvec_2$, where $\scoresvec_1$ and $\scoresvec_2$ are the attribution scores computed for the input $\inputvec$, reference input $\refinputvec$ and the models $\model_1$ and $\model_2$, respectively.
	\end{property}
	
	\begin{proposition}
		\label{prop:cafe-linearity}
		CAFE is linear for any choice of $\cancelconst{}$, considering the joint attribution scores $\scoresvec^\combs = \scoresvec^\poss - \scoresvec^\negs$.
	\end{proposition}
	
	We restrict Proposition \ref{prop:cafe-linearity} to the joint feature attribution scores, as the linearity property was originally proposed for methods returning a single score for each feature. The corresponding proof is given in Proof \ref{proof:cafe-linearity}.
	
	\begin{myproof}
		\label{proof:cafe-linearity}
		Take arbitrary models $\model_1$ and $\model_2$, coefficients $a$ and $b$, a combined model such that $\model_3(\inputvec) \triangleq a \cdot \model_1(\inputvec) + b \cdot \model_2(\inputvec)$, current input $\inputvec$ and reference input $\refinputvec$. We aim to show that, if CAFE returns scores $\scoresvec^\poss_1$ and $\scoresvec^\negs_1$ for $\model_1$, scores $\scoresvec^\poss_2$ and $\scoresvec^\negs_2$ for $\model_2$ and scores $\scoresvec^\poss_3$ and $\scoresvec^\negs_3$ for $\model_3$ (while considering the current input $\inputvec$ and the reference input $\refinputvec$), the following equality will hold: $\scoresvec^\poss_3 - \scoresvec^\negs_3 = a \cdot (\scoresvec^\poss_1 - \scoresvec^\negs_1) + b \cdot (\scoresvec^\poss_2 - \scoresvec^\negs_2)$. Using the definitions \ref{def:linear-layer-rule} and \ref{def:backward-attribution-scores}, we get:
		\begin{align*}
			&\scoresvec^\poss_3 - \scoresvec^\negs_3 \\
			&= \frac{\partial \scoresvec^\poss_3}{\partial \scoresvec^\poss_1} \odot \scoresvec^\poss_1 + \frac{\partial \scoresvec^\poss_3}{\partial \scoresvec^\negs_1} \odot \scoresvec^\negs_1 + \frac{\partial \scoresvec^\poss_3}{\partial \scoresvec^\poss_2} \odot \scoresvec^\poss_2 + \frac{\partial \scoresvec^\poss_3}{\partial \scoresvec^\negs_2} \odot \scoresvec^\negs_2 \\
			&\phantom{=)}- \frac{\partial \scoresvec^\negs_3}{\partial \scoresvec^\poss_1} \odot \scoresvec^\poss_1 - \frac{\partial \scoresvec^\negs_3}{\partial \scoresvec^\negs_1} \odot \scoresvec^\negs_1 - \frac{\partial \scoresvec^\negs_3}{\partial \scoresvec^\poss_2} \odot \scoresvec^\poss_2 - \frac{\partial \scoresvec^\negs_3}{\partial \scoresvec^\negs_2} \odot \scoresvec^\negs_2 \\
			&= \clip(a) \cdot \scoresvec^\poss_1 + \clip(-a) \cdot \scoresvec^\negs_1 + \clip(b) \cdot \scoresvec^\poss_2 + \clip(-b) \cdot \scoresvec^\negs_2 \\
			&\phantom{=(}-\clip(-a) \cdot \scoresvec^\poss_1 - \clip(a) \cdot \scoresvec^\negs_1 - \clip(-b) \cdot \scoresvec^\poss_2 - \clip(b) \cdot \scoresvec^\negs_2 \\
			&= (\clip(a) - \clip(-a)) \cdot (\scoresvec^\poss_1 - \scoresvec^\negs_1) \\
			&\phantom{=(}+ (\clip(b) - \clip(-b)) \cdot (\scoresvec^\poss_2 - \scoresvec^\negs_2) \\
			&= a \cdot (\scoresvec^\poss_1 - \scoresvec^\negs_1) + b \cdot (\scoresvec^\poss_2 - \scoresvec^\negs_2)
		\end{align*}
		as required. This concludes the proof. \aptLtoX[graphic=no,type=html]{&#x25A1;}{{$\square$}}
	\end{myproof}
	
	\subsection{Completeness}
	As our final property sourced from the literature, we consider completeness, which has been previously considered by several as a desirable property for feature attribution methods \cite{sundararajan-integrated-gradients, shrikumar-deeplift, ancona-towards-attribution}:
	
	\begin{property}[Completeness]
		A feature attribution method is complete iff, for any model $\model$, input $\inputvec$ and reference input $\refinputvec$, for which the attribution method computes scores $\mathbf{s}$:
		\begin{enumerate}
			\item[(i)] $\model(\inputvec) = \sum_{f} s_{\featidx}$, i.e. the sum of all returned feature attribution scores is equal to the model output, if the feature attribution method does not use a reference input.
			\item[(ii)] $\model(\inputvec) - \model(\refinputvec) = \sum_{f} s_{\featidx}$, i.e. the sum of all returned feature attribution scores is equal to the difference between the model outputs for the input $\inputvec$ and the reference input $\refinputvec$, if the feature attribution method uses a reference input.
		\end{enumerate}
	\end{property}
	
	Intuitively, the completeness property states that the attribution method should exactly account for all the changes to the model output as a result of the model input or the differences between the current input and the reference input. For CAFE, we wish to show the following proposition:
	
	\begin{proposition}
		\label{prop:cafe-completeness}
		CAFE satisfies the completeness property for any choice of the constant $\cancelconst{}$, provided that the negative attribution scores $\scoresvec^\negs$ and $\scorebias^\negs$ are considered to be negative when summing the scores and that $\model_{\reference}(\refinputvec)$ is considered as the baseline output.
	\end{proposition}
	
	As previously with linearity, our proposition needs to specify the handling of the two different kinds of scores returned by CAFE, as the original property was formulated for attribution methods only computing a single score. Similarly, since CAFE additionally captures the effects of the biases, we adjust the proposition to consider the reference output from a model without such biases $\model_{\reference}$. The proof of Proposition \ref{prop:cafe-completeness} is given in Proof \ref{proof:cafe-completeness}.
	
	\begin{myproof}
		\label{proof:cafe-completeness}
		We start by showing that completeness holds for the scores computed during the forward propagation, i.e.,  $\forall \currlayer. [\scoresvecn{\currlayer}{\poss} - \scoresvecn{\currlayer}{\negs} = \actvec{\currlayer} - \refactvec{\currlayer}]$, using structural induction on CAFE forward rules.
		
		For the \textbf{base case}, we aim to show that, for an arbitrary input $\inputvec$, reference input $\refinputvec$ and input layer $\layer{1}$, we have $\scoresvecn{1}{\poss} - \scoresvecn{1}{\negs} = \actvec{1} - \refactvec{1} = \inputvec - \refinputvec$. By Definition \ref{def:input-layer-rule}:
		\begin{align*}
			\scoresvecn{1}{\poss} - \scoresvecn{1}{\negs} = \clip(\vfoundation) - \clip(-\vfoundation) = \vfoundation = \inputvec - \refinputvec
		\end{align*}
		as required. This concludes the proof of the base case.
		
		For the \textbf{linear rule case}, we aim to show that, for an arbitrary linear layer $\layer{\currlayer + 1}$ where $\currlayer \geq 1$, we have $\scoresvecn{\currlayer + 1}{\poss} - \scoresvecn{\currlayer + 1}{\negs} = \actvec{\currlayer + 1} - \refactvec{\currlayer + 1}$, given the inductive hypothesis $\scoresvecn{\currlayer}{\poss} - \scoresvecn{\currlayer}{\negs} = \actvec{\currlayer} - \refactvec{\currlayer}$. Using Definition \ref{def:linear-layer-rule} and the inductive hypothesis, we get:
		{\allowdisplaybreaks
			\begin{align*}
				&\scoresvecn{\currlayer + 1}{\poss} - \scoresvecn{\currlayer + 1}{\negs} \\
				&= \scoresvecn{\currlayer}{\poss} \clip(\weightmat^{(\currlayer + 1)}) + \scoresvecn{\currlayer}{\negs} \clip(-\weightmat^{(\currlayer + 1)}) + \clip(\biasvec{\currlayer + 1}) \\
				&\phantom{=(}- \scoresvecn{\currlayer}{\poss} \clip(-\weightmat^{(\currlayer + 1)}) - \scoresvecn{\currlayer}{\negs} \clip(\weightmat^{(\currlayer + 1)}) - \clip(-\biasvec{\currlayer + 1}) \\
				&= (\clip(\weightmat^{(\currlayer + 1)}) - \clip(-\weightmat^{(\currlayer + 1)})) (\scoresvecn{\currlayer}{\poss} - \scoresvecn{\currlayer}{\negs}) + \clip(\biasvec{\currlayer + 1}) - \clip(-\biasvec{\currlayer + 1}) \\
				&= \weightmat^{(\currlayer + 1)} (\actvec{\currlayer} - \refactvec{\currlayer}) + \biasvec{\currlayer + 1} \\
				&= \weightmat^{(\currlayer + 1)} \actvec{\currlayer} + \biasvec{\currlayer + 1} - \weightmat^{(\currlayer + 1)} \refactvec{\currlayer} \\
				&= \actvec{\currlayer + 1} - \refactvec{\currlayer + 1}
			\end{align*}
		}
		as required. This concludes the proof of the linear rule case.
		
		Finally, we consider the \textbf{activation rule case}. We aim to show that, for an arbitrary activation layer $\layer{\currlayer + 1}$ where $\currlayer \geq 1$, we have $\scoresvecn{\currlayer + 1}{\poss} - \scoresvecn{\currlayer + 1}{\negs} = \actvec{\currlayer + 1} - \refactvec{\currlayer + 1}$, given the inductive hypothesis $\scoresvecn{\currlayer}{\poss} - \scoresvecn{\currlayer}{\negs} = \actvec{\currlayer} - \refactvec{\currlayer}$. In order to prove this statement, we will first show completeness for peak and linear score flows and later demonstrate that this implies completeness for the final activation layer scores. For the peak score flows in particular, we aim to show that $\vpeak{\currlayer + 1}{\poss}{\poss} + \vpeak{\currlayer + 1}{\negs}{\poss} - \vpeak{\currlayer + 1}{\poss}{\negs} - \vpeak{\currlayer + 1}{\negs}{\negs} = \actvec{\currlayer + 1} - \refactvec{\currlayer + 1}$. By applying Definition \ref{def:peak-effects-full} as well as the inductive hypothesis, we obtain:
		{\allowdisplaybreaks
			\begin{align*}
				&\vpeak{\currlayer + 1}{\poss}{\poss} + \vpeak{\currlayer + 1}{\negs}{\poss} - \vpeak{\currlayer + 1}{\poss}{\negs} - \vpeak{\currlayer + 1}{\negs}{\negs} \\
				&= \indvector{\currlayer}{\poss} \odot \clip(\simpleactfun(\oinputs{\currlayer + 1}{\combs}) - \simpleactfun(\refactvec{\currlayer})) +\max(\clip(\simpleactfun(\oinputs{\currlayer + 1}{\poss}) - \simpleactfun(\indvector{\currlayer}{\poss} \odot \oinputs{\currlayer + 1}{\combs} + \indvector{\currlayer}{\negs} \odot \refactvec{\currlayer})), \\
				&\phantom{=ssss}\clip(\simpleactfun(\indvector{\currlayer}{\poss} \odot \refactvec{\currlayer} + \indvector{\currlayer}{\negs} \odot \oinputs{\currlayer + 1}{\combs}) - \simpleactfun(\oinputs{\currlayer + 1}{\negs}))) \\
				&\phantom{=(}+ \indvector{\currlayer}{\negs} \odot \clip(\simpleactfun(\oinputs{\currlayer + 1}{\combs}) - \simpleactfun(\refactvec{\currlayer})) +\max(\clip(\simpleactfun(\simpleactfun(\indvector{\currlayer}{\poss} \odot \oinputs{\currlayer + 1}{\combs} + \indvector{\currlayer}{\negs} \odot \refactvec{\currlayer}) - \oinputs{\currlayer + 1}{\poss})), \\
				&\phantom{=ssss}\clip(\simpleactfun(\oinputs{\currlayer + 1}{\negs}) - \simpleactfun(\indvector{\currlayer}{\poss} \odot \refactvec{\currlayer} + \indvector{\currlayer}{\negs} \odot \oinputs{\currlayer + 1}{\combs}))) \\
				&\phantom{=(}- \indvector{\currlayer}{\poss} \odot \clip(\simpleactfun(\refactvec{\currlayer}) - \simpleactfun(\oinputs{\currlayer + 1}{\combs})) -\max(\clip(\simpleactfun(\simpleactfun(\indvector{\currlayer}{\poss} \odot \oinputs{\currlayer + 1}{\combs} + \indvector{\currlayer}{\negs} \odot \refactvec{\currlayer}) - \oinputs{\currlayer + 1}{\poss})), \\
				&\phantom{=ssss}\clip(\simpleactfun(\oinputs{\currlayer + 1}{\negs}) - \simpleactfun(\indvector{\currlayer}{\poss} \odot \refactvec{\currlayer} + \indvector{\currlayer}{\negs} \odot \oinputs{\currlayer + 1}{\combs}))) \displaybreak[0] \\
				&\phantom{=(}- \indvector{\currlayer}{\negs} \odot \clip(\simpleactfun(\refactvec{\currlayer}) - \simpleactfun(\oinputs{\currlayer + 1}{\combs})) -\max(\clip(\simpleactfun(\oinputs{\currlayer + 1}{\poss}) - \simpleactfun(\indvector{\currlayer}{\poss} \odot \oinputs{\currlayer + 1}{\combs} + \indvector{\currlayer}{\negs} \odot \refactvec{\currlayer})), \\
				&\phantom{=ssss}\clip(\simpleactfun(\indvector{\currlayer}{\poss} \odot \refactvec{\currlayer} + \indvector{\currlayer}{\negs} \odot \oinputs{\currlayer + 1}{\combs}) - \simpleactfun(\oinputs{\currlayer + 1}{\negs}))) \\
				&= \indvector{\currlayer}{\poss} \odot \clip(\simpleactfun(\oinputs{\currlayer + 1}{\combs}) - \simpleactfun(\refactvec{\currlayer})) + \indvector{\currlayer}{\negs} \odot \clip(\simpleactfun(\oinputs{\currlayer + 1}{\combs}) - \simpleactfun(\refactvec{\currlayer})) \\
				&\phantom{=(}- \indvector{\currlayer}{\poss} \odot \clip(\simpleactfun(\refactvec{\currlayer}) - \simpleactfun(\oinputs{\currlayer + 1}{\combs})) - \indvector{\currlayer}{\negs} \odot \clip(\simpleactfun(\refactvec{\currlayer}) - \simpleactfun(\oinputs{\currlayer + 1}{\combs})) \\
				&= \clip(\simpleactfun(\oinputs{\currlayer + 1}{\combs}) - \simpleactfun(\refactvec{\currlayer})) - \clip(\simpleactfun(\refactvec{\currlayer}) - \simpleactfun(\oinputs{\currlayer + 1}{\combs})) \\
				&= \simpleactfun(\oinputs{\currlayer + 1}{\combs}) - \simpleactfun(\refactvec{\currlayer}) \\
				&= \simpleactfun(\scoresvecn{\currlayer}{\combs} + \refactvec{\currlayer}) - \simpleactfun(\refactvec{\currlayer}) \\
				&= \simpleactfun(\actvec{\currlayer} - \refactvec{\currlayer} + \refactvec{\currlayer}) - \simpleactfun(\refactvec{\currlayer}) \\
				&= \simpleactfun(\actvec{\currlayer}) - \simpleactfun(\refactvec{\currlayer}) \\
				&= \actvec{\currlayer + 1} - \refactvec{\currlayer + 1}
			\end{align*}
		}
		as required.
		
		To show completeness for the linear score flows, we first prove a further lemma:
		\begin{align*}
			&\scoresvecn{\currlayer}{\poss} \odot \frac{\clip \left( \text{sgn}(\scoresvecn{\currlayer}{\combs}) \odot (\phi(\oinputs{\currlayer + 1}{\combs}) - \phi(\refactvec{\currlayer})) \right)}{|\scoresvecn{\currlayer}{\combs}| + \epsilon} + \scoresvecn{\currlayer}{\negs} \odot \frac{\clip \left( \text{sgn}(\scoresvecn{\currlayer}{\combs}) \odot (\phi(\refactvec{\currlayer}) - \phi(\oinputs{\currlayer + 1}{\combs})) \right)}{|\scoresvecn{\currlayer}{\combs}| + \epsilon} \\
			&- \scoresvecn{\currlayer}{\poss} \odot \frac{\clip \left( \text{sgn}(\scoresvecn{\currlayer}{\combs}) \odot (\phi(\refactvec{\currlayer}) - \phi(\oinputs{\currlayer + 1}{\combs})) \right)}{|\scoresvecn{\currlayer}{\combs}| + \epsilon} - \scoresvecn{\currlayer}{\negs} \odot \frac{\clip \left( \text{sgn}(\scoresvecn{\currlayer}{\combs}) \odot (\phi(\oinputs{\currlayer + 1}{\combs}) - \phi(\refactvec{\currlayer})) \right)}{|\scoresvecn{\currlayer}{\combs}| + \epsilon} \\
			&= \actvec{\currlayer + 1} - \refactvec{\currlayer + 1}
		\end{align*}
		
		Using the inductive hypothesis, we have:
		{\allowdisplaybreaks
			\begin{align*}
				&\scoresvecn{\currlayer}{\poss} \odot \frac{\clip \left( \text{sgn}(\scoresvecn{\currlayer}{\combs}) \odot (\phi(\oinputs{\currlayer + 1}{\combs}) - \phi(\refactvec{\currlayer})) \right)}{|\scoresvecn{\currlayer}{\combs}| + \epsilon} + \scoresvecn{\currlayer}{\negs} \odot \frac{\clip \left( \text{sgn}(\scoresvecn{\currlayer}{\combs}) \odot (\phi(\refactvec{\currlayer})  - \phi(\oinputs{\currlayer + 1}{\combs})) \right)}{|\scoresvecn{\currlayer}{\combs}| + \epsilon} \\
				&- \scoresvecn{\currlayer}{\poss} \odot \frac{\clip \left( \text{sgn}(\scoresvecn{\currlayer}{\combs}) \odot (\phi(\refactvec{\currlayer}) - \phi(\oinputs{\currlayer + 1}{\combs})) \right)}{|\scoresvecn{\currlayer}{\combs}| + \epsilon} - \scoresvecn{\currlayer}{\negs} \odot \frac{\clip \left( \text{sgn}(\scoresvecn{\currlayer}{\combs}) \odot (\phi(\oinputs{\currlayer + 1}{\combs}) - \phi(\refactvec{\currlayer})) \right)}{|\scoresvecn{\currlayer}{\combs}| + \epsilon} \\
				&= \frac{(\scoresvecn{\currlayer}{\poss} - \scoresvecn{\currlayer}{\negs}) \odot \left( \clip \left( \text{sgn}(\scoresvecn{\currlayer}{\combs}) \odot (\phi(\oinputs{\currlayer + 1}{\combs}) - \phi(\refactvec{\currlayer})) \right) - \clip \left( \text{sgn}(\scoresvecn{\currlayer}{\combs}) \odot (\phi(\refactvec{\currlayer}) - \phi(\oinputs{\currlayer + 1}{\combs})) \right) \right)}{|\scoresvecn{\currlayer}{\combs}| + \epsilon} \\
				&= \frac{\scoresvecn{\currlayer}{\combs}}{|\scoresvecn{\currlayer}{\combs}| + \epsilon} \odot \text{sgn}(\scoresvecn{\currlayer}{\combs}) \odot (\phi(\oinputs{\currlayer + 1}{\combs}) - \phi(\refactvec{\currlayer})) \\
				&= \text{sgn}(\scoresvecn{\currlayer}{\combs}) \odot \text{sgn}(\scoresvecn{\currlayer}{\combs}) \odot (\phi(\oinputs{\currlayer + 1}{\combs}) - \phi(\refactvec{\currlayer})) \\
				&= \phi(\oinputs{\currlayer + 1}{\combs}) - \phi(\refactvec{\currlayer}) \\
				&= ... \\
				&= \actvec{\currlayer + 1} - \refactvec{\currlayer + 1}
			\end{align*}
		}
		as required. Note that we assumed that $\epsilon$ is negligible. We also omitted several of the proof steps, as they would be identical to the steps in the proof for peak attribution scores. Given the completeness result for peak attribution scores and the lemma shown immediately above, we can now show that linear score flows themselves also satisfy completeness. We take an arbitrary index $\idxtwo$ and consider the following exhaustive cases:
		\begin{enumerate}
			\item[(i)] $\text{sgn}(\scoresvecnelem{\currlayer}{\combs}{\idxtwo}) \cdot (\phi(\oinputselem{\currlayer + 1}{\combs}{\idxtwo}) - \phi(\refactelem{\currlayer}{\idxtwo})) > 0$. Then necessarily $\indvectorelem{\currlayer}{\negs}{\idxtwo} \clip(\simpleactfun(\oinputselem{\currlayer + 1}{\combs}{\idxtwo}) - \simpleactfun(\refactelem{\currlayer}{\idxtwo})) = \indvectorelem{\currlayer}{\poss}{\idxtwo} \clip(\simpleactfun(\refactelem{\currlayer}{\idxtwo}) - \simpleactfun(\oinputselem{\currlayer + 1}{\combs}{\idxtwo})) = 0$ and $\peak{\currlayer + 1}{\negs}{\poss}{\idxtwo} = \peak{\currlayer + 1}{\poss}{\negs}{\idxtwo}$. Thus, by peak score flows completeness, $\peak{\currlayer + 1}{\poss}{\poss}{\idxtwo} - \peak{\currlayer + 1}{\negs}{\negs}{\idxtwo} = \actvalelem{\currlayer + 1}{\idxtwo} - \refactelem{\currlayer + 1}{\idxtwo}$. Since $\text{sgn}(\scoresvecnelem{\currlayer}{\combs}{\idxtwo}) \cdot (\phi(\refactelem{\currlayer}{\idxtwo}) - \phi(\oinputselem{\currlayer + 1}{\combs}{\idxtwo})) = 0$, we also get that $\linear{\currlayer + 1}{\negs}{\poss}{\idxtwo} = \linear{\currlayer + 1}{\poss}{\negs}{\idxtwo} = 0$. By the lemma, it must also necessarily hold that $\scoresvecnelem{\currlayer}{\poss}{\idxtwo} \frac{\clip \left( \text{sgn}(\scoresvecnelem{\currlayer}{\combs}{\idxtwo}) \cdot (\phi(\oinputselem{\currlayer + 1}{\combs}{\idxtwo}) - \phi(\refactelem{\currlayer}{\idxtwo})) \right)}{|\scoresvecnelem{\currlayer}{\combs}{\idxtwo}| + \epsilon} - \scoresvecnelem{\currlayer}{\negs}{\idxtwo} \frac{\clip \left( \text{sgn}(\scoresvecnelem{\currlayer}{\combs}{\idxtwo}) \cdot (\phi(\oinputselem{\currlayer + 1}{\combs}{\idxtwo}) - \phi(\refactelem{\currlayer}{\idxtwo})) \right)}{|\scoresvecnelem{\currlayer}{\combs}{\idxtwo}| + \epsilon} = \actvalelem{\currlayer + 1}{\idxtwo} - \refactelem{\currlayer + 1}{\idxtwo}$. Then, since it must either be the case that $\peak{\currlayer + 1}{\poss}{\poss}{\idxtwo} \geq \scoresvecnelem{\currlayer}{\poss}{\idxtwo} \frac{\clip \left( \text{sgn}(\scoresvecnelem{\currlayer}{\combs}{\idxtwo}) \cdot (\phi(\oinputselem{\currlayer + 1}{\combs}{\idxtwo}) - \phi(\refactelem{\currlayer}{\idxtwo})) \right)}{|\scoresvecnelem{\currlayer}{\combs}{\idxtwo}| + \epsilon} \land \peak{\currlayer + 1}{\negs}{\negs}{\idxtwo} \geq \scoresvecnelem{\currlayer}{\negs}{\idxtwo} \frac{\clip \left( \text{sgn}(\scoresvecnelem{\currlayer}{\combs}{\idxtwo}) \cdot (\phi(\oinputselem{\currlayer + 1}{\combs}{\idxtwo}) - \phi(\refactelem{\currlayer}{\idxtwo})) \right)}{|\scoresvecnelem{\currlayer}{\combs}{\idxtwo}| + \epsilon}$ or that $\peak{\currlayer + 1}{\poss}{\poss}{\idxtwo} < \scoresvecnelem{\currlayer}{\poss}{\idxtwo} \frac{\clip \left( \text{sgn}(\scoresvecnelem{\currlayer}{\combs}{\idxtwo}) \cdot (\phi(\oinputselem{\currlayer + 1}{\combs}{\idxtwo}) - \phi(\refactelem{\currlayer}{\idxtwo})) \right)}{|\scoresvecnelem{\currlayer}{\combs}{\idxtwo}| + \epsilon} \land \peak{\currlayer + 1}{\negs}{\negs}{\idxtwo} < \scoresvecnelem{\currlayer}{\negs}{\idxtwo} \frac{\clip \left( \text{sgn}(\scoresvecnelem{\currlayer}{\combs}{\idxtwo}) \cdot (\phi(\oinputselem{\currlayer + 1}{\combs}{\idxtwo}) - \phi(\refactelem{\currlayer}{\idxtwo})) \right)}{|\scoresvecnelem{\currlayer}{\combs}{\idxtwo}| + \epsilon}$, we get that $\linear{\currlayer + 1}{\poss}{\poss}{\idxtwo} - \linear{\currlayer + 1}{\negs}{\negs}{\idxtwo} = \actvalelem{\currlayer + 1}{\idxtwo} - \refactelem{\currlayer + 1}{\idxtwo}$, and thus completeness holds in this case.
			\item[(ii)] $\text{sgn}(\scoresvecnelem{\currlayer}{\combs}{\idxtwo}) \cdot (\phi(\refactelem{\currlayer}{\idxtwo}) - \phi(\oinputselem{\currlayer + 1}{\combs}{\idxtwo})) > 0$. Then necessarily $\indvectorelem{\currlayer}{\poss}{\idxtwo} \clip(\simpleactfun(\refactelem{\currlayer}{\idxtwo}) - \simpleactfun(\oinputselem{\currlayer + 1}{\combs}{\idxtwo})) = \indvectorelem{\currlayer}{\negs}{\idxtwo} \clip(\simpleactfun(\oinputselem{\currlayer + 1}{\combs}{\idxtwo}) - \simpleactfun(\refactelem{\currlayer}{\idxtwo})) = 0$ and $\peak{\currlayer + 1}{\poss}{\poss}{\idxtwo} = \peak{\currlayer + 1}{\negs}{\negs}{\idxtwo}$. Thus, by peak score flows completeness, $\peak{\currlayer + 1}{\negs}{\poss}{\idxtwo} - \peak{\currlayer + 1}{\poss}{\negs}{\idxtwo} = \actvalelem{\currlayer + 1}{\idxtwo} - \refactelem{\currlayer + 1}{\idxtwo}$. Since $\text{sgn}(\scoresvecnelem{\currlayer}{\combs}{\idxtwo}) \cdot (\phi(\oinputselem{\currlayer + 1}{\combs}{\idxtwo}) - \phi(\refactelem{\currlayer}{\idxtwo})) = 0$, we also get that $\linear{\currlayer + 1}{\poss}{\poss}{\idxtwo} = \linear{\currlayer + 1}{\negs}{\negs}{\idxtwo} = 0$. By the lemma, it must also necessarily hold that
			$\scoresvecnelem{\currlayer}{\negs}{\idxtwo} \frac{\clip \left( \text{sgn}(\scoresvecnelem{\currlayer}{\combs}{\idxtwo}) \cdot (\phi(\refactelem{\currlayer}{\idxtwo})  - \phi(\oinputselem{\currlayer + 1}{\combs}{\idxtwo})) \right)}{|\scoresvecnelem{\currlayer}{\combs}{\idxtwo}| + \epsilon} - \scoresvecnelem{\currlayer}{\poss}{\idxtwo} \frac{\clip \left( \text{sgn}(\scoresvecnelem{\currlayer}{\combs}{\idxtwo}) \cdot (\phi(\refactelem{\currlayer}{\idxtwo}) - \phi(\oinputselem{\currlayer + 1}{\combs}{\idxtwo})) \right)}{|\scoresvecnelem{\currlayer}{\combs}{\idxtwo}| + \epsilon} = \actvalelem{\currlayer + 1}{\idxtwo} - \refactelem{\currlayer + 1}{\idxtwo}$. Then, since it must either be the case that $\peak{\currlayer + 1}{\negs}{\poss}{\idxtwo} \geq \scoresvecnelem{\currlayer}{\negs}{\idxtwo} \frac{\clip \left( \text{sgn}(\scoresvecnelem{\currlayer}{\combs}{\idxtwo}) \cdot (\phi(\refactelem{\currlayer}{\idxtwo})  - \phi(\oinputselem{\currlayer + 1}{\combs}{\idxtwo})) \right)}{|\scoresvecnelem{\currlayer}{\combs}{\idxtwo}| + \epsilon} \land \peak{\currlayer + 1}{\poss}{\negs}{\idxtwo} \geq \scoresvecnelem{\currlayer}{\poss}{\idxtwo} \frac{\clip \left( \text{sgn}(\scoresvecnelem{\currlayer}{\combs}{\idxtwo}) \cdot (\phi(\refactelem{\currlayer}{\idxtwo}) - \phi(\oinputselem{\currlayer + 1}{\combs}{\idxtwo})) \right)}{|\scoresvecnelem{\currlayer}{\combs}{\idxtwo}| + \epsilon}$ or that $\peak{\currlayer + 1}{\negs}{\poss}{\idxtwo} < \scoresvecnelem{\currlayer}{\negs}{\idxtwo} \frac{\clip \left( \text{sgn}(\scoresvecnelem{\currlayer}{\combs}{\idxtwo}) \cdot (\phi(\refactelem{\currlayer}{\idxtwo})  - \phi(\oinputselem{\currlayer + 1}{\combs}{\idxtwo})) \right)}{|\scoresvecnelem{\currlayer}{\combs}{\idxtwo}| + \epsilon} \land \peak{\currlayer + 1}{\poss}{\negs}{\idxtwo} < \scoresvecnelem{\currlayer}{\poss}{\idxtwo} \frac{\clip \left( \text{sgn}(\scoresvecnelem{\currlayer}{\combs}{\idxtwo}) \cdot (\phi(\refactelem{\currlayer}{\idxtwo}) - \phi(\oinputselem{\currlayer + 1}{\combs}{\idxtwo})) \right)}{|\scoresvecnelem{\currlayer}{\combs}{\idxtwo}| + \epsilon}$, we get that $\linear{\currlayer + 1}{\negs}{\poss}{\idxtwo} - \linear{\currlayer + 1}{\poss}{\negs}{\idxtwo} = \actvalelem{\currlayer + 1}{\idxtwo} - \refactelem{\currlayer + 1}{\idxtwo}$, and thus completeness holds in this case.
			\item[(iii)] $\text{sgn}(\scoresvecnelem{\currlayer}{\combs}{\idxtwo}) \cdot (\phi(\oinputselem{\currlayer + 1}{\combs}{\idxtwo}) - \phi(\refactelem{\currlayer}{\idxtwo})) = \text{sgn}(\scoresvecnelem{\currlayer}{\combs}{\idxtwo}) \cdot (\phi(\refactelem{\currlayer}{\idxtwo}) - \phi(\oinputselem{\currlayer + 1}{\combs}{\idxtwo})) = 0$. Then, it is necessarily the case that $\linear{\currlayer + 1}{\poss}{\poss}{\idxtwo} = \linear{\currlayer + 1}{\negs}{\poss}{\idxtwo} = \linear{\currlayer + 1}{\poss}{\negs}{\idxtwo} = \linear{\currlayer + 1}{\negs}{\negs}{\idxtwo} = 0$. Since necessarily $\actvalelem{\currlayer + 1}{\idxtwo} - \refactelem{\currlayer + 1}{\idxtwo}$, completeness is also satisfied in this case.
		\end{enumerate}
		
		By showing the above for an arbitrary index $\idxtwo$, we have also shown that $\vlinear{\currlayer + 1}{\poss}{\poss} + \vlinear{\currlayer + 1}{\negs}{\poss} - \vlinear{\currlayer + 1}{\poss}{\negs} - \vlinear{\currlayer + 1}{\negs}{\negs} = \actvec{\currlayer + 1} - \refactvec{\currlayer + 1}$, as desired. Now that we have proven completeness for both peak attribution flows and clipped linear attribution flows, it remains to show that completeness also holds for the final attribution scores. Using definitions \ref{def:activation-multipliers} and \ref{def:activation-scores}, we get:
		
		{\allowdisplaybreaks
			\begin{align*}
				&\scoresvecn{\currlayer + 1}{\poss} - \scoresvecn{\currlayer}{\negs} \\
				&= \vmultiplier{\currlayer + 1}{\poss}{\poss} \odot \scoresvecn{\currlayer}{\poss} + \vmultiplier{\currlayer + 1}{\negs}{\poss} \odot \scoresvecn{\currlayer}{\negs} \\
				&\phantom{=(}-\vmultiplier{\currlayer + 1}{\poss}{\negs} \odot \scoresvecn{\currlayer}{\poss} - \vmultiplier{\currlayer + 1}{\negs}{\negs} \odot \scoresvecn{\currlayer}{\negs} \\
				&= \Bigg( \frac{(1 - \cancelconst{\currlayer + 1}) \vlinear{\currlayer + 1}{\poss}{\poss} + \cancelconst{\currlayer + 1} \vpeak{\currlayer + 1}{\poss}{\poss}}{\scoresvecn{\currlayer}{\poss} + \epsilon} \\
				&\phantom{\Bigg(=}- \frac{(1 - \cancelconst{\currlayer + 1}) \vlinear{\currlayer + 1}{\poss}{\negs} + \cancelconst{\currlayer + 1} \vpeak{\currlayer + 1}{\poss}{\negs}}{\scoresvecn{\currlayer}{\poss} + \epsilon} \Bigg) \odot \scoresvecn{\currlayer}{\poss} \\
				&\phantom{=(}+\Bigg( \frac{(1 - \cancelconst{\currlayer + 1}) \vlinear{\currlayer + 1}{\negs}{\poss} + \cancelconst{\currlayer + 1} \vpeak{\currlayer + 1}{\negs}{\poss}}{\scoresvecn{\currlayer}{\negs} + \epsilon} \\
				&\phantom{=(\Bigg(=}-\frac{(1 - \cancelconst{\currlayer + 1}) \vlinear{\currlayer + 1}{\negs}{\negs} + \cancelconst{\currlayer + 1} \vpeak{\currlayer + 1}{\negs}{\negs}}{\scoresvecn{\currlayer}{\negs} + \epsilon} \Bigg) \odot \scoresvecn{\currlayer}{\negs} \\
				&= (1 - \cancelconst{\currlayer + 1}) \cdot (\vlinear{\currlayer + 1}{\poss}{\poss} + \vlinear{\currlayer + 1}{\negs}{\poss} - \vlinear{\currlayer + 1}{\poss}{\negs} - \vlinear{\currlayer + 1}{\negs}{\negs}) \\
				&\phantom{=(}+ \cancelconst{\currlayer + 1} \cdot (\vpeak{\currlayer + 1}{\poss}{\poss} + \vpeak{\currlayer + 1}{\negs}{\poss} - \vpeak{\currlayer + 1}{\poss}{\negs} - \vpeak{\currlayer + 1}{\negs}{\negs}) \\
				&= (1 - \cancelconst{\currlayer + 1}) (\actvec{\currlayer + 1} - \refactvec{\currlayer + 1}) + \cancelconst{\currlayer + 1} (\actvec{\currlayer + 1} - \refactvec{\currlayer + 1}) \\
				&= \actvec{\currlayer + 1} - \refactvec{\currlayer + 1}
			\end{align*}
		}
		which concludes the proof for the activation layer rule. Note that completeness holds even if $\scoresvecn{\currlayer}{\poss} = 0$ or $\scoresvecn{\currlayer}{\negs} = 0$, as the corresponding attribution flows will necessarily be zero as well in such a case, ensuring that no attribution is ``lost". We have again employed the assumption that $\epsilon$ is negligible.
		
		Now that we have shown that $\forall \currlayer. [\scoresvecn{\currlayer}{\poss} - \scoresvecn{\currlayer}{\negs} = \actvec{\currlayer} - \refactvec{\currlayer}]$, it is clear that for any selected target neuron index $\targetneuron$, $\scoresvecnelem{\nlayers}{\poss}{\targetneuron} - \scoresvecnelem{\nlayers}{\negs}{\targetneuron} = \actvalelem{\nlayers}{\targetneuron} - \refactelem{\nlayers}{\targetneuron}$. Since the score multipliers are treated as constants during differentiation, for the purposes of the backward pass, the scores $\scoresvecnelem{\nlayers}{\poss}{\targetneuron}$ and $\scoresvecnelem{\nlayers}{\negs}{\targetneuron}$ are a linear combination of the individual elements of $\scoresvec^\poss$, $\scoresvec^\negs$ as well as $\scorebias^\poss$ and $\scorebias^\negs$. Thus, by Definition \ref{def:backward-attribution-scores}, we have:
		\begin{align*}
			&\sum^{\ldim{0}}_\idxone \scoresvecelem{\poss}{\idxone} + \scorebias^\poss - \sum^{\ldim{0}}_\idxone \scoresvecelem{\negs}{\idxone} - \scorebias^\negs \\
			&= \sum^{\ldim{0}}_\idxone \left[ \frac{\partial \scoresvecnelem{\nlayers}{\poss}{\targetneuron}}{\partial \vfoundation} \odot \vfoundation \right]_i + \sum^{\nlayers}_{\currlayer} \sum^{\ldim{\currlayer}}_{\idxone} \left[ \frac{\partial \scoresvecnelem{\nlayers}{\poss}{\targetneuron}}{\partial \biasvec{\currlayer}} \odot \biasvec{\currlayer} \right]_i - \sum^{\ldim{0}}_\idxone \left[ \frac{\partial \scoresvecnelem{\nlayers}{\negs}{\targetneuron}}{\partial \vfoundation} \odot \vfoundation \right]_i - \sum^{\nlayers}_{n} \sum^{\ldim{\currlayer}}_{\idxone} \left[ \frac{\partial \scoresvecnelem{\nlayers}{\negs}{\targetneuron}}{\partial \biasvec{\currlayer}} \odot \biasvec{\currlayer} \right]_i \\
			&= \scoresvecnelem{\nlayers}{\poss}{\targetneuron} - \scoresvecnelem{\nlayers}{\negs}{\targetneuron} \\
			&= \actvalelem{\nlayers}{\targetneuron} - \refactelem{\nlayers}{\targetneuron}
		\end{align*}
		which completes the proof. \aptLtoX[graphic=no,type=html]{&#x25A1;}{{$\square$}}
	\end{myproof}
	
	\subsection{Local Conflict Awareness}
	The proof of Theorem \ref{theo:cafe-conflict-awareness} is given in Proof \ref{proof:cafe-conflict-awareness}.
	
	\begin{myproof}[Theorem \ref{theo:cafe-conflict-awareness}]
		\label{proof:cafe-conflict-awareness}
		We aim to show that CAFE with $\cancelconst{} = 1$ computed with respect to a reference input of zero does not underreport any minimal local conflicts for ReLU neurons. Consider an arbitrary ReLU neuron with $\actelem{\currlayer+1}{\idxthree} = \relu(\actelem{\currlayer}{\idxthree}) = \relu(\sum^{\ldim{\currlayer - 1}}_\idxone \weightelem{\currlayer}{\idxone}{\idxthree} \actelem{\currlayer - 1}{\idxone})$, a zero reference activation vector $\refactvec{\currlayer - 1}$, a feature set $C'$ and a set $S^C$ containing all sets $C$ such that $C$ counteracts $C'$ in a minimal conflict with respect to $\actelem{\currlayer + 1}{\idxthree}$ and $\refactvec{\currlayer - 1}$. By Propositions \ref{prop:conflict-relu-positive} and \ref{prop:conflict-relu-negative}, we know that $C' = \{\idxtwo\}$ for some $\idxtwo$ (unless $S^C$ is empty, which would make the statement vacuously true).
		
		Before we start proving the main statement, we will first show the following lemma:
		\begin{equation*}
			\argmax_{C \in S^C} \cmag(\model[\currlayer:\currlayer + 1]_{\idxthree}, C, C', \actelem{\currlayer - 1}{\idxthree}, \refactvec{\currlayer - 1}) \leq \left|\weightelem{\currlayer}{\idxtwo}{\idxthree} (\actelem{\currlayer - 1}{\idxtwo} - \refactelem{\currlayer - 1}{\idxtwo})\right|
		\end{equation*}
		
		Take an arbitrary feature set $C$ from $S^C$. Then,
		{\allowdisplaybreaks
			\begin{align*}
				&\countermag(\model[\currlayer:\currlayer + 1]_{\idxthree}, C, C', \actelem{\currlayer - 1}{\idxthree}, \refactvec{\currlayer - 1}) \\
				&= \min\left(
				\left|\layer{\currlayer:\currlayer + 1}\left(\actvec{\currlayer - 1}\right)_\idxthree - \layer{\currlayer:\currlayer + 1}\left(\actvec{\currlayer - 1} \big| \remove(C)\right)_\idxthree\right|, 
				\left|\layer{\currlayer:\currlayer + 1}\left(\actvec{\currlayer - 1} \big| \remove(C)\right)_\idxthree - \layer{\currlayer:\currlayer + 1}\left(\actvec{\currlayer - 1} \big| \remove(C \cup C')\right)_\idxthree\right|
				\right) \\
				&\leq \left|\layer{\currlayer:\currlayer + 1}\left(\actvec{\currlayer - 1} \big| \remove(C)\right)_\idxthree - \layer{\currlayer:\currlayer + 1}\left(\actvec{\currlayer - 1} \big| \remove(C \cup C')\right)_\idxthree\right| \\
				&= \left|\layer{\currlayer:\currlayer + 1}\left(\actvec{\currlayer - 1} \big| \left[\actelem{\currlayer - 1}{\idxone} := \refactelem{\currlayer}{\idxone}\right]_{\idxone \in C} \right)_\idxthree - \layer{\currlayer:\currlayer + 1}\left(\actvec{\currlayer - 1} \big| \left[\actelem{\currlayer - 1}{\idxone} := \refactelem{\currlayer}{\idxone}\right]_{\idxone \in C \cup C'}\right)_\idxthree\right| \\
				&= \Bigg| \relu\left(\actelem{n}{\idxthree} - \sum^{\ldim{\currlayer - 1}}_{\idxone \in C} \weightelem{\currlayer}{\idxone}{\idxthree} (\actelem{\currlayer - 1}{\idxone} - \refactelem{\currlayer - 1}{\idxone})\right) \\
				&\phantom{=(}- \relu\left(\actelem{n}{\idxthree} - \sum^{\ldim{\currlayer - 1}}_{\idxone \in C} \weightelem{\currlayer}{\idxone}{\idxthree} (\actelem{\currlayer - 1}{\idxone} - \refactelem{\currlayer - 1}{\idxone}) - \weightelem{\currlayer}{\idxtwo}{\idxthree} (\actelem{\currlayer - 1}{\idxtwo} - \refactelem{\currlayer - 1}{\idxtwo})\right) \Bigg| \\
				&\leq \left|\weightelem{\currlayer}{\idxtwo}{\idxthree} (\actelem{\currlayer - 1}{\idxtwo} - \refactelem{\currlayer - 1}{\idxtwo})\right|
			\end{align*}
		}
		
		Similarly,
		{\allowdisplaybreaks
			\begin{align*}
				&\countermag(\model[\currlayer:\currlayer + 1]_{\idxthree}, C', C, \actelem{\currlayer - 1}{\idxthree}, \refactvec{\currlayer - 1}) \\
				&\leq \min\left(
				\left|\layer{\currlayer:\currlayer + 1}\left(\actvec{\currlayer - 1}\right)_\idxthree - \layer{\currlayer:\currlayer + 1}\left(\actvec{\currlayer - 1} \big| \remove(C')\right)_\idxthree\right|, 
				\left|\layer{\currlayer:\currlayer + 1}\left(\actvec{\currlayer - 1} \big| \remove(C')\right)_\idxthree - \layer{\currlayer:\currlayer + 1}\left(\actvec{\currlayer - 1} \big| \remove(C' \cup C)\right)_\idxthree\right|
				\right) \\
				&\leq \left|\layer{\currlayer:\currlayer + 1}\left(\actvec{\currlayer - 1}\right)_\idxthree - \layer{\currlayer:\currlayer + 1}\left(\actvec{\currlayer - 1} \big| \remove(C')\right)_\idxthree\right| \\
				&= \left|\layer{\currlayer:\currlayer + 1}\left(\actvec{\currlayer - 1}\right)_\idxthree - \layer{\currlayer:\currlayer + 1}\left(\actvec{\currlayer - 1} \big| \actelem{\currlayer - 1}{\idxtwo} := \refactelem{\currlayer}{\idxtwo}\right)_\idxthree\right| \\
				&= \Bigg| \relu\left(\actelem{n}{\idxthree}\right) - \relu\left(\actelem{n}{\idxthree} - \weightelem{\currlayer}{\idxtwo}{\idxthree} (\actelem{\currlayer - 1}{\idxtwo} - \refactelem{\currlayer - 1}{\idxtwo})\right) \Bigg| \\
				&\leq \left|\weightelem{\currlayer}{\idxtwo}{\idxthree} (\actelem{\currlayer - 1}{\idxtwo} - \refactelem{\currlayer - 1}{\idxtwo})\right|
		\end{align*}}
		
		From the above results, it immediately follows that:
		\begin{align*}
			&\cmag(\model[\currlayer:m]_\idxthree, C, C', \actvec{\currlayer - 1}, \refactvec{\currlayer - 1}) \\
			&= \max\left(\countermag(\model[\currlayer:m]_\idxthree, C, C', \actvec{\currlayer - 1}, \refactvec{\currlayer - 1}), \countermag(\model[\currlayer:m]_\idxthree, C', C, \actvec{\currlayer - 1}, \refactvec{\currlayer - 1})\right) \\
			&\leq \left|\weightelem{\currlayer}{\idxtwo}{\idxthree} (\actelem{\currlayer - 1}{\idxtwo} - \refactelem{\currlayer - 1}{\idxtwo})\right|
		\end{align*}
		which is sufficient for proving the lemma.
		
		Next, we aim to show that:
		\begin{align*}
			&\left|\scoresvecelem{\combs}{\idxtwo}\right|
			= \left|\scoresvecelem{\poss}{\idxtwo} - \scoresvecelem{\negs}{\idxtwo} \right|
			= \left| \frac{\partial (\scoresvecnelem{\currlayer + 1}{\poss}{\idxthree} - \scoresvecnelem{\currlayer + 1}{\negs}{\idxthree})}{{\partial(\actelem{\currlayer - 1}{\idxtwo} - \refactelem{\currlayer - 1}{\idxtwo})}} (\actelem{\currlayer - 1}{\idxtwo} - \refactelem{\currlayer - 1}{\idxtwo}) \right|
			\geq \left|\weightelem{\currlayer}{\idxtwo}{\idxthree} (\actelem{\currlayer - 1}{\idxtwo} - \refactelem{\currlayer - 1}{\idxtwo})\right|
		\end{align*}
		
		Therefore, it is sufficient to show:
		\begin{align*}
			\left| \frac{\partial (\scoresvecnelem{\currlayer + 1}{\poss}{\idxthree} - \scoresvecnelem{\currlayer + 1}{\negs}{\idxthree})}{{\partial(\actelem{\currlayer - 1}{\idxtwo} - \refactelem{\currlayer - 1}{\idxtwo})}} \right|
			\geq \left|\weightelem{\currlayer}{\idxtwo}{\idxthree} \right|
		\end{align*}
		
		As a first step, we determine the possible values of $\frac{\partial \scoresvecnelem{\currlayer}{\poss}{\idxthree}}{\partial(\actelem{\currlayer - 1}{\idxtwo} - \refactelem{\currlayer - 1}{\idxtwo})}$ and $\frac{\partial \scoresvecnelem{\currlayer}{\negs}{\idxthree}}{\partial(\actelem{\currlayer - 1}{\idxtwo} - \refactelem{\currlayer - 1}{\idxtwo})}$. We consider four possible cases:
		
		{\allowdisplaybreaks
			\begin{enumerate}
				\item[(i)] $\weightelem{\currlayer}{\idxtwo}{\idxthree} \geq 0$ and $\actelem{\currlayer - 1}{\idxtwo} - \refactelem{\currlayer - 1}{\idxtwo} \geq 0$. Then, by Definitions \ref{def:input-layer-rule} and \ref{def:linear-layer-rule}:
				\begin{align*}
					\frac{\partial \scoresvecnelem{\currlayer}{\poss}{\idxthree}}{\partial (\actelem{\currlayer - 1}{\idxtwo} - \refactelem{\currlayer - 1}{\idxtwo})} &= \weightelem{\currlayer}{\idxtwo}{\idxthree} \frac{\partial \scoresvecnelem{\currlayer - 1}{\poss}{\idxthree}}{\partial (\actelem{\currlayer - 1}{\idxtwo} - \refactelem{\currlayer - 1}{\idxtwo})} = \weightelem{\currlayer}{\idxtwo}{\idxthree} \qquad
					\frac{\partial \scoresvecnelem{\currlayer}{\negs}{\idxthree}}{\partial (\actelem{\currlayer - 1}{\idxtwo} - \refactelem{\currlayer - 1}{\idxtwo})} &= \weightelem{\currlayer}{\idxtwo}{\idxthree} \frac{\partial \scoresvecnelem{\currlayer - 1}{\negs}{\idxthree}}{\partial (\actelem{\currlayer - 1}{\idxtwo} - \refactelem{\currlayer - 1}{\idxtwo})} = 0
				\end{align*}
				\item[(ii)] $\weightelem{\currlayer}{\idxtwo}{\idxthree} < 0$ and $\actelem{\currlayer - 1}{\idxtwo} - \refactelem{\currlayer - 1}{\idxtwo} \geq 0$. Then, by Definitions \ref{def:input-layer-rule} and \ref{def:linear-layer-rule}:
				\begin{align*}
					\frac{\partial \scoresvecnelem{\currlayer}{\poss}{\idxthree}}{\partial (\actelem{\currlayer - 1}{\idxtwo} - \refactelem{\currlayer - 1}{\idxtwo})} &= \weightelem{\currlayer}{\idxtwo}{\idxthree} \frac{\partial \scoresvecnelem{\currlayer - 1}{\negs}{\idxthree}}{\partial (\actelem{\currlayer - 1}{\idxtwo} - \refactelem{\currlayer - 1}{\idxtwo})} = 0 \qquad
					\frac{\partial \scoresvecnelem{\currlayer}{\negs}{\idxthree}}{\partial (\actelem{\currlayer - 1}{\idxtwo} - \refactelem{\currlayer - 1}{\idxtwo})} &= \weightelem{\currlayer}{\idxtwo}{\idxthree} \frac{\partial \scoresvecnelem{\currlayer - 1}{\poss}{\idxthree}}{\partial (\actelem{\currlayer - 1}{\idxtwo} - \refactelem{\currlayer - 1}{\idxtwo})} = \weightelem{\currlayer}{\idxtwo}{\idxthree}
				\end{align*}
				\item[(iii)] $\weightelem{\currlayer}{\idxtwo}{\idxthree} \geq 0$ and $\actelem{\currlayer - 1}{\idxtwo} - \refactelem{\currlayer - 1}{\idxtwo} < 0$. Then, by Definitions \ref{def:input-layer-rule} and \ref{def:linear-layer-rule}:
				\begin{align*}
					\frac{\partial \scoresvecnelem{\currlayer}{\poss}{\idxthree}}{\partial (\actelem{\currlayer - 1}{\idxtwo} - \refactelem{\currlayer - 1}{\idxtwo})} &= \weightelem{\currlayer}{\idxtwo}{\idxthree} \frac{\partial \scoresvecnelem{\currlayer - 1}{\poss}{\idxthree}}{\partial (\actelem{\currlayer - 1}{\idxtwo} - \refactelem{\currlayer - 1}{\idxtwo})} = 0 \qquad
					\frac{\partial \scoresvecnelem{\currlayer}{\negs}{\idxthree}}{\partial (\actelem{\currlayer - 1}{\idxtwo} - \refactelem{\currlayer - 1}{\idxtwo})} &= \weightelem{\currlayer}{\idxtwo}{\idxthree} \frac{\partial \scoresvecnelem{\currlayer - 1}{\negs}{\idxthree}}{\partial (\actelem{\currlayer - 1}{\idxtwo} - \refactelem{\currlayer - 1}{\idxtwo})} = -\weightelem{\currlayer}{\idxtwo}{\idxthree}
				\end{align*}
				\item[(iv)] $\weightelem{\currlayer}{\idxtwo}{\idxthree} < 0$ and $\actelem{\currlayer - 1}{\idxtwo} - \refactelem{\currlayer - 1}{\idxtwo} < 0$. Then, by Definitions \ref{def:input-layer-rule} and \ref{def:linear-layer-rule}:
				\begin{align*}
					\frac{\partial \scoresvecnelem{\currlayer}{\poss}{\idxthree}}{\partial (\actelem{\currlayer - 1}{\idxtwo} - \refactelem{\currlayer - 1}{\idxtwo})} &= \weightelem{\currlayer}{\idxtwo}{\idxthree} \frac{\partial \scoresvecnelem{\currlayer - 1}{\negs}{\idxthree}}{\partial (\actelem{\currlayer - 1}{\idxtwo} - \refactelem{\currlayer - 1}{\idxtwo})} = -\weightelem{\currlayer}{\idxtwo}{\idxthree} \qquad
					\frac{\partial \scoresvecnelem{\currlayer}{\negs}{\idxthree}}{\partial (\actelem{\currlayer - 1}{\idxtwo} - \refactelem{\currlayer - 1}{\idxtwo})} &= \weightelem{\currlayer}{\idxtwo}{\idxthree} \frac{\partial \scoresvecnelem{\currlayer - 1}{\poss}{\idxthree}}{\partial (\actelem{\currlayer - 1}{\idxtwo} - \refactelem{\currlayer - 1}{\idxtwo})} = 0
				\end{align*}
			\end{enumerate}
		}
		
		Next, we determine the possible values of $\frac{\partial \scoresvecnelem{\currlayer + 1}{\poss}{\idxthree}}{\partial \scoresvecnelem{\currlayer}{\poss}{\idxthree}}$, $\frac{\partial \scoresvecnelem{\currlayer + 1}{\poss}{\idxthree}}{\partial \scoresvecnelem{\currlayer}{\negs}{\idxthree}}$, $\frac{\partial \scoresvecnelem{\currlayer + 1}{\negs}{\idxthree}}{\partial \scoresvecnelem{\currlayer}{\negs}{\idxthree}}$ and $\frac{\partial \scoresvecnelem{\currlayer + 1}{\negs}{\idxthree}}{\partial \scoresvecnelem{\currlayer}{\poss}{\idxthree}}$. We consider two possible cases:
		
		\begin{enumerate}
			\item[(i)] $\actelem{\currlayer}{\idxthree} > 0$. Then also $\scoresvecnelem{\currlayer}{\poss}{\idxthree} > \scoresvecnelem{\currlayer}{\negs}{\idxthree}$, and by Definitions \ref{def:peak-effects-full}, \ref{def:activation-multipliers} and \ref{def:activation-scores}, we have:
			{\allowdisplaybreaks
				\begin{align*}
					% + → +
					&\frac{\partial \scoresvecnelem{\currlayer + 1}{\poss}{\idxthree}}{\partial \scoresvecnelem{\currlayer}{\poss}{\idxthree}} = \multiplier{\currlayer + 1}{\poss}{\poss}{\idxthree} = \frac{(1 - \cancelconst{}) \linear{\currlayer}{\poss}{\poss}{\idxthree} + \cancelconst{} \peak{\currlayer}{\poss}{\poss}{\idxthree}}{\scoresvecnelem{\currlayer}{\poss}{\idxthree} + \epsilon} = \frac{\peak{\currlayer}{\poss}{\poss}{\idxthree}}{\scoresvecnelem{\currlayer}{\poss}{\idxthree} + \epsilon} \\
					&= \frac{\relu(\scoresvecnelem{\currlayer}{\poss}{\idxthree} - \scoresvecnelem{\currlayer}{\negs}{\idxthree}) - \relu(0) + \max(\relu(\scoresvecnelem{\currlayer}{\poss}{\idxthree}) - \relu(\scoresvecnelem{\currlayer}{\poss}{\idxthree} - \scoresvecnelem{\currlayer}{\negs}{\idxthree}), 0)}{\scoresvecnelem{\currlayer}{\poss}{\idxthree} + \epsilon} \\
					&= \frac{\relu(\scoresvecnelem{\currlayer}{\poss}{\idxthree})}{\scoresvecnelem{\currlayer}{\poss}{\idxthree} + \epsilon} = \frac{\scoresvecnelem{\currlayer}{\poss}{\idxthree}}{\scoresvecnelem{\currlayer}{\poss}{\idxthree} + \epsilon} \approx 1 \\
					% - → +
					&\frac{\partial \scoresvecnelem{\currlayer + 1}{\poss}{\idxthree}}{\partial \scoresvecnelem{\currlayer}{\negs}{\idxthree}} = \multiplier{\currlayer + 1}{\negs}{\poss}{\idxthree} = \frac{(1 - \cancelconst{}) \linear{\currlayer}{\negs}{\poss}{\idxthree} + \cancelconst{} \peak{\currlayer}{\negs}{\poss}{\idxthree}}{\scoresvecnelem{\currlayer}{\negs}{\idxthree} + \epsilon} = \frac{\peak{\currlayer}{\negs}{\poss}{\idxthree}}{\scoresvecnelem{\currlayer}{\negs}{\idxthree} + \epsilon} = \frac{0 + \max(0, 0)}{\scoresvecnelem{\currlayer}{\negs}{\idxthree} + \epsilon} = 0 \\
					% - → -
					&\frac{\partial \scoresvecnelem{\currlayer + 1}{\negs}{\idxthree}}{\partial \scoresvecnelem{\currlayer}{\negs}{\idxthree}} = \multiplier{\currlayer + 1}{\negs}{\negs}{\idxthree} = \frac{(1 - \cancelconst{}) \linear{\currlayer}{\negs}{\negs}{\idxthree} + \cancelconst{} \peak{\currlayer}{\negs}{\negs}{\idxthree}}{\scoresvecnelem{\currlayer}{\negs}{\idxthree} + \epsilon} = \frac{\peak{\currlayer}{\negs}{\negs}{\idxthree}}{\scoresvecnelem{\currlayer}{\negs}{\idxthree} + \epsilon} \\
					&= \frac{0 + \max(\relu(\scoresvecnelem{\currlayer}{\poss}{\idxthree}) - \relu(\scoresvecnelem{\currlayer}{\poss}{\idxthree} - \scoresvecnelem{\currlayer}{\negs}{\idxthree}), 0)}{\scoresvecnelem{\currlayer}{\negs}{\idxthree} + \epsilon} = \frac{\scoresvecnelem{\currlayer}{\negs}{\idxthree}}{\scoresvecnelem{\currlayer}{\negs}{\idxthree} + \epsilon} \approx 1 \\
					% + → -
					&\frac{\partial \scoresvecnelem{\currlayer + 1}{\negs}{\idxthree}}{\partial \scoresvecnelem{\currlayer}{\poss}{\idxthree}} = \multiplier{\currlayer + 1}{\poss}{\negs}{\idxthree} = \frac{(1 - \cancelconst{}) \linear{\currlayer}{\poss}{\negs}{\idxthree} + \cancelconst{} \peak{\currlayer}{\poss}{\negs}{\idxthree}}{\scoresvecnelem{\currlayer}{\poss}{\idxthree} + \epsilon} = \frac{\peak{\currlayer}{\poss}{\negs}{\idxthree}}{\scoresvecnelem{\currlayer}{\poss}{\idxthree} + \epsilon} = \frac{0 + \max(0, 0)}{\scoresvecnelem{\currlayer}{\poss}{\idxthree} + \epsilon} = 0
				\end{align*}
			}
			Note that we used the inequalities from the case assumptions and assumed that $\epsilon$ is negligible.
			\item[(ii)] $\actelem{\currlayer}{\idxthree} \leq 0$. Then also $\scoresvecnelem{\currlayer}{\poss}{\idxthree} \leq \scoresvecnelem{\currlayer}{\negs}{\idxthree}$, and by Definitions \ref{def:peak-effects-full}, \ref{def:activation-multipliers} and \ref{def:activation-scores}, we have:
			{\allowdisplaybreaks
				\begin{align*}
					% + → +
					&\frac{\partial \scoresvecnelem{\currlayer + 1}{\poss}{\idxthree}}{\partial \scoresvecnelem{\currlayer}{\poss}{\idxthree}} = \multiplier{\currlayer + 1}{\poss}{\poss}{\idxthree} = \frac{(1 - \cancelconst{}) \linear{\currlayer}{\poss}{\poss}{\idxthree} + \cancelconst{} \peak{\currlayer}{\poss}{\poss}{\idxthree}}{\scoresvecnelem{\currlayer}{\poss}{\idxthree} + \epsilon} = \frac{\peak{\currlayer}{\poss}{\poss}{\idxthree}}{\scoresvecnelem{\currlayer}{\poss}{\idxthree} + \epsilon} \\
					&= \frac{0 + \max(\relu(\scoresvecnelem{\currlayer}{\poss}{\idxthree}) - \relu(0), 0)}{\scoresvecnelem{\currlayer}{\poss}{\idxthree} + \epsilon}
					= \frac{\relu(\scoresvecnelem{\currlayer}{\poss}{\idxthree})}{\scoresvecnelem{\currlayer}{\poss}{\idxthree} + \epsilon} = \frac{\scoresvecnelem{\currlayer}{\poss}{\idxthree}}{\scoresvecnelem{\currlayer}{\poss}{\idxthree} + \epsilon} \approx 1 \\
					% - → +
					&\frac{\partial \scoresvecnelem{\currlayer + 1}{\poss}{\idxthree}}{\partial \scoresvecnelem{\currlayer}{\negs}{\idxthree}} = \multiplier{\currlayer + 1}{\negs}{\poss}{\idxthree} = \frac{(1 - \cancelconst{}) \linear{\currlayer}{\negs}{\poss}{\idxthree} + \cancelconst{} \peak{\currlayer}{\negs}{\poss}{\idxthree}}{\scoresvecnelem{\currlayer}{\negs}{\idxthree} + \epsilon} = \frac{\peak{\currlayer}{\negs}{\poss}{\idxthree}}{\scoresvecnelem{\currlayer}{\negs}{\idxthree} + \epsilon} = \frac{0 + \max(0, 0)}{\scoresvecnelem{\currlayer}{\negs}{\idxthree} + \epsilon} = 0 \\
					% - → -
					&\frac{\partial \scoresvecnelem{\currlayer + 1}{\negs}{\idxthree}}{\partial \scoresvecnelem{\currlayer}{\negs}{\idxthree}} = \multiplier{\currlayer + 1}{\negs}{\negs}{\idxthree} = \frac{(1 - \cancelconst{}) \linear{\currlayer}{\negs}{\negs}{\idxthree} + \cancelconst{} \peak{\currlayer}{\negs}{\negs}{\idxthree}}{\scoresvecnelem{\currlayer}{\negs}{\idxthree} + \epsilon} = \frac{\peak{\currlayer}{\negs}{\negs}{\idxthree}}{\scoresvecnelem{\currlayer}{\negs}{\idxthree} + \epsilon} \\
					&= \frac{0 + \max(\relu(\scoresvecnelem{\currlayer}{\poss}{\idxthree}) - \relu(0), 0)}{\scoresvecnelem{\currlayer}{\negs}{\idxthree} + \epsilon} = \frac{\scoresvecnelem{\currlayer}{\poss}{\idxthree}}{\scoresvecnelem{\currlayer}{\negs}{\idxthree} + \epsilon} \approx \frac{\scoresvecnelem{\currlayer}{\poss}{\idxthree}}{\scoresvecnelem{\currlayer}{\negs}{\idxthree}} \\
					% + → -
					&\frac{\partial \scoresvecnelem{\currlayer + 1}{\negs}{\idxthree}}{\partial \scoresvecnelem{\currlayer}{\poss}{\idxthree}} = \multiplier{\currlayer + 1}{\poss}{\negs}{\idxthree} = \frac{(1 - \cancelconst{}) \linear{\currlayer}{\poss}{\negs}{\idxthree} + \cancelconst{} \peak{\currlayer}{\poss}{\negs}{\idxthree}}{\scoresvecnelem{\currlayer}{\poss}{\idxthree} + \epsilon} = \frac{\peak{\currlayer}{\poss}{\negs}{\idxthree}}{\scoresvecnelem{\currlayer}{\poss}{\idxthree} + \epsilon} = \frac{0 + \max(0, 0)}{\scoresvecnelem{\currlayer}{\poss}{\idxthree} + \epsilon} = 0
				\end{align*}
			}
		\end{enumerate}
		
		Connecting the above results for the intermediate partial derivatives, we consider two final cases:
		\begin{enumerate}
			\item[(i)] $\actelem{\currlayer}{\idxthree} > 0$. Then, per the case analysis above, $\frac{\partial \scoresvecnelem{\currlayer + 1}{\poss}{\idxthree}}{\partial \scoresvecnelem{\currlayer}{\poss}{\idxthree}} = 1$, $\frac{\partial \scoresvecnelem{\currlayer + 1}{\poss}{\idxthree}}{\partial \scoresvecnelem{\currlayer}{\negs}{\idxthree}} = 0$, $\frac{\partial \scoresvecnelem{\currlayer + 1}{\negs}{\idxthree}}{\partial \scoresvecnelem{\currlayer}{\negs}{\idxthree}} = 1$ and $\frac{\partial \scoresvecnelem{\currlayer + 1}{\negs}{\idxthree}}{\partial \scoresvecnelem{\currlayer}{\poss}{\idxthree}} = 0$. Considering the results for $\frac{\partial \scoresvecnelem{\currlayer}{\poss}{\idxthree}}{\partial(\actelem{\currlayer - 1}{\idxtwo} - \refactelem{\currlayer - 1}{\idxtwo})}$ and $\frac{\partial \scoresvecnelem{\currlayer}{\negs}{\idxthree}}{\partial(\actelem{\currlayer - 1}{\idxtwo} - \refactelem{\currlayer - 1}{\idxtwo})}$, we can immediately see that
			\begin{align*}
				\left| \frac{\partial (\scoresvecnelem{\currlayer + 1}{\poss}{\idxthree} - \scoresvecnelem{\currlayer + 1}{\negs}{\idxthree})}{{\partial(\actelem{\currlayer - 1}{\idxtwo} - \refactelem{\currlayer - 1}{\idxtwo})}} \right|
				= \left|\weightelem{\currlayer}{\idxtwo}{\idxthree} \right|
			\end{align*}
			which implies the statement we aimed to prove.
			\item[(ii)] $\actelem{\currlayer}{\idxthree} \leq 0$. Then, by Property \ref{prop:conflict-relu-negative}, $\weightelem{\currlayer}{\idxtwo}{\idxthree} (\actelem{\currlayer - 1}{\idxtwo} - \refactelem{\currlayer - 1}{\idxtwo}) > 0$. According to the case analysis above, this means that $\left|\frac{\partial \scoresvecnelem{\currlayer}{\poss}{\idxthree}}{\partial(\actelem{\currlayer - 1}{\idxtwo} - \refactelem{\currlayer - 1}{\idxtwo})}\right| = \left|\weightelem{\currlayer}{\idxtwo}{\idxthree}\right|$ and $\frac{\partial \scoresvecnelem{\currlayer}{\negs}{\idxthree}}{\partial(\actelem{\currlayer - 1}{\idxtwo} - \refactelem{\currlayer - 1}{\idxtwo})} = 0$. Given that $\frac{\partial \scoresvecnelem{\currlayer + 1}{\poss}{\idxthree}}{\partial \scoresvecnelem{\currlayer}{\poss}{\idxthree}} = 1$ and $\frac{\partial \scoresvecnelem{\currlayer + 1}{\negs}{\idxthree}}{\partial \scoresvecnelem{\currlayer}{\poss}{\idxthree}} = 0$, we obtain
			\begin{align*}
				\left|\frac{\partial (\scoresvecnelem{\currlayer + 1}{\poss}{\idxthree} - \scoresvecnelem{\currlayer + 1}{\negs}{\idxthree})}{{\partial(\actelem{\currlayer - 1}{\idxtwo} - \refactelem{\currlayer - 1}{\idxtwo})}}\right|
				= \weightelem{\currlayer}{\idxtwo}{\idxthree}
			\end{align*}
			which also implies the statement we aimed to prove.
		\end{enumerate}
		We have now shown that $|\sum_{\idxtwo \in C'} \scoresvecelem{\combs}{\idxtwo}| = |\scoresvecelem{\combs}{\idxtwo}| \geq \argmax_{C \in S^C} \cmag(\model[\currlayer:\currlayer + 1]_{\idxthree}, C, C', \actelem{\currlayer - 1}{\idxthree}, \refactvec{\currlayer - 1})$. It remains to be shown that $\forall C \in S^C, \sgn(\sum_{\idxone \in C} \scoresvecelem{\combs}{\idxone}) = - \sgn(\scoresvecelem{\combs}{\idxtwo})$. Take an arbitrary feature set $C$ from $S^C$ and an arbitrary feature $\idxone$ from $C$. By Properties \ref{prop:conflict-relu-positive} and \ref{prop:conflict-relu-negative}, we have $\sgn(\weightelem{\currlayer}{\idxone}{\idxthree} (\actelem{\currlayer - 1}{\idxone} - \refactelem{\currlayer - 1}{\idxone})) = - \sgn(\weightelem{\currlayer}{\idxtwo}{\idxthree} (\actelem{\currlayer - 1}{\idxtwo} - \refactelem{\currlayer - 1}{\idxtwo}))$. Without loss of generality, assume that $\weightelem{\currlayer}{\idxtwo}{\idxthree} (\actelem{\currlayer - 1}{\idxtwo} - \refactelem{\currlayer - 1}{\idxtwo}) > 0 > \weightelem{\currlayer}{\idxone}{\idxthree} (\actelem{\currlayer - 1}{\idxone} - \refactelem{\currlayer - 1}{\idxone})$. We consider eight possible cases, using the results for partial derivatives that we obtained above (note that these are identical for $\idxone$):
		\begin{enumerate}
			\item[(i)] $(\actelem{\currlayer - 1}{\idxtwo} - \refactelem{\currlayer - 1}{\idxtwo}) > 0$, $(\actelem{\currlayer - 1}{\idxone} - \refactelem{\currlayer - 1}{\idxone}) > 0$ and $\actelem{\currlayer}{\idxthree} > 0$. This means that $\weightelem{\currlayer}{\idxtwo}{\idxthree} > 0$ and $\weightelem{\currlayer}{\idxone}{\idxthree} < 0$. By the above results for partial derivatives, we obtain:
			\begin{align*}
				\scoresvecelem{\combs}{\idxtwo} = \weightelem{\currlayer}{\idxtwo}{\idxthree} (\actelem{\currlayer - 1}{\idxtwo} - \refactelem{\currlayer - 1}{\idxtwo}) > 0 \qquad
				\scoresvecelem{\combs}{\idxone} = -\weightelem{\currlayer}{\idxone}{\idxthree} (\actelem{\currlayer - 1}{\idxtwo} - \refactelem{\currlayer - 1}{\idxtwo}) < 0
			\end{align*}
			\item[(ii)] $(\actelem{\currlayer - 1}{\idxtwo} - \refactelem{\currlayer - 1}{\idxtwo}) < 0$, $(\actelem{\currlayer - 1}{\idxone} - \refactelem{\currlayer - 1}{\idxone}) > 0$ and $\actelem{\currlayer}{\idxthree} > 0$. This means that $\weightelem{\currlayer}{\idxtwo}{\idxthree} < 0$ and $\weightelem{\currlayer}{\idxone}{\idxthree} < 0$. By the above results for partial derivatives, we obtain:
			\begin{align*}
				\scoresvecelem{\combs}{\idxtwo} = -\weightelem{\currlayer}{\idxtwo}{\idxthree} (\actelem{\currlayer - 1}{\idxtwo} - \refactelem{\currlayer - 1}{\idxtwo}) > 0 \qquad
				\scoresvecelem{\combs}{\idxone} = -\weightelem{\currlayer}{\idxone}{\idxthree} (\actelem{\currlayer - 1}{\idxtwo} - \refactelem{\currlayer - 1}{\idxtwo}) < 0
			\end{align*}
			\item[(iii)] $(\actelem{\currlayer - 1}{\idxtwo} - \refactelem{\currlayer - 1}{\idxtwo}) > 0$, $(\actelem{\currlayer - 1}{\idxone} - \refactelem{\currlayer - 1}{\idxone}) < 0$ and $\actelem{\currlayer}{\idxthree} > 0$. This means that $\weightelem{\currlayer}{\idxtwo}{\idxthree} > 0$ and $\weightelem{\currlayer}{\idxone}{\idxthree} > 0$. By the above results for partial derivatives, we obtain:
			\begin{align*}
				\scoresvecelem{\combs}{\idxtwo} = \weightelem{\currlayer}{\idxtwo}{\idxthree} (\actelem{\currlayer - 1}{\idxtwo} - \refactelem{\currlayer - 1}{\idxtwo}) > 0 \qquad
				\scoresvecelem{\combs}{\idxone} = \weightelem{\currlayer}{\idxone}{\idxthree} (\actelem{\currlayer - 1}{\idxtwo} - \refactelem{\currlayer - 1}{\idxtwo}) < 0
			\end{align*}
			\item[(iv)] $(\actelem{\currlayer - 1}{\idxtwo} - \refactelem{\currlayer - 1}{\idxtwo}) < 0$, $(\actelem{\currlayer - 1}{\idxone} - \refactelem{\currlayer - 1}{\idxone}) < 0$ and $\actelem{\currlayer}{\idxthree} > 0$. This means that $\weightelem{\currlayer}{\idxtwo}{\idxthree} < 0$ and $\weightelem{\currlayer}{\idxone}{\idxthree} > 0$. By the above results for partial derivatives, we obtain:
			\begin{align*}
				\scoresvecelem{\combs}{\idxtwo} = -\weightelem{\currlayer}{\idxtwo}{\idxthree} (\actelem{\currlayer - 1}{\idxtwo} - \refactelem{\currlayer - 1}{\idxtwo}) > 0 \qquad
				\scoresvecelem{\combs}{\idxone} = \weightelem{\currlayer}{\idxone}{\idxthree} (\actelem{\currlayer - 1}{\idxtwo} - \refactelem{\currlayer - 1}{\idxtwo}) < 0
			\end{align*}
			%%%%%%%%%%
			\item[(v)] $(\actelem{\currlayer - 1}{\idxtwo} - \refactelem{\currlayer - 1}{\idxtwo}) > 0$, $(\actelem{\currlayer - 1}{\idxone} - \refactelem{\currlayer - 1}{\idxone}) > 0$ and $\actelem{\currlayer}{\idxthree} \leq 0$. This means that $\weightelem{\currlayer}{\idxtwo}{\idxthree} > 0$ and $\weightelem{\currlayer}{\idxone}{\idxthree} < 0$. By the above results for partial derivatives, we obtain:
			\begin{align*}
				\scoresvecelem{\combs}{\idxtwo} = \weightelem{\currlayer}{\idxtwo}{\idxthree} (\actelem{\currlayer - 1}{\idxtwo} - \refactelem{\currlayer - 1}{\idxtwo}) > 0 \qquad
				\scoresvecelem{\combs}{\idxone} = -\frac{\scoresvecnelem{\currlayer}{\poss}{\idxthree}}{\scoresvecnelem{\currlayer}{\negs}{\idxthree}} \weightelem{\currlayer}{\idxone}{\idxthree} (\actelem{\currlayer - 1}{\idxtwo} - \refactelem{\currlayer - 1}{\idxtwo}) < 0
			\end{align*}
			\item[(vi)] $(\actelem{\currlayer - 1}{\idxtwo} - \refactelem{\currlayer - 1}{\idxtwo}) < 0$, $(\actelem{\currlayer - 1}{\idxone} - \refactelem{\currlayer - 1}{\idxone}) > 0$ and $\actelem{\currlayer}{\idxthree} \leq 0$. This means that $\weightelem{\currlayer}{\idxtwo}{\idxthree} < 0$ and $\weightelem{\currlayer}{\idxone}{\idxthree} < 0$. By the above results for partial derivatives, we obtain:
			\begin{align*}
				\scoresvecelem{\combs}{\idxtwo} = -\weightelem{\currlayer}{\idxtwo}{\idxthree} (\actelem{\currlayer - 1}{\idxtwo} - \refactelem{\currlayer - 1}{\idxtwo}) > 0 \qquad
				\scoresvecelem{\combs}{\idxone} = -\frac{\scoresvecnelem{\currlayer}{\poss}{\idxthree}}{\scoresvecnelem{\currlayer}{\negs}{\idxthree}} \weightelem{\currlayer}{\idxone}{\idxthree} (\actelem{\currlayer - 1}{\idxtwo} - \refactelem{\currlayer - 1}{\idxtwo}) < 0
			\end{align*}
			\item[(vii)] $(\actelem{\currlayer - 1}{\idxtwo} - \refactelem{\currlayer - 1}{\idxtwo}) > 0$, $(\actelem{\currlayer - 1}{\idxone} - \refactelem{\currlayer - 1}{\idxone}) < 0$ and $\actelem{\currlayer}{\idxthree} \leq 0$. This means that $\weightelem{\currlayer}{\idxtwo}{\idxthree} > 0$ and $\weightelem{\currlayer}{\idxone}{\idxthree} > 0$. By the above results for partial derivatives, we obtain:
			\begin{align*}
				\scoresvecelem{\combs}{\idxtwo} = \weightelem{\currlayer}{\idxtwo}{\idxthree} (\actelem{\currlayer - 1}{\idxtwo} - \refactelem{\currlayer - 1}{\idxtwo}) > 0 \qquad
				\scoresvecelem{\combs}{\idxone} = \frac{\scoresvecnelem{\currlayer}{\poss}{\idxthree}}{\scoresvecnelem{\currlayer}{\negs}{\idxthree}} \weightelem{\currlayer}{\idxone}{\idxthree} (\actelem{\currlayer - 1}{\idxtwo} - \refactelem{\currlayer - 1}{\idxtwo}) < 0
			\end{align*}
			\item[(viii)] $(\actelem{\currlayer - 1}{\idxtwo} - \refactelem{\currlayer - 1}{\idxtwo}) < 0$, $(\actelem{\currlayer - 1}{\idxone} - \refactelem{\currlayer - 1}{\idxone}) < 0$ and $\actelem{\currlayer}{\idxthree} \leq 0$. This means that $\weightelem{\currlayer}{\idxtwo}{\idxthree} < 0$ and $\weightelem{\currlayer}{\idxone}{\idxthree} > 0$. By the above results for partial derivatives, we obtain:
			\begin{align*}
				\scoresvecelem{\combs}{\idxtwo} = -\weightelem{\currlayer}{\idxtwo}{\idxthree} (\actelem{\currlayer - 1}{\idxtwo} - \refactelem{\currlayer - 1}{\idxtwo}) > 0 \qquad
				\scoresvecelem{\combs}{\idxone} = \frac{\scoresvecnelem{\currlayer}{\poss}{\idxthree}}{\scoresvecnelem{\currlayer}{\negs}{\idxthree}} \weightelem{\currlayer}{\idxone}{\idxthree} (\actelem{\currlayer - 1}{\idxtwo} - \refactelem{\currlayer - 1}{\idxtwo}) < 0
			\end{align*}
		\end{enumerate}
		As we can observe, in all the possible cases, and for any $\idxone \in C$ and $C \in S^C$, $\sgn(\scoresvecelem{\combs}{\idxone}) = - \sgn(\scoresvecelem{\combs}{\idxtwo})$. This also means that $\forall C \in S^C, \sgn(\sum_{\idxone \in C} \scoresvecelem{\combs}{\idxone}) = - \sgn(\scoresvecelem{\combs}{\idxtwo})$, as required. Thus, we have shown that CAFE with $c = 1$ computed with respect to a reference input of zero does not underreport any minimal local conflicts for ReLU neurons, which concludes the proof. \aptLtoX[graphic=no,type=html]{&#x25A1;}{{$\square$}}
	\end{myproof}
	
	\subsection{Computational Complexity}
	\label{sec:comp-complexity-runtimes}
	Apart from the asymptotic complexity outlined in the main text, we also evaluate the computational efficiency of CAFE in practice by measuring its runtimes in our experiments and comparing them with other methods. The results are captured in Tables \ref{tab:real-data-runtimes} and \ref{tab:ft-real-data-runtimes}. We can observe that the runtimes of CAFE are roughly comparable with other gradient-based methods while being significantly better than the runtimes of more complex approaches, especially those requiring extensive sampling. Note that we did not make any significant effort to optimize our implementation of CAFE for speed, so it is possible that its performance could be further improved.
	
	\begin{table*}[!t]
		\caption{Absolute runtimes for different attribution methods applied to MLPs trained on various datasets. Best results in bold, second-best results are underlined. \faHourglassHalf{} marks methods more than 10-times slower than CAFE on average.}
		\label{tab:real-data-runtimes}
		\footnotesize
		\centering
		\begin{center}
			\begin{tabular}{ c c c c c c c c }
				\toprule
				\multirow{3}{*}{\textbf{Method}} & \multicolumn{7}{c}{\textbf{Runtime (Seconds) ($\downarrow$)}} \\
				\cmidrule(r){2-8}
				& \textbf{COMPAS} & \textbf{HELOC} & \textbf{Adult} &\textbf{German} & \textbf{Titanic} & \textbf{MIMIC-IV} & \textbf{Covertype} \\
				\midrule
				Gradient $\cdot$ Input & 0.012 & 0.029 & 0.202 & 0.002 & 0.005 & 0.078 & 0.221 \\
				LRP & 0.018 & 0.029 & 0.107 & 0.006 & 0.008 & 0.123 & 0.431 \\
				DeepLIFT Rescale & 0.041 & 0.051 & 0.201 & 0.008 & 0.010 & 0.182 & 0.554 \\
				GradientSHAP & 0.049 & 0.139 & 0.288 & 0.021 & 0.014 & 0.252 & 0.708 \\
				Integrated Gradients & 0.491 & 0.719 & 3.483 & 0.066 & 0.062 & 1.750 & 4.448 \\
				SmoothGrad & 0.039 & 0.091 & 0.398 & 0.009 & 0.012 & 0.219 & 0.580 \\
				KernelSHAP \faHourglassHalf{} & 14.539 & 50.157 & 87.740 & 4.266 & 4.139 & 46.606 & 128.514 \\
				Shapley Value Sampling \faHourglassHalf{} & 0.164 & 2.059 & 2.015 & 0.338 & 0.159 & 3.996 & 16.275 \\
				LIME \faHourglassHalf{} & 9.382 & 68.264 & 90.473 & 5.077 & 5.636 & 56.109 & 183.548 \\
				\midrule
				CAFE ($c = 0.0$) & 0.040 & 0.114 & 0.214 & 0.015 & 0.021 & 0.366 & 1.475 \\
				CAFE ($c = 0.5$) & 0.032 & 0.061 & 0.265 & 0.007 & 0.020 & 0.364 & 1.520 \\
				CAFE ($c = 1.0$) & 0.086 & 0.142 & 0.221 & 0.009 & 0.020 & 0.367 & 1.493 \\
				\bottomrule
			\end{tabular}
		\end{center}
	\end{table*}
	
	\begin{table*}[!t]
		\caption{Absolute runtimes for different attribution methods applied to FT-Transformers trained on various datasets. Best results in bold, second-best results are underlined. \faHourglassHalf{} marks methods more than 10-times slower than CAFE on average.}
		\label{tab:ft-real-data-runtimes}
		\footnotesize
		\centering
		\begin{center}
			\begin{tabular}{ c c c c c c c c }
				\toprule
				\multirow{3}{*}{\textbf{Method}} & \multicolumn{7}{c}{\textbf{Runtime (Seconds) ($\downarrow$)}} \\
				\cmidrule(r){2-8}
				& \textbf{COMPAS} & \textbf{HELOC} & \textbf{Adult} &\textbf{German} & \textbf{Titanic} & \textbf{MIMIC-IV} & \textbf{Covertype} \\
				\midrule
				Gradient $\cdot$ Input & 0.088 & 0.193 & 0.393 & 0.037 & 0.069 & 0.518 & 1.014 \\
				DeepLIFT Rescale & 0.089 & 0.324 & 0.785 & 0.075 & 0.103 & 0.991 & 2.066 \\
				GradientSHAP & 0.158 & 0.455 & 1.125 & 0.061 & 0.075 & 0.716 & 1.406 \\
				Integrated Gradients & 1.215 & 4.125 & 11.647 & 0.417 & 0.161 & 2.780 & 6.480 \\
				SmoothGrad & 0.117 & 0.499 & 1.224 & 0.099 & 0.074 & 0.651 & 1.271 \\
				Shapley Value Sampling \faHourglassHalf{} & 1.727 & 27.478 & 33.437 & 7.793 & 3.609 & 40.375 & 85.793 \\
				\midrule
				CAFE ($c = 0.0$) & 0.171 & 0.630 & 1.654 & 0.100 & 0.207 & 2.144 & 4.252 \\
				CAFE ($c = 0.5$) & 0.159 & 0.657 & 1.639 & 0.100 & 0.157 & 1.888 & 4.115 \\
				CAFE ($c = 1.0$) & 0.162 & 0.634 & 1.668 & 0.115 & 0.165 & 1.875 & 3.865 \\
				\bottomrule
			\end{tabular}
		\end{center}
	\end{table*}
	
	\subsection{Comparison to Related Gradient-Based Methods}
	\label{sec:cafe-comparison}
	\begin{table*}[!tp]
		\caption{Qualitative comparison between CAFE and related attribution methods. The used method abbreviations are the same as in the beginning of Section \ref{sec:introduction}. Details on the considered properties are given in Appendix \ref{sec:cafe-comparison}.}
		\label{tab:cafe-comparison}
		\centering
		\scriptsize
		\begin{center}
			\begin{tabular}{ c c c c c c c } 
				\toprule
				\multirow{2}{*}{\textbf{Property}} & \multicolumn{6}{c}{\textbf{Method}} \\
				\cmidrule(r){2-7}
				& \textbf{G $\cdot$ I} & \textbf{LRP} & \textbf{DL-R} & \textbf{DL-RC} & \textbf{IG} & \textbf{CAFE} \\
				\midrule
				Missingness & \cmark & \cmark & \cmark & \cmark & \cmark & \cmark \\
				Linearity & \cmark & \cmark & \cmark & \cmark & \cmark & \cmark \\
				Completeness & \xmark & \cmark & \cmark & \cmark & \cmark & \cmark \\
				Implementation invariance & \cmark & \xmark & \xmark & \xmark & \cmark & \xmark \\
				Exactly computable without sampling & \cmark & \cmark & \cmark & \cmark & \xmark & \cmark \\
				Handles $\simpleactfun(0) \neq 0$ nonlinearities & \cmark & \xmark & \cmark & \cmark & \cmark & \cmark \\
				Accepts reference input & \xmark & \xmark & \cmark & \cmark & \cmark & \cmark \\
				Safeguards against attribution explosion & \xmark & \xmark & \xmark & \cmark & \xmark & \cmark \\
				Considers conflicts & \xmark & \xmark & \xmark & \cmark & \xmark & \cmark \\
				Local conflict awareness for ReLU & \xmark & \xmark & \xmark & \xmark & \xmark & \cmark \\
				Adjustable conflict sensitivity & \xmark & \xmark & \xmark & \xmark & \xmark & \cmark \\
				Separates $+$ and $-$ feature attributions & \xmark & \xmark & \xmark & \xmark & \xmark & \cmark \\
				Captures effects of biases & \xmark & \xmark & \xmark & \xmark & \xmark & \cmark \\
				\bottomrule
			\end{tabular}
		\end{center}
	\end{table*}
	
	To better position CAFE with respect to prior work and to provide guidance to users who may be deciding between multiple different feature attribution methods, we compare the properties of CAFE to several other relevant gradient-based attribution methods. In particular, we consider Gradient $\cdot$ Input (G $\cdot$ I) \cite{shrikumar-old-deeplift-gxi}, LRP \cite{bach-on-pixel-wise-lrp, montavon-lrp-overview}, DeepLIFT Rescale (DL-R)  and DeepLIFT RevealCancel (DL-RC) \cite{shrikumar-deeplift}, and Integrated Gradients (IG) \cite{sundararajan-integrated-gradients}. An overview of our comparison is presented in Table \ref{tab:cafe-comparison}. We provide further details about each of the considered properties below:
	
	\paragraph{Missingness, Linearity and Completeness} We have already explained these properties and shown that CAFE satisfies them in the above sections. Most of the other gradient-based approaches also satisfy these desiderata. A notable exception is Gradient $\cdot$ Input, which does not satisfy completeness. This is because it only considers the local gradient at a single point, which may be arbitrarily large or small regardless of the model output.
	
	\paragraph{Implementation Invariance} Implementation invariance was originally introduced as a desirable property for Integrated Gradients \cite{sundararajan-integrated-gradients} and also holds for Gradient $\cdot$ Input explanations. The property stipulates that the attribution scores for two functions should be identical if these functions return the same value for all inputs. While such property may seem intuitive, it is at odds with the design goals of CAFE, which aims to identify conflicting features even if they ultimately don't change the output of the model. Thus, CAFE doesn't satisfy implementation invariance, similarly to several other attribution methods.
	
	\paragraph{Exact Computability Without Sampling} CAFE, along with the majority of the other considered methods, is computable with a single forward-backward pass through the considered model (respectively two passes, if we additionally consider the computation of the activations for the reference input for methods that use such an input, although these activations can also be computed simultaneously during the forward pass), and thus does not require any sampling. The only considered method that does not satisfy this are the Integrated Gradients, which require sampling to approximate the mean gradient of the considered model along the path from the reference input to the current input.
	
	\paragraph{Handling of $\simpleactfun(0) \neq 0$ Nonlinearities} CAFE, as well as most of the other considered feature attribution approaches, is largely agnostic to the choice of the activation function (although some methods may still behave non-intuitively for non-monotonic or piecewise-linear functions, as we demonstrated on several examples above). The LRP method is an exception, as it was specifically designed for networks using ReLU or tanh activations. Thus, as was earlier noted by \cite{ancona-towards-attribution}, it may return erratic scores for models using activation functions that return non-zero values for $x = 0$, such as sigmoid or softplus.
	
	\paragraph{Reference Input} As described above, CAFE explanations can be computed with respect to a specified reference input. The attribution scores returned in this case can be seen as explaining the differences between the behaviour of the model for the reference input and the current input at hand. Integrated Gradients and both variants of DeepLIFT also accept reference input, while Gradient $\cdot$ Input and LRP do not. Gradient $\cdot$ Input only estimates the behaviour of the model at a single point, resulting in potentially less precise and noisier attribution scores, while LRP can be argued to be implicitly using a reference input of $\textbf{0}$.
	
	\paragraph{Safeguards Against Attribution Score Explosion} As we explained in the main text, CAFE prevents the attribution scores from becoming excessively large or low by checking whether the output estimated by a linear approximation of the activation function is achievable. Meanwhile, as we have shown in several examples, most of the alternative attribution methods are relatively sensitive to this phenomenon and can easily return scores of excessive magnitudes. The only exception is DeepLIFT RevealCancel, which computes attribution scores solely based on the activation function outputs for different values, which eliminates the issue. As far as we are aware, CAFE is the only method capable of ignoring feature conflicts that safeguards against attribution score explosion.
	
	\paragraph{Consideration of conflicts} One of the primary goals of CAFE is to reliably capture the effects of canceled features if the user commands so by choosing suitably large values for the $\cancelconst{\currlayer}$ constants. This is in contrast to the majority of alternative feature attribution methods, which typically ignore the effects of conflicting features altogether, unless they are not already captured in the gradient information. This is problematic for functions such as ReLU or GELU that are inactive or non-monotonic on parts of their domain. As far as we are aware, the only other attribution method that explicitly strives to capture canceled features is DeepLIFT RevealCancel. While it performs better in this sense than other approaches, we argue that it is still limited in this regard, as we have demonstrated on the examples in the main text. Nevertheless, we still mark DeepLIFT RevealCancel as attempting to consider conflicts.
	
	\paragraph{Local conflict awareness for ReLU} In addition to the general aim of surfacing conflicting or canceled features considered under the previous point, our work has formally defined the notion of local conflict awareness (Property \ref{def:local-conflict-awareness}), which can be used to determine whether a feature attribution method is always successful in correctly representing local conflicts on a certain type of models. While CAFE satisfies this property for ReLU networks as shown by Theorem \ref{theo:cafe-conflict-awareness}, DeepLIFT RevealCancel does not, as it still underestimated the effects of conflicting features. This is shown on the example in Appendix \ref{apd:real-world-conflict}.
	
	\paragraph{Adjustable Conflict Sensitivity} As much as the capability to capture conflicts is desirable, there may be cases in which we wish to obtain more focused and abstract attribution scores partially ignoring the conflicting features. For example, we may opt to largely ignore conflicts in the earlier layers of a deep learning model, as these layers typically operate on more noisy, lower-level features, while taking into account conflicts in the later layers, which may reason in more high-level and human-understandable terms. Past research has suggested that adapting the propagation rules per layer may be beneficial \cite{montavon-lrp-overview}. CAFE enables users to do so seamlessly by simply adjusting the $\cancelconst{\currlayer}$ parameters, controlling how sensitive the attributions are to the canceled effects. To the best of our knowledge, no other feature attribution method enables these kinds of adjustments.
	
	\paragraph{Separate Positive and Negative Feature Attributions} CAFE is also unique in separately identifying the positive and negative contributions of each feature. This may provide more rich information in cases in which the conflicts between features are more complex or when a single feature contributes both positively and negatively (e.g., via two separate neurons). When desired, the positive and negative attributions can still be combined into a single joint score by computing the operation $\scoresvec^\combs = \scoresvec^\poss - \scoresvec^\negs$.
	
	\paragraph{Bias Effects} Most of the existing attribution methods ignore the effects of the network bias terms on the output. This may result in the generation of misleading scores in cases in which the output of the model is mainly or partially driven by the biases. Additionally, the interactions between the bias terms and the input features may further distort the returned attributions. In contrast, CAFE explicitly considers biases in its rules and computes separate attribution scores $s_{\text{bias}}^+$ and $s_{\text{bias}}^-$ indicating their combined effects on the output.
	
	\section{Experimental Details}
	\label{sec:experiment-details}
	In this section, we provide additional details regarding our experiments. All our experimental evaluation was performed on a standard x86\_64 machine running a customised distribution of Ubuntu 22.04.2 with NVIDIA GeForce RTX 4090 graphical accelerators. In all experiments, we used the baseline attribution method implementations available in Captum \cite{kokhlikyan-captum} with their default parameters. The used perturbation-based methods were additionally provided with a feature mask specifying the one-hot-encoded categorical features. The infidelity and max-sensitivity metrics were computed directly using Captum, structural infidelity was implemented using Captum's input-output infidelity as a backbone and complexity was evaluated using its implementation from Quantus \cite{hedstrom-quantus}. For computing model performance metrics, we used the metric implementations from TorchEval\footnote{\url{https://pytorch.org/torcheval/stable/}}, with the exception of RMSE, which was computed using scikit-learn \cite{pedregosa-scikit-learn}. For computing means and other statistical measures, we used the respective methods from NumPy\footnote{\url{https://numpy.org/}}. Code for all our experiments is available at \url{https://github.com/adamdejl/hidden-conflicts}, including all the hyperparameters and random seeds necessary for reproducing the results.
	
	\subsection{Synthetic Data Experiments}
	\label{apd:synthetic-details}
	
	\begin{table*}[!tb]
		\caption{Performance of various attribution methods on trained (Train.) models with conflicting features. Results in each row are averaged over five different randomly initialised datasets and (in case of trained models) neural models.}
		\label{tab:synthetic-data-results-train}
		\footnotesize
		\centering
		\begin{tabular}{ c c c c c c c c c }
			\toprule
			\textbf{Configuration / Method} & \multicolumn{8}{c}{\textbf{Value / Attribution RMSE ($\downarrow$)}} \\
			\midrule
			Model type & Train. & Train. & Train. & Train. & Train. & Train. & Train. & Train. \\
			Continuous data dimension & 2 & 3 & 4 & 5 & 2 & 3 & 4 & 5 \\
			Conflict likelihood & 0.30 & 0.25 & 0.20 & 0.15 & 0.30 & 0.25 & 0.20 & 0.15 \\
			Hidden layer sizes & 16, 16 & 24, 24 & 32, 32 & 40, 40 & 16, 16 & 24, 24 & 32, 32 & 40, 40 \\
			Activation function & ReLU & ReLU & ReLU & ReLU & GELU & GELU & GELU & GELU \\
			Validation RMSE & 0.03 & 0.02 & 0.10 & 0.09 & 0.02 & 0.03 & 0.09 & 0.09 \\
			Test RMSE & 0.07 & 0.04 & 0.08 & 0.11 & 0.06 & 0.04 & 0.09 & 0.12 \\
			\midrule
			Gradient $\cdot$ Input & 3.27 & 2.66 & 2.57 & 2.13 & 3.28 & 2.68 & 2.55 & 2.14 \\
			LRP & 3.27 & 2.66 & 2.57 & 2.13 & 3.14 & 2.56 & 2.37 & 2.05 \\
			DeepLIFT Rescale & 3.19 & 2.57 & 2.46 & 2.04 & 3.07 & 2.50 & 2.24 & 2.02 \\
			Integrated Gradients & 3.18 & 2.59 & 2.47 & 2.06 & 3.07 & 2.52 & 2.28 & 2.03 \\
			SmoothGrad & 4.86 & 3.96 & 4.33 & 4.25 & 4.89 & 3.94 & 4.27 & 2.26 \\
			Gradient SHAP & 3.19 & 2.59 & 2.49 & 2.07 & 3.09 & 2.53 & 2.30 & 2.04 \\
			Kernel SHAP & 1.81 & 1.60 & 1.66 & 1.53 & 1.81 & 1.60 & 1.66 & 1.53 \\
			Shapley Value Sampling & 1.59 & 1.27 & 1.29 & 1.12 & 1.59 & 1.27 & 1.29 & 1.13 \\
			LIME & 1.69 & 1.49 & 1.51 & 1.33 & 1.69 & 1.48 & 1.51 & 1.33 \\
			\midrule
			CAFE ($c = 0.00$) & 3.27 & 2.66 & 2.57 & 2.13 & 3.14 & 2.56 & 2.37 & 2.05 \\
			CAFE ($c = 0.50$) & 1.85 & 1.44 & 1.39 & 1.13 & 1.71 & 1.29 & 1.24 & 1.03 \\
			CAFE ($c = 1.00$) & \textbf{0.35} & \textbf{0.43} & \textbf{0.52} & \textbf{0.42} & \textbf{0.74} & \textbf{0.60} & \textbf{1.01} & \textbf{0.88} \\
			\bottomrule
		\end{tabular}
	\end{table*}
	
	\begin{table*}[!tb]
		\caption{Performance of various attribution methods on procedurally constructed (Proc.) synthetic models with conflicting features. Results in each row are averaged over five different randomly initialised datasets and (in case of trained models) neural models.}
		\label{tab:synthetic-data-results-proc}
		\footnotesize
		\centering
		\begin{tabular}{ c c c c c c c }
			\toprule
			\textbf{Configuration / Method} & \multicolumn{6}{c}{\textbf{Value / Attribution RMSE ($\downarrow$)}} \\
			\midrule
			Model type & Proc. & Proc. & Proc. & Proc. & Proc. & Proc. \\
			Continuous data dim & 2 & 3 & 4 & 5 & 100 & 1000 \\
			Conflict likelihood & 0.30 & 0.25 & 0.20 & 0.15 & 0.05 & 0.01 \\
			Hidden layer sizes & 6, 4 & 9, 6 & 12, 8 & 15, 10 & 300, 200 & 3000, 2000 \\
			Activation function & ReLU & ReLU & ReLU & ReLU & ReLU & ReLU \\
			Validation RMSE & 0.00 & 0.00 & 0.00 & 0.00 & 0.00 & 0.00 \\
			Test RMSE & 0.00 & 0.00 & 0.00 & 0.00 & 0.00 & 0.00 \\
			\midrule
			Gradient $\cdot$ Input & 3.31 & 2.69 & 2.62 & 2.18 & 1.14 & 0.5 \\
			LRP & 3.31 & 2.69 & 2.62 & 2.18 & 1.14 & 0.5 \\
			DeepLIFT Rescale & 3.31 & 2.69 & 2.62 & 2.18 & 1.14 & 0.5 \\
			Integrated Gradients & 3.31 & 2.69 & 2.62 & 2.18 & 1.14 & 0.5 \\
			SmoothGrad & 4.98 & 4.08 & 4.34 & 4.17 & 3.36 & 2.92 \\
			Gradient SHAP & 3.31 & 2.69 & 2.62 & 2.18 & 1.14 & 0.5 \\
			Kernel SHAP & 1.81 & 1.60 & 1.66 & 1.53 & 3.24 & 2.84 \\
			Shapley Value Sampling & 1.59 & 1.27 & 1.29 & 1.12 & 0.57 & 0.25 \\
			LIME & 1.69 & 1.49 & 1.51 & 1.33 & 3.59 & 3.09 \\
			\midrule
			CAFE ($c = 0.00$) & 3.31 & 2.69 & 2.62 & 2.18 & 1.14 & 0.5 \\
			CAFE ($c = 0.50$) & 1.66 & 1.35 & 1.31 & 1.09 & 0.57 & 0.25 \\
			CAFE ($c = 1.00$) & \textbf{0.00} & \textbf{0.00} & \textbf{0.00} & \textbf{0.00} & \textbf{0.00} & \textbf{0.00} \\
			\bottomrule
		\end{tabular}
	\end{table*}
	
	\begin{table*}[tbp]
		\caption{Structural Infidelity of the attribution methods applied to MLPs, with an identical setup as in Table \ref{tab:real-data-infidelity}.}
		\label{tab:real-data-struct-infidelity}
		\scriptsize
		\centering
		\begin{center}
			\begin{tabular}{ c c c c c c c c c c c c c c c }
				\toprule
				\multirow{4}{*}{\textbf{Method}} & \multicolumn{14}{c}{\textbf{Attribution Structural Infidelity ($\downarrow$)}} \\
				\cmidrule(r){2-15}
				& \multicolumn{2}{c}{\textbf{COMPAS}} & \multicolumn{2}{c}{\textbf{HELOC}} & \multicolumn{2}{c}{\textbf{Adult}} & \multicolumn{2}{c}{\textbf{German}} & \multicolumn{2}{c}{\textbf{Titanic}} & \multicolumn{2}{c}{\textbf{MIMIC-IV}} & \multicolumn{2}{c}{\textbf{Covertype}} \\
				\cmidrule(r){2-3} \cmidrule(r){4-5} \cmidrule(r){6-7} \cmidrule(r){8-9} \cmidrule(r){10-11} \cmidrule(r){12-13} \cmidrule(r){14-15}
				& \textbf{S} & \textbf{L} & \textbf{S} & \textbf{L} & \textbf{S} & \textbf{L} & \textbf{S} & \textbf{L} & \textbf{S} & \textbf{L} & \textbf{S} & \textbf{L} & \textbf{S} & \textbf{L} \\
				\midrule
				Gradient $\cdot$ Input & 6.93 & 17.64 & 7.64 & 20.21 & 14.419 & 27.119 & \textbf{0.013} & \textbf{0.028} & 1.58 & 3.87 & 0.81 & 1.76 & 391.16 & 1148.58 \\
				DeepLIFT Rescale & 6.42 & 16.29 & 7.26 & 19.11 & 14.435 & 27.162 & \textbf{0.013} & \textbf{0.028} & \underline{1.54} & \underline{3.71} & 0.78 & 1.70 & 399.04 & 1159.86 \\
				GradientSHAP & 6.47 & 16.41 & 7.40 & 19.38 & 14.440 & 27.175 & \textbf{0.013} & \textbf{0.028} & 1.56 & 3.76 & 0.80 & 1.74 & 402.37 & 1167.22 \\
				Integrated Gradients & 6.42 & 16.29 & 7.26 & 19.11 & 14.435 & 27.162 & \textbf{0.013} & \textbf{0.028} & \underline{1.54} & \underline{3.71} & 0.78 & 1.71 & 402.18 & 1166.66 \\
				SmoothGrad & 16.60 & 39.71 & 8.97 & 22.54 & 14.748 & 27.585 & 0.029 & 0.062 & 2.14 & 4.87 & 1.17 & 2.36 & 436.43 & 1246.41 \\
				Shapley Value Sampling \faHourglassHalf{} & \textbf{6.36} & 16.13 & 6.53 & 16.70 & 14.442 & 27.175 & \underline{0.027} & \underline{0.052} & 1.69 & 3.94 & \underline{0.77} & \underline{1.69} & \underline{382.37} & \underline{1127.39} \\
				\midrule
				CAFE ($c = 0.0$) & 6.93 & 17.64 & 7.64 & 20.21 & 14.419 & 27.119 & \textbf{0.013} & \textbf{0.028} & 1.58 & 3.87 & 0.81 & 1.76 & \textbf{379.70} & \textbf{1124.19} \\
				CAFE ($c = 0.5$) & \underline{6.40} & \underline{16.11} & \textbf{6.21} & \underline{15.77} & \textbf{14.416} & \textbf{27.111} & \textbf{0.013} & \textbf{0.028} & \textbf{1.52} & \textbf{3.59} & \textbf{0.75} & \textbf{1.57} & 395.06 & 1147.14 \\
				CAFE ($c = 1.0$) & 6.49 & \textbf{16.10} & \underline{6.34} & \textbf{15.53} & \underline{14.418} & \underline{27.112} & \textbf{0.013} & \textbf{0.028} & 1.71 & 3.87 & 0.86 & 1.73 & 404.92 & 1164.35 \\
				\bottomrule
			\end{tabular}
		\end{center}
	\end{table*}
	
	\begin{table*}[tbp]
		\caption{Sensitivity of the attribution methods applied to MLPs, with an identical setup as in Table \ref{tab:real-data-infidelity}.}
		\label{tab:real-data-sensitivity}
		\footnotesize
		\centering
		\begin{center}
			\begin{tabular}{ c c c c c c c c }
				\toprule
				\multirow{3}{*}{\textbf{Method}} & \multicolumn{7}{c}{\textbf{Attribution Sensitivity ($\downarrow$)}} \\
				\cmidrule(r){2-8}
				& \textbf{COMPAS} & \textbf{HELOC} & \textbf{Adult} &\textbf{German} & \textbf{Titanic} & \textbf{MIMIC-IV} & \textbf{Covertype} \\
				\midrule
				Gradient $\cdot$ Input & 0.231 & 0.194 & 0.050 & \underline{0.025} & 0.131 & 0.264 & 0.588 \\
				LRP & 0.231 & 0.194 & 0.050 & \underline{0.025} & 0.131 & 0.264 & 0.404 \\
				DeepLIFT Rescale & 0.149 & 0.110 & \textbf{0.045} & \textbf{0.023} & \underline{0.044} & \underline{0.124} & 0.229 \\
				GradientSHAP & 0.311 & 0.648 & 0.188 & 0.030 & 0.266 & 0.444 & 1.080 \\
				Integrated Gradients & 0.150 & 0.112 & \textbf{0.045} & \textbf{0.023} & 0.045 & 0.126 & 0.253 \\
				SmoothGrad & 0.383 & 0.496 & 0.212 & 0.258 & 0.474 & 0.592 & 0.631 \\
				KernelSHAP \faHourglassHalf{} & 0.255 & 2.570 & 0.692 & 0.297 & 0.284 & 0.666 & 1.620 \\
				Shapley Value Sampling \faHourglassHalf{} & 0.168 & 0.325 & 0.088 & 0.040 & 0.109 & 0.308 & 0.664 \\
				LIME \faHourglassHalf{} & 0.228 & 1.194 & 0.205 & 0.645 & 0.236 & 0.546 & 1.411 \\
				\midrule
				CAFE ($c = 0.0$) & 0.231 & 0.194 & 0.050 & \underline{0.025} & 0.131 & 0.264 & 0.404 \\
				CAFE ($c = 0.5$) & \underline{0.148} & \underline{0.102} & \underline{0.046} & \textbf{0.023} & 0.064 & 0.159 & \underline{0.096} \\
				CAFE ($c = 1.0$) & \textbf{0.124} & \textbf{0.059} & \textbf{0.045} & \textbf{0.023} & \textbf{0.042} & \textbf{0.108} & \textbf{0.093} \\
				\bottomrule
			\end{tabular}
		\end{center}
	\end{table*}
	
	\textbf{Data with Conflicting Features.} In these experiments, we empirically test the ability of CAFE and various other baselines to correctly identify the effects of conflicting features. To this end, we construct a collection of several datasets with different parameters using a controlled data generation process, which enables us to establish the expected ``ground-truth" attribution scores for the different features. The generated data contains pairs of continuous and binary features, with the binary features canceling the effects of the continuous features when positive. We experiment with two kinds of models: randomly initialised MLPs trained on the training partitions of the generated datasets and procedurally constructed models with known reasoning. Both types of models provide unique advantages for our evaluation: the trained models are likely to be more realistic while the procedurally constructed models are guaranteed to be perfectly aligned with the data generation process.
	
	The data generation process we use for our conflicting features experiments is parameterised with a dimension $D$ specifying the number of continuous features, a cancelation likelihood $l$ determining the likelihood of each feature in each sample being canceled, a standard deviation $s$ to be used for constructing a diagonal covariance matrix of a multivariate Gaussian distribution and a range $[a, b]$ for the randomly generated feature weights. During the data generation, we first sample the weights $\vw \in \mathbb{R}^D$ for each of $D$ continuous features from a uniform distribution $U(a, b)$. We then generate each sample $[\vx, \vc]$ by drawing the continuous features $\vx \in \mathbb{R}^D$ from a Gaussian distribution $\mathcal{N}(\mathbf{0}, \text{diag}(s^2))$ and independently drawing the binary cancelation features $\vc \in \{0, 1\}^D$ from a Bernoulli distribution $\text{Bernoulli}(l)$. The continuous label for each sample is then derived as $y = \sum_i \mathbb{1}[c_i = 0] w_i x_i$. That is, the influence of each continuous feature is only reflected in the label if the corresponding cancelation feature is zero and is canceled otherwise. With the labels derived in this way, the natural ``ground-truth" attribution score for the $i$-th continuous feature is $x_i w_i$, while the score for the corresponding cancelation feature is $-\mathbb{1}[c_i = 1] w_i x_i$.
	
	Using the described data generation procedure, we generate a collection of datasets with different parameters and conduct a series of experiments. For each experiment, we generated five datasets with $10000$ samples using different random seeds. These datasets were split into training, validation and test portions in the ratio of $6 : 2 : 2$. On each of the training portions, we trained five differently initialised neural models with two hidden layers for $2000$ epochs using the AdamW optimizer \cite{loschchilov-adamw} and selected the model with the best performance on the validation dataset. All the selected models achieved a low error on the test dataset, giving us confidence that their internal reasoning should be relatively close to the data generation process. In addition to the trained models, we also constructed several synthetic models in a procedural fashion, which solve the task exactly by computing positive and negative versions of each feature, subtracting the corresponding cancelation feature multiplied by a large edge weight and weighing the result by the weight corresponding with the given feature. While these models may potentially be less realistic than the trained models, they are guaranteed to precisely match the data generation process.
	
	For both the trained and the procedurally constructed models, we then evaluated CAFE and the baseline feature attribution methods\footnote{Our experimental comparison does not include DeepLIFT RevealCancel, as the only available software implementation is for Tensorflow v1, which is incompatible with the rest of our software tooling.} by computing the attribution scores and comparing them with the expected ``ground-truth" scores using the root-mean-square error (RMSE). Attribution scores for methods taking in a reference input (including CAFE) were computed with respect to a zero vector. In order to make the comparison fair to methods not distinguishing between positive and negative scores, we only considered the combined CAFE scores obtained by subtracting the negative scores from the positive ones. In each experiment, we averaged the results over the five generated datasets and (in case of the trained models) selected models. The detailed results are captured in Tables \ref{tab:synthetic-data-results-train} and \ref{tab:synthetic-data-results-proc}.
	
	\begin{table*}[tbp]
		\caption{Complexity of attribution methods on MLPs trained on various datasets. Best results overall are bolded, second-best results are underlined. \faHourglassHalf{} marks methods more than 10-times slower than CAFE on average.}
		\label{tab:real-data-complexity}
		\footnotesize
		\centering
		\begin{center}
			\begin{tabular}{ c c c c c c c c }
				\toprule
				\multirow{3}{*}{\textbf{Method}} & \multicolumn{7}{c}{\textbf{Attribution Complexity ($\downarrow$)}} \\
				\cmidrule(r){2-8}
				& \textbf{COMPAS} & \textbf{HELOC} & \textbf{Adult} &\textbf{German} & \textbf{Titanic} & \textbf{MIMIC-IV} & \textbf{Covertype} \\
				\midrule
				Gradient $\cdot$ Input & 1.011 & 2.487 & 1.536 & \textbf{2.486} & 0.869 & \textbf{1.731} & 1.864 \\
				LRP & 1.011 & 2.487 & 1.536 & \textbf{2.486} & 0.869 & \textbf{1.731} & \underline{1.848} \\
				DeepLIFT Rescale & 1.004 & 2.515 & 1.563 & \underline{2.488} & \underline{0.841} & 1.747 & \textbf{1.781} \\
				GradientSHAP & 1.011 & 2.504 & 1.590 & \underline{2.488} & 0.845 & 1.744 & \underline{1.848} \\
				Integrated Gradients & 1.004 & 2.515 & 1.563 & \underline{2.488} & \textbf{0.839} & \underline{1.743} & 1.853 \\
				SmoothGrad & 1.056 & 2.598 & 1.809 & \underline{2.488} & 0.956 & 1.864 & 2.052 \\
				KernelSHAP \faHourglassHalf{} & 1.174 & 2.718 & 2.034 & 3.792 & 1.333 & 2.570 & 3.745 \\
				Shapley Value Sampling \faHourglassHalf{} & 1.026 & 2.529 & 1.552 & 3.778 & 1.191 & 2.528 & 3.671 \\
				LIME \faHourglassHalf{} & \textbf{0.941} & 2.488 & 1.580 & 3.296 & 1.238 & 2.414 & 3.708 \\
				\midrule
				CAFE ($c = 0.0$) & 1.011 & 2.487 & 1.536 & \textbf{2.486} & 0.869 & \textbf{1.731} & \underline{1.848} \\
				CAFE ($c = 0.5$) & \underline{0.952} & \underline{2.484} & \textbf{1.517} & 2.489 & 0.874 & 1.766 & 1.904 \\
				CAFE ($c = 1.0$) & 0.984 & \textbf{2.430} & \underline{1.524} & 2.492 & 0.895 & 1.789 & 1.925 \\
				\bottomrule
			\end{tabular}
		\end{center}
	\end{table*}
	
	\begin{table*}[tbp]
		\caption{Structural Infidelity of the attribution methods applied to FT-Transformers, with an identical setup as in Table \ref{tab:ft-real-data-infidelity}.}
		\label{tab:ft-real-data-struct-infidelity}
		\scriptsize
		\centering
		\begin{center}
			\begin{tabular}{ c c c c c c c c c c c c c c c }
				\toprule
				\multirow{4}{*}{\textbf{Method}} & \multicolumn{14}{c}{\textbf{Attribution Structural Infidelity ($\downarrow$)}} \\
				\cmidrule(r){2-15}
				& \multicolumn{2}{c}{\textbf{COMPAS}} & \multicolumn{2}{c}{\textbf{HELOC}} & \multicolumn{2}{c}{\textbf{Adult}} & \multicolumn{2}{c}{\textbf{German}} & \multicolumn{2}{c}{\textbf{Titanic}} & \multicolumn{2}{c}{\textbf{MIMIC-IV}} & \multicolumn{2}{c}{\textbf{Covertype}} \\
				\cmidrule(r){2-3} \cmidrule(r){4-5} \cmidrule(r){6-7} \cmidrule(r){8-9} \cmidrule(r){10-11} \cmidrule(r){12-13} \cmidrule(r){14-15}
				& \textbf{S} & \textbf{L} & \textbf{S} & \textbf{L} & \textbf{S} & \textbf{L} & \textbf{S} & \textbf{L} & \textbf{S} & \textbf{L} & \textbf{S} & \textbf{L} & \textbf{S} & \textbf{L} \\
				\midrule
				Gradient $\cdot$ Input & 2.004 & 2.418 & 1.318 & 1.707 & 21.106 & 24.270 & 0.471 & 0.745 & 1.292 & 2.324 & 8.343 & 14.832 & 13.524 & 17.249 \\
				DeepLIFT Rescale & 2.004 & 2.419 & 1.317 & 1.706 & 21.061 & 24.225 & 0.469 & 0.740 & 1.304 & 2.343 & 8.332 & 14.586 & 13.599 & 17.284 \\
				GradientSHAP & 1.932 & 2.340 & 1.307 & 1.692 & 21.006 & 24.136 & 0.470 & 0.741 & 1.300 & 2.325 & 8.648 & 15.085 & 13.620 & 17.301 \\
				Integrated Gradients & 1.921 & 2.332 & 1.303 & 1.691 & 20.964 & 24.100 & 0.468 & 0.739 & 1.294 & 2.309 & 8.573 & 15.009 & 13.565 & 17.259 \\
				SmoothGrad & 2.501 & 2.861 & 1.578 & 1.895 & 21.377 & 24.390 & 0.532 & 0.821 & 1.318 & 2.303 & 9.727 & 16.130 & 13.872 & 17.412 \\
				Shapley Value Sampling \faHourglassHalf{} & 1.931 & 2.342 & 1.308 & 1.693 & 20.934 & 24.051 & 0.473 & 0.736 & 1.344 & 2.384 & 8.402 & 14.541 & \underline{13.391} & 17.116 \\
				\midrule
				CAFE ($c = 0.0$) & 1.917 & 2.326 & \textbf{1.295} & \textbf{1.682} & 20.980 & 24.126 & \textbf{0.410} & \underline{0.639} & \underline{1.169} & 2.103 & 8.211 & 14.473 & \textbf{13.155} & \textbf{16.867} \\
				CAFE ($c = 0.5$) & \textbf{1.900} & \textbf{2.309} & \underline{1.296} & \underline{1.683} & \textbf{20.825} & \textbf{23.984} & \underline{0.411} & \textbf{0.637} & \textbf{1.166} & \textbf{2.076} & \textbf{7.987} & \textbf{13.903} & 13.409 & \underline{17.018} \\
				CAFE ($c = 1.0$) & \underline{1.903} & \underline{2.311} & 1.304 & 1.688 & \underline{20.842} & \underline{23.987} & 0.419 & 0.647 & 1.180 & \underline{2.092} & \underline{8.106} & \underline{13.991} & 13.515 & 17.096 \\
				\bottomrule
			\end{tabular}
		\end{center}
	\end{table*}
	
	\begin{table*}[tbp]
		\caption{Sensitivity of the attribution methods applied to FT-Transformers, with an identical setup as in Table \ref{tab:ft-real-data-infidelity}.}
		\label{tab:ft-real-data-sensitivity}
		\footnotesize
		\centering
		\begin{center}
			\begin{tabular}{ c c c c c c c c }
				\toprule
				\multirow{3}{*}{\textbf{Method}} & \multicolumn{7}{c}{\textbf{Attribution Sensitivity ($\downarrow$)}} \\
				\cmidrule(r){2-8}
				& \textbf{COMPAS} & \textbf{HELOC} & \textbf{Adult} &\textbf{German} & \textbf{Titanic} & \textbf{MIMIC-IV} & \textbf{Covertype} \\
				\midrule
				Gradient $\cdot$ Input & 0.493 & 0.144 & 0.420 & 0.097 & 0.456 & 0.292 & 0.481 \\
				DeepLIFT Rescale & 0.553 & 0.136 & 0.274 & 0.077 & 1.271 & 0.138 & 0.280 \\
				GradientSHAP & 0.530 & 0.441 & 0.732 & 0.499 & 2.679 & 0.983 & 1.250 \\
				Integrated Gradients & \underline{0.169} & \textbf{0.061} & \textbf{0.094} & 0.051 & 0.339 & 0.106 & 0.180 \\
				SmoothGrad & 3.262 & 1.871 & 2.657 & 2.093 & 3.455 & 1.027 & 0.973 \\
				Shapley Value Sampling \faHourglassHalf{} & \textbf{0.163} & 0.214 & 0.178 & 0.358 & 0.143 & \textbf{0.037} & 0.413 \\
				\midrule
				CAFE ($c = 0.0$) & 0.196 & 0.067 & 0.283 & 0.057 & \textbf{0.038} & 0.143 & 0.361 \\
				CAFE ($c = 0.5$) & 0.187 & \underline{0.065} & 0.102 & \underline{0.036} & \underline{0.040} & 0.055 & \underline{0.168} \\
				CAFE ($c = 1.0$) & 0.203 & 0.074 & \underline{0.095} & \textbf{0.032} & 0.043 & \underline{0.052} & \textbf{0.085} \\
				\bottomrule
			\end{tabular}
		\end{center}
	\end{table*}
	
	\begin{table*}[tbp]
		\caption{Complexity of attribution methods on FT-Transformers trained on various datasets. Best results overall are bolded, second-best results are underlined. \faHourglassHalf{} marks methods more than 10-times slower than CAFE on average.}
		\label{tab:ft-real-data-complexity}
		\footnotesize
		\centering
		\begin{center}
			\begin{tabular}{ c c c c c c c c }
				\toprule
				\multirow{3}{*}{\textbf{Method}} & \multicolumn{7}{c}{\textbf{Attribution Complexity ($\downarrow$)}} \\
				\cmidrule(r){2-8}
				& \textbf{COMPAS} & \textbf{HELOC} & \textbf{Adult} &\textbf{German} & \textbf{Titanic} & \textbf{MIMIC-IV} & \textbf{Covertype} \\
				\midrule
				Gradient $\cdot$ Input & 1.015 & 2.251 & \textbf{1.330} & \textbf{1.107} & 0.473 & \textbf{1.381} & \textbf{1.434} \\
				DeepLIFT Rescale & 1.035 & \underline{2.227} & \underline{1.347} & \underline{1.118} & \underline{0.462} & \underline{1.421} & 1.524 \\
				GradientSHAP & 0.946 & 2.366 & 1.705 & 1.138 & 0.465 & 1.460 & \underline{1.494} \\
				Integrated Gradients & 0.949 & 2.373 & 1.738 & 1.140 & \textbf{0.448} & 1.474 & 1.523 \\
				SmoothGrad & \textbf{0.927} & \textbf{2.157} & 1.653 & 1.185 & 0.510 & 1.571 & 1.635 \\
				Shapley Value Sampling \faHourglassHalf{} & 0.973 & 2.376 & 1.749 & 3.521 & 0.747 & 2.310 & 3.739 \\
				\midrule
				CAFE ($c = 0.0$) & \underline{0.943} & 2.292 & 1.399 & 2.389 & 0.672 & 1.619 & 1.636 \\
				CAFE ($c = 0.5$) & 1.052 & 2.329 & 1.589 & 2.398 & 0.753 & 1.602 & 1.671 \\
				CAFE ($c = 1.0$) & 1.084 & 2.355 & 1.645 & 2.400 & 0.796 & 1.651 & 1.713 \\
				\bottomrule
			\end{tabular}
		\end{center}
	\end{table*}
	
	\subsection{Real Data Experiments}
	\label{sec:real-details}
	\textbf{Datasets.} As described in the main text, for our real data experiments, we used four datasets and pre-trained MLPs from the OpenXAI benchmark \cite{agarwal-openxai} in addition to our own MLP models trained on Titanic \cite{titanic-frank} and Covertype \cite{covertype-blackard} datasets, as well as a subset of the MIMIC-IV medical database \cite{johnson-mimic-iv-nature}. We also used a set of FT-Transformer models trained on the same datasets (including those from OpenXAI).
	
	Our dataset for the MIMIC-IV experiment was constructed from a cohort of all patients with recorded ICU stays. For each patient, we extracted the admission weight, heart rate (HR), mean blood pressure (MBP), respiratory rate (RR), oxygen saturation (SpO2), body temperature, admission type, gender and admission age values from the first hour of their first ICU stay. These variables were selected as the most relevant in cooperation with our clinical collaborators. Patients for which some of the data were unavailable or contained clinically unrealistic values were excluded from the dataset. We took the 30-day patient mortality as the label and our prediction target. For the purposes of training our models on the three additional datasets, the continuous features were scaled using a robust scaler from scikit-learn \cite{pedregosa-scikit-learn} using the quantile range $[10, 90]$ while the categorical features were encoded using a one-hot encoding. For MIMIC-IV only, we randomly upsampled the patients with the positive label to form one-third of all samples, so as to reduce label imbalance. The upsampling was performed for a training set only.
	
	\textbf{Models.} In each of the experiments using newly trained models (all except the experiments using pretrained MLPs from OpenXAI), we trained a collection of five differently initialised models. The used model architectures in our MLP experiments varied for each dataset and were informed by the performance of the corresponding models. For Titanic, we used a ReLU network with two hidden layers and a hidden dimension of 256, for MIMIC-IV, we used a ReLU network with two hidden layers and a hidden dimension of 32, and for Covertype, we used a larger GELU network with four hidden layers and a hidden dimension of 256. All models were trained for 100 epochs using the AdamW optimizer. Our MLP models reached an average accuracy of $0.78$ for Titanic, $0.86$ for MIMIC-IV and $0.95$ for Covertype, as evaluated on a withheld test set. In the FT-Transformer experiments, we used an identical architecture for all datasets, with internal model dimension of 128, 8 attention heads, feedforward layer expansion factor 4/3, attention dropout 0.2, feedforward layer dropout 0.1, two residual blocks and GELU as the internal activation function. The FT-Transformer models reached an average accuracy of $0.85$ for COMPAS, $0.73$ for HELOC, $0.83$ for Adult, $0.64$ for German, $0.77$ for Titanic, $0.86$ for MIMIC-IV and $0.94$ for Covertype. Note that we did not make any significant effort to reach the maximum possible accuracy on the given datasets, as the attribution methods should be equally applicable to less performant models.
	
	\textbf{Evaluation metrics.} We consider faithfulness as our primary performance indicator on real data, as it amounts to the explanation accurately describing the behaviour of the model. We operationalise faithfulness through the inverse notions of infidelity~\cite{infidelity-sensitivity-explanations} and structural infidelity~\cite{ayoobi-sparx}. These metrics perform meaningful perturbations to the explained data samples (Gaussian for continuous features and categorical for discrete features) and measure the error between the predicted model outputs based on a linear extrapolation from the attribution scores and actual model outputs for the perturbed samples. Structural infidelity additionally considers attributions with respect to internal model neurons, evaluating the utility of the attribution method for explaining intermediate model representations in addition to the final outputs. Infidelity generalises several other notions of explanation correctness, which only consider more constrained perturbations, such as faithfulness correlation~\cite{bhatt-faithfulness-corr} and sensitivity-n~\cite{ancona-towards-attribution}.
	
	We also use the \emph{max-sensitivity} metric to assesses robustness of explanations by observing how much the attribution scores change for small perturbations in the inputs \cite{infidelity-sensitivity-explanations}, though robustness evaluated in this way can be at odds with fidelity if the behaviour of the explained model rapidly changes for slightly different inputs. Finally, the \emph{complexity} metric aims to gauge the difficulty of understanding the explanation by considering the entropy of the fractional contribution of the attribution scores the sum of these scores \cite{bhatt-faithfulness-corr}. We consider complexity to be of relatively lower importance (a complex but correct explanation is likely to be more useful than a simple but incorrect explanation), but report it for completeness.
	
	For computing the infidelity metrics in all experiments (including the ones for the pre-trained OpenXAI models), we used a perturbation function adding Gaussian noise with zero mean and a specified standard deviation to continuous features and randomly resampling categorical variables with a specified likelihood. We evaluated the infidelity of the different attribution methods both on smaller perturbations (with a standard deviation of the Gaussian noise of 0.5 and categorical resampling probability of 0.1) and larger perturbations (with a standard deviation of the Gaussian noise of 0.75 and categorical resampling probability of 0.2). To compute the infidelity metric itself, we used the respective implementation from Captum \cite{kokhlikyan-captum}, using the normalization option to ensure that the metric is invariant to a constant scaling of the attributions. The sensitivity metric was also computed by its implementation in Captum, using the default perturbation radius of $0.02$.
	
	\textbf{Results.} The results of our key experiments were reported or linked in the main text. In Tables \ref{tab:real-data-complexity} and \ref{tab:ft-real-data-complexity}, we additionally report the complexity of the explanations computed by the different methods. CAFE is highly competitive in terms of explanation complexity on MLP models, reaching similar results as other methods and sometimes outperforming them. On FT-Transformer models, CAFE explanations were generally more complex compared to explanations returned by other methods. We believe that this might be due to the more complex architecture and behaviour of FT-Transformers requiring potentially more complex explanations. As we mentioned above, we consider complexity to be of lower importance than infidelity, as a complex but correct explanation is likely to be more helpful than a simple but incorrect explanation.
	
	\subsection{Conflict Prevalence in Real Models}
	\label{apd:conflict-prevalence}
	While our experiments have demonstrated that reflecting conflicts can improve attribution infidelity and sensitivity, a natural question to consider is how prevalent these conflicts tend to be. To address this question, we computed the average share of inactive neurons with non-zero inputs for each of the models used in our real-data evaluation when applied to samples from its corresponding test set. The values observed for MLPs ranged from $31\%$ for the MIMIC-IV model to $80\%$ for the HELOC model, with the average over different models being $49\%$. For FT-Transformers, we observed values from $24\%$ for the HELOC model to $37\%$ for the Covertype model, with the average over different models being $30\%$. This suggests that conflicts are relatively common in real models.
	
	\subsection{Real-World Conflict Example}
	\label{apd:real-world-conflict}
	
	\begin{figure*}[!htb]
		\centering
		\begin{subfigure}[m]{0.50\textwidth}
			\centering
			\includegraphics[width=0.7\textwidth]{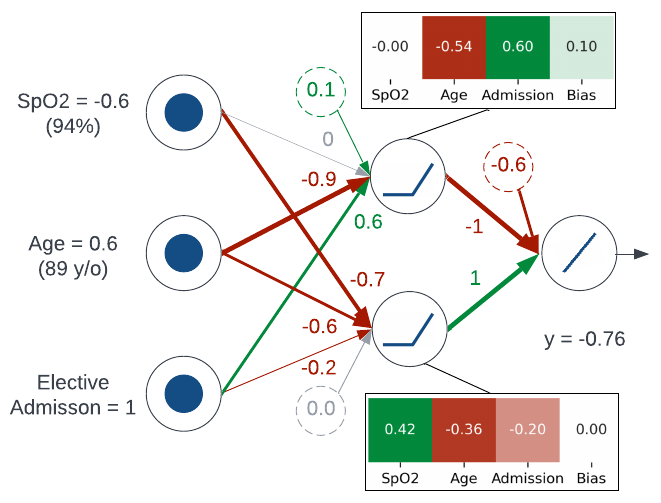}
			\caption{A MIMIC-IV network}
			\label{fig:example-mimic-iv-nn}
		\end{subfigure}
		\begin{subfigure}[m]{0.45\textwidth}
			\tiny
			\centering
			\begin{center}
				\begin{tabular}{ c c c c c }
					\toprule
					\multirow{2}{*}{\textbf{Method}} & \multicolumn{4}{c}{\textbf{Attribution Scores}} \\
					\cmidrule(r){2-5}
					& \textbf{SpO2} & \textbf{Age} & \textbf{Admission} & \textbf{Bias} \\
					\midrule
					G $\cdot$ I & 0.0 & 0.54 & -0.6 & --- \\
					LRP & 0.0 & 0.54 & -0.6 & --- \\
					DL-R & 0.0 & 0.54 & -0.6 & --- \\
					DL-RC & 0.21 & 0.18 & -0.46 & --- \\
					GS & 0.0 & 0.54 & -0.6 & --- \\
					IG & 0.0 & 0.54 & -0.6 & --- \\
					SG & -0.17 & 0.22 & 0.32 & --- \\
					KS & 0.14 & 0.24 & -0.44 & --- \\
					SVS & 0.20 & 0.15 & -0.4 & --- \\
					LIME & 0.13 & 0.16 & -0.4 & --- \\
					\midrule
					CAFE (0.0) & 0.0$^+$/0.0$^-$ & 0.54$^+$/0.0$^-$ & 0.0$^+$/0.6$^-$ & 0.0$^+$/0.7$^-$ \\
					CAFE (1.0) & 0.42$^+$/0.0$^-$ & 0.54$^+$/0.27$^-$ & 0.0$^+$/0.75$^-$ & 0.0$^+$/0.7$^-$ \\
					\bottomrule
				\end{tabular}
			\end{center}
			\caption{
				\label{fig:example-mimic-iv-scores}Attribution scores for the NN in Figure \ref{fig:example-mimic-iv-nn}. The used baseline methods are the same and in the same order as in Table \ref{tab:real-data-infidelity}, except DeepLIFT RevealCancel (DL-RC).}
		\end{subfigure}
		
		\caption{
			A real-world example of a network with conflicting features. The heatmaps in the network visualization indicate the contributions of the input features to the pre-activations of the corresponding neurons.
		}
		\label{fig:example-mimic-iv}
	\end{figure*}
	
	To further exemplify the phenomena of feature conflicts and bias effects in real models, we present a simplified circuit (i.e., a subset of a neural model) extracted from an MLP trained on MIMIC-IV mortality prediction. The circuit is visualised in Figure \ref{fig:example-mimic-iv}. The top hidden neuron activates more strongly for patients with an elective hospital admission (i.e., pre-arranged, non-urgent admission) and patients that are younger, lowering the predicted risk of death. In contrast, the bottom hidden neuron increases the predicted risk. The neuron activates for patients with lower blood oxygen levels (SpO2), but this effect can be counteracted by higher age and elective hospital admission. This behaviour is reasonable from a clinical perspective, as elderly patients can exhibit lower normal levels of blood oxygen \cite{britto-spo2-aging}. Additionally, the biases of the model reduce the predicted risk, which is likely because the underlying data is imbalanced with a majority of patients surviving.
	
	For a patient with features as illustrated in the figure, only the first neuron activated, while the positive effect of a low SpO2 on the second neuron got canceled by the age and elective admission features. This prompted all gradient-based methods (except for SmoothGrad) to conclude that the low SpO2 had no effect on the output. While the resulting explanations are simpler, they also miss an important aspect of the model's reasoning. Sampling-based methods and DeepLIFT RevealCancel surface some of the canceled effects, but only partially and imperfectly. CAFE is the only method enabling fine-grained control over the degree to which internal conflicts are reflected in the scores, distinguishing between positive and negative features (highlighting the dual effect of age in CAFE (1.0) scores), and capturing the influence of the biases.
	
	\subsection{Conflicts and Distributional Shifts}
	\begin{figure*}[tb]
		\centering
		\begin{subfigure}[m]{0.3\textwidth}
			\includegraphics[width=0.9\textwidth]{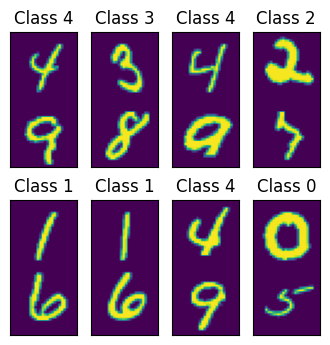}
			\caption{Top and bottom digit both predict class label.}
			\label{fig:subfig1}
		\end{subfigure}
		\hfill
		\begin{subfigure}[m]{0.3\textwidth}
			\includegraphics[width=0.9\textwidth]{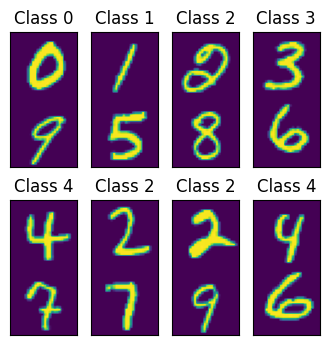}
			\caption{Only top digit predicts class label. Bottom digit randomised.}
			\label{fig:subfig2}
		\end{subfigure}
		\hfill
		\begin{subfigure}[m]{0.3\textwidth}
			\includegraphics[width=0.9\textwidth]{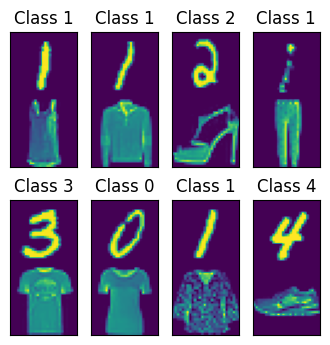}
			\caption{Only top digit predicts class label. Bottom replaced with Fashion-MNIST.}
			\label{fig:subfig3}
		\end{subfigure}
		\caption{A random selection of instances from three concatenations of MNIST and Fashion-MNIST.}
		\label{fig:concatmnist}
	\end{figure*}
	
	\textbf{Details on Input Feature Conflict Experiments.} For each input $[c,s,n]$ we generate categorical features $c, s \in C = \{0,1,2,3,4\}$ and a continuous noise feature $n$ drawn from a Gaussian $\mathcal{N}(0, 1)$. The labels are simply follow the first categorical feature as $y=c$. On the training dataset, we let the categorical feature $s$ be perfectly correlated with $c$ as $s=c$, so that a model trained on this distribution could equally well use $c$ or $s$ to predict $y$. We consider MLPs with two hidden layers of size $\{16,24,32,40\}$ and an FT-Transformer with one transformer layer with $\{4,8\}$ heads and size $\{32,64\}$. We induce a correlation shift on $s$ by letting it be randomly chosen from the set of categories other than the current value of $c$, i.e., $s \in_R C \setminus c$. We simulate a missing feature $s$ by setting it to zero.
	
	\textbf{Details on Latent Feature Conflicts Experiments.} We train four different CNNs. CNN Shallow consists of two convolutional layers of size 128 followed by two fully connected layers with a hidden dimension of 128. CNN Deep doubles the number of convolutional layers. CNN Narrow consists of four convolutional layers of sizes 8, 16, 32, 64 with two fully connected layers with a hidden dimension of 64. CNN Wide doubles the sizes of each layer.
	
	We construct our custom version of MNIST by concatenating images of the first five digits at the top with images of the second five digits at the bottom. The labels are constructed by taking the labels of the first five digits only. For the training dataset, the concatenation is such that both the top and bottom of the image are correlated with the label, while for the test dataset the bottom digit is randomised so that only the top digit is predictive. For the concatenation of MNIST with Fashion-MNIST, we simply replace the bottom digit with images from the Fashion-MNIST dataset. The label remains determined by the top digit. We illustrate the datasets in Figure~\ref{fig:concatmnist}
	
	\section{CAFE Attribution Examples}
	\label{sec:cafe-attriubtion-examples}
	
	\subsection{Tabular Data Examples}
\begin{figure*}[!p]
		\centering
		\includegraphics[width=0.6\linewidth]{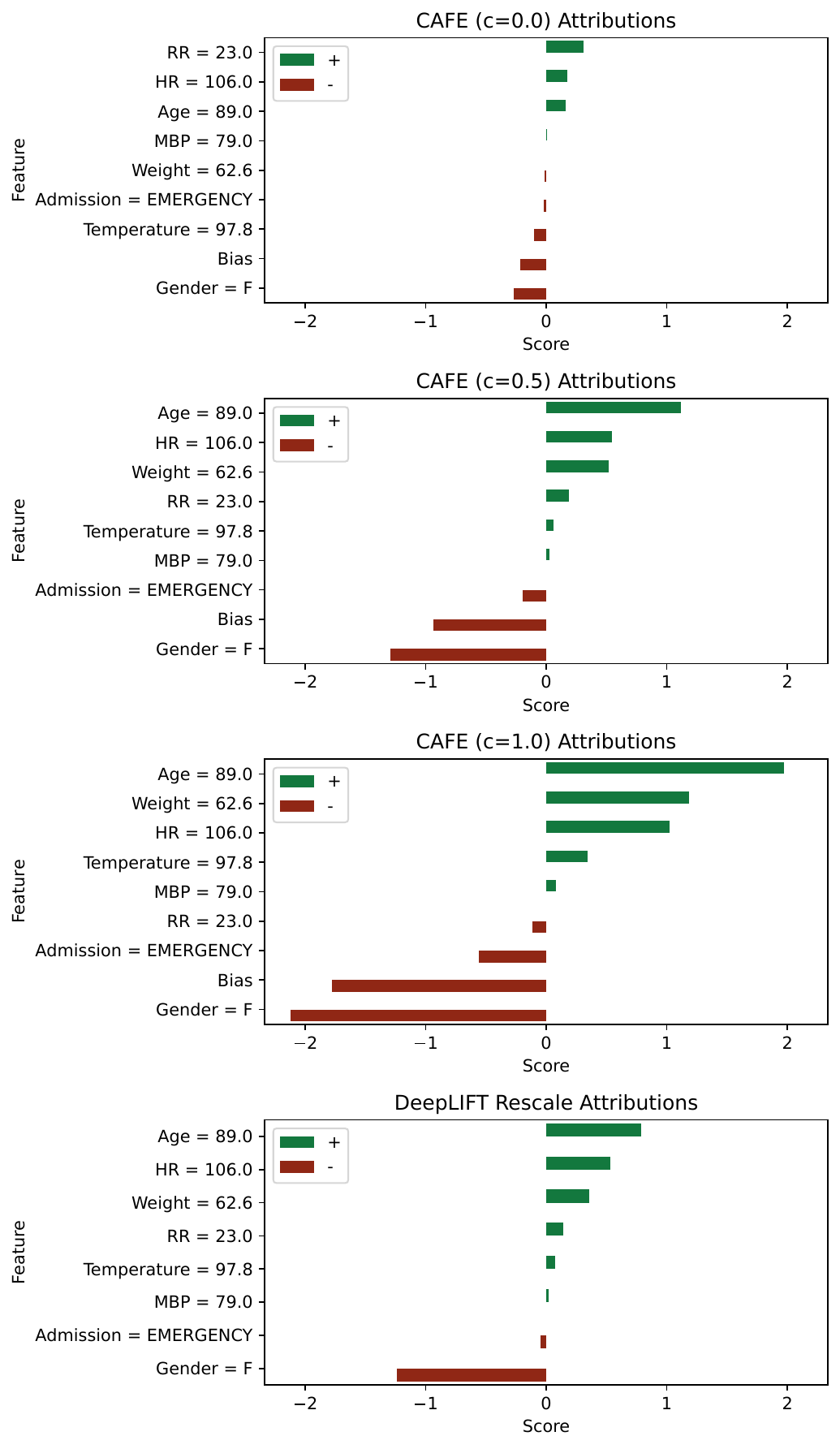}
		\caption{CAFE and DeepLIFT Rescale attribution scores for a correctly classified positive sample from our MIMIC-IV dataset (prediction score 0.51)}
		\label{fig:tp-mimic-explanations}
	\end{figure*}

\begin{figure*}[!p]
		\centering
		\includegraphics[width=0.6\linewidth]{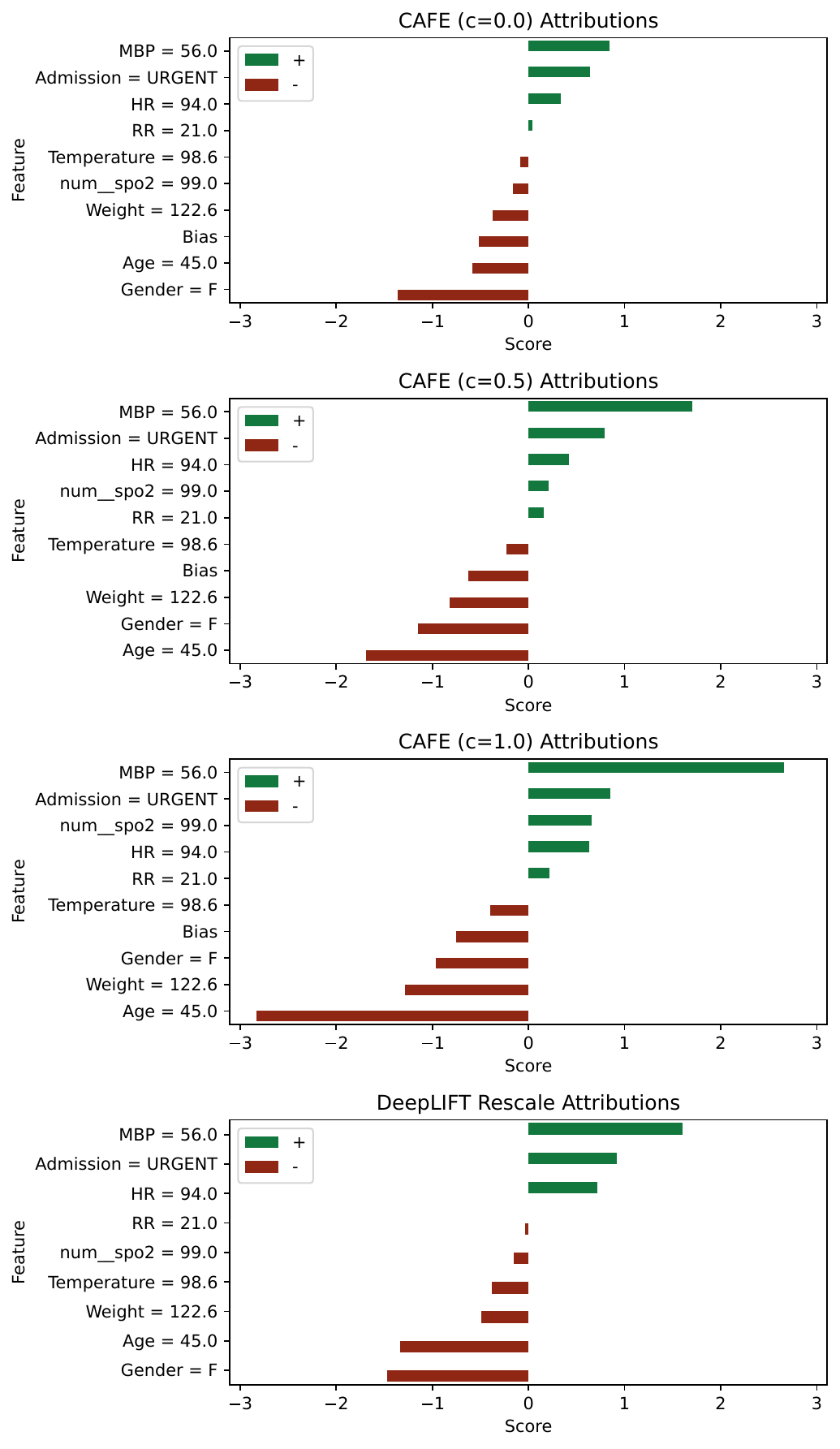}
		\caption{CAFE and DeepLIFT Rescale attribution scores for a correctly classified negative sample from our MIMIC-IV dataset (prediction score 0.23)}
		\label{fig:tn-mimic-explanations}
	\end{figure*}

\begin{figure*}[!p]
		\centering
		\includegraphics[width=0.6\linewidth]{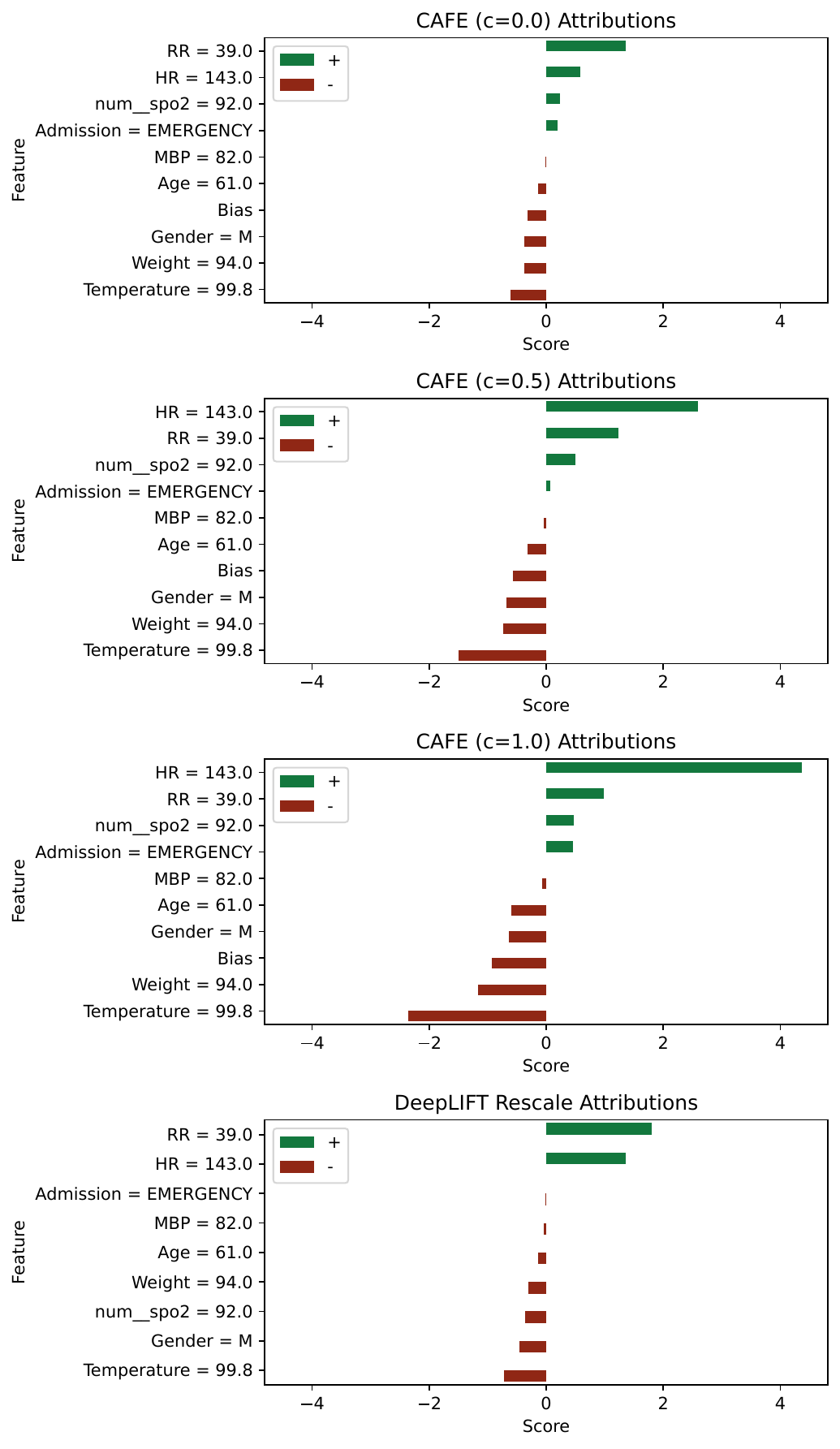}
		\caption{CAFE and DeepLIFT Rescale attribution scores for an incorrectly classified negative sample from our MIMIC-IV dataset (prediction score 0.63)}
		\label{fig:fp-mimic-explanations}
	\end{figure*}

\begin{figure*}[!p]
		\centering
		\includegraphics[width=0.6\linewidth]{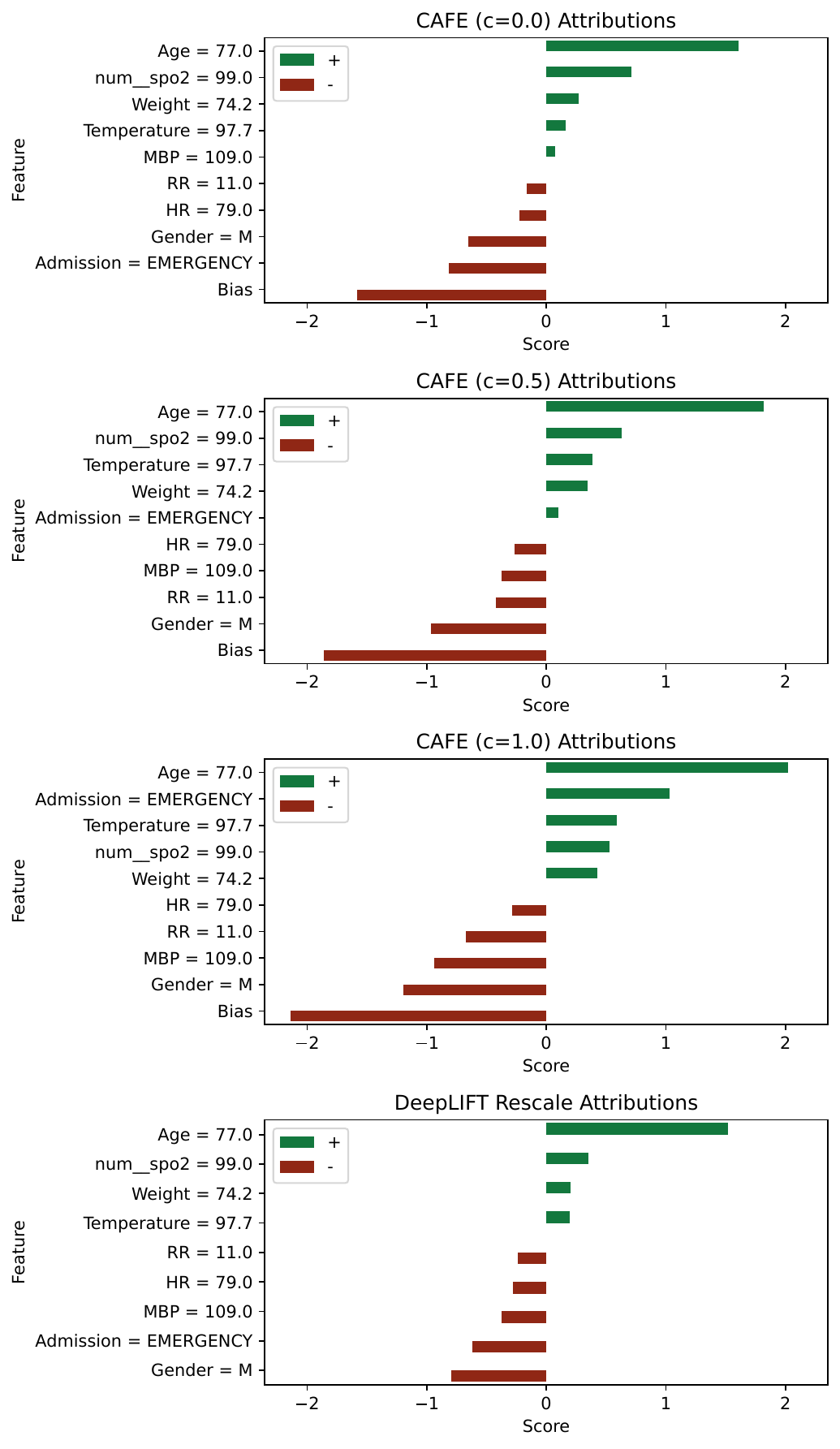}
		\caption{CAFE and DeepLIFT Rescale attribution scores for an incorrectly classified positive sample from our MIMIC-IV dataset (prediction score 0.35)}
		\label{fig:fn-mimic-explanations}
	\end{figure*}

	In this section, we consider example CAFE attribution scores for four data points from our MIMIC-IV dataset — one for a true positive sample (Figure \ref{fig:tp-mimic-explanations}), a true negative sample (Figure \ref{fig:tn-mimic-explanations}), a false positive sample (Figure \ref{fig:fp-mimic-explanations}) and a false negative sample (Figure \ref{fig:fn-mimic-explanations}). The data points were selected as the first occurring representatives of the given category in our randomly ordered test set. In all examples, we compare the CAFE scores with DeepLIFT Rescale, as it was the overall best-performing gradient-based baseline in our MIMIC experiments. For fairness and ease of comparison between the two methods, we only consider the combined positive and negative scores for CAFE.
	
	In the first example (Figure \ref{fig:tp-mimic-explanations}) showing attributions for a true positive, we can see that all methods except for CAFE (0.0) considered the high age to be the key feature for predicting a higher likelihood of 30-day mortality, which seems intuitive. It is possible that the age feature partially interfered with the other features or the bias, explaining why it was not well captured by CAFE (0.0). Additionally, we can observe that in comparison to DeepLIFT, CAFE identifies a strong negative effect of the bias. This is to be expected, as the model was trained on a partially imbalanced dataset with a majority of negative samples. Female gender is identified by all methods as a strongly negatively contributing feature, which may be somewhat counterintuitive. Our hypothesis is that the model may be using the gender attribute for a partial representation of its default bias against predicting a positive label, for both of its possible values. This hypothesis also seems to be supported by the attributions scores for the other examples, where the gender attribute always has a relatively strong negative effect on the prediction, regardless of whether it is male or female (though a female gender seems to be associated with stronger negative biases in our examples).
	
	In the second example (Figure \ref{fig:tn-mimic-explanations}), we consider the attribution scores for a correctly classified negative sample. We can see that the most strongly contributing features identified by all the considered methods, are a low mean blood pressure and an urgent admission, both reasonable features in the given context. In comparison with other methods, CAFE (0.5) and CAFE (1.0) identify the relatively low age of 45 as a more important feature, which again seems intuitive. All variants of CAFE capture a non-negligible negative effect of the bias, though relatively less important than in the previous example.
	
	The third example (Figure \ref{fig:fp-mimic-explanations}) considers an incorrectly classified negative sample. In this case, the relatively high respiratory rate and heart rate are most important for the positive classification. This is reasonable, as high rates for these vital signs typically indicate higher distress and an overall worse clinical state of the patient. Variants of CAFE with higher cancelation sensitivity constant consider the heart rate to be the more important of the two features, while CAFE (0.0) and DeepLIFT state the opposite.
	
	Finally, the fourth example (Figure \ref{fig:fn-mimic-explanations}) perhaps most clearly demonstrates the advantage of considering the effect of the bias in the attribution scores. Here, the incorrect negative classification of the model seems to be predominantly caused by the negative bias, which counteracted otherwise strong positively contributing features, such as the high age (77 years). While all variants of CAFE clearly show this, DeepLIFT does not.
	
	\subsection{Image Examples}
	\label{apd:image-examples}
	In this section, we give examples of CAFE image attributions for convolutional neural networks trained on three different datasets — MNIST \cite{lecun1998mnist}, CIFAR-10 \cite{cifar-10} and a synthetic dataset of geometric shapes overlayed on randomly sampled textures from the Describable Textures Dataset \cite{cimpoi-describing-textures}. For comparison, we also visualise the corresponding attributions produced by DeepLIFT Rescale, which was chosen as a fast and high-performing baseline. The CAFE attributions were computed by linearising the convolutional and batch normalisation CNN layers, with a special rule for max-pooling layers performing linear interpolation between the maximum elements for the reference and current inputs.
	
\begin{figure*}[!htbp]
		\centering
		\begin{subfigure}[m]{0.4\textwidth}
			\centering
			\includegraphics[width=0.6\textwidth]{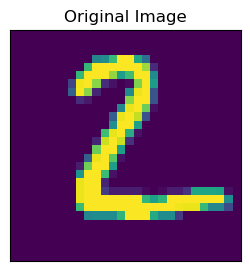}
			\caption{Original sample}
			\label{fig:mnist-sample}
		\end{subfigure}
		\begin{subfigure}[m]{0.57\textwidth}
			\centering
			\includegraphics[width=0.9\textwidth]{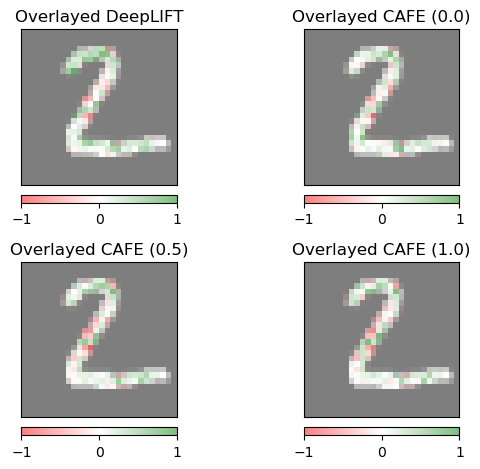}
			\caption{Computed Feature Attributions}
			\label{fig:mnist-attributions}
		\end{subfigure}
		\caption{
			CAFE and DeepLIFT Rescale attributions for a correctly classified sample from the MNIST dataset.
		}
		\label{fig:example-mnist}
	\end{figure*}

	\textbf{MNIST.} An example for the MNIST dataset, showing a number 2, is given in Figure \ref{fig:example-mnist}. We can observe that CAFE with a higher conflict sensitivity highlights more pixels on the diagonal segment near the base of the digit as negatively contributing, possibly since they also occur in number 7.
	
\begin{figure*}[!htbp]
		\centering
		\begin{subfigure}[m]{0.4\textwidth}
			\centering
			\includegraphics[width=0.6\textwidth]{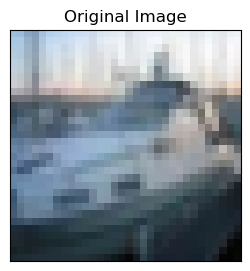}
			\caption{Original sample}
			\label{fig:mnist-cifar}
		\end{subfigure}
		\begin{subfigure}[m]{0.57\textwidth}
			\centering
			\includegraphics[width=0.9\textwidth]{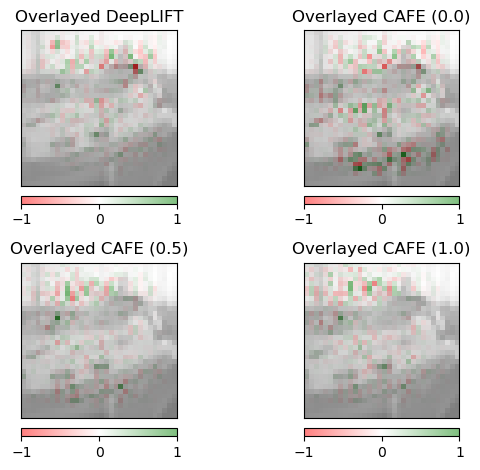}
			\caption{Computed Feature Attributions}
			\label{fig:cifar-attributions}
		\end{subfigure}
		\caption{
			CAFE and DeepLIFT Rescale attributions for an image of a ship from CIFAR-10, incorrectly classified as a truck.
		}
		\label{fig:example-cifar}
	\end{figure*}

	\textbf{CIFAR-10.} In the example for CIFAR-10 (Figure \ref{fig:example-cifar}, we show an explanation for an incorrect model prediction, identifying a ship as a truck. Note that CAFE(0.5) and CAFE(1.0) explanations highlight more of the background of the image, suggesting a possible reason why it has been misclassified.
	
\begin{figure*}[!htbp]
		\centering
		\begin{subfigure}[m]{0.4\textwidth}
			\centering
			\includegraphics[width=0.6\textwidth]{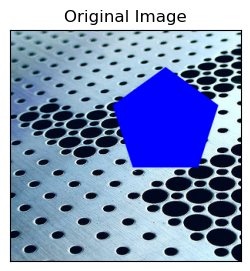}
			\caption{Original sample}
			\label{fig:synthetic-shapes-cifar}
		\end{subfigure}
		\begin{subfigure}[m]{0.57\textwidth}
			\centering
			\includegraphics[width=0.9\textwidth]{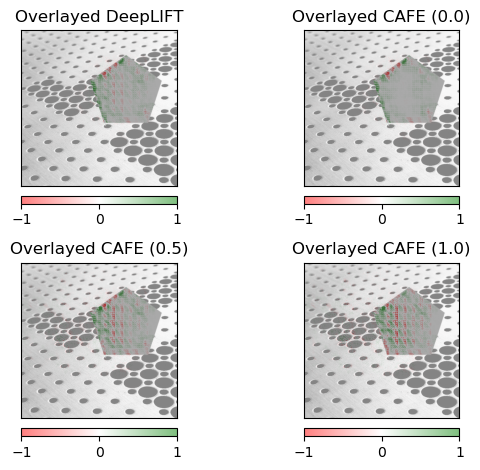}
			\caption{Computed Feature Attributions}
			\label{fig:synthetic-shapes-attributions}
		\end{subfigure}
		\caption{
			CAFE and DeepLIFT Rescale attributions for a correctly classified image of a pentagon from the synthetic shapes dataset.
		}
		\label{fig:example-synthetic-shapes}
	\end{figure*}

	\textbf{Synthetic Shapes.} In the final example, we show an explanation for a correctly classified pentagon from the synthetic shapes dataset (Figure \ref{fig:example-synthetic-shapes}). We can observe that CAFE with an increased conflict sensitivity highlights more of the internal pentagon area as negatively contributing to the classification, while the positive contributions are focused near its leftmost vertex. This is intuitive, as the vertex is the distinguishing factor of the shape, while its interior is similar to a rectangle and other shapes in the dataset.
	
\end{document}